\documentclass[11pt]{article}

\usepackage[utf8]{inputenc}
\usepackage[T1]{fontenc}
\usepackage{lmodern}
\usepackage[scaled=0.92]{helvet}
\usepackage[a4paper,margin=24mm,headheight=22pt,headsep=10pt,footskip=22pt]{geometry}
\usepackage{microtype}
\usepackage{setspace}
\usepackage{graphicx}
\usepackage{multirow}
\usepackage{amsmath,amssymb,amsfonts}
\usepackage{amsthm}
\usepackage{mathrsfs}
\usepackage[title]{appendix}
\usepackage[table]{xcolor}
\usepackage{colortbl}
\usepackage{threeparttable}
\usepackage{subcaption}
\usepackage{wrapfig}
\usepackage{textcomp}
\usepackage{manyfoot}
\usepackage{booktabs}
\usepackage{tabularx}
\usepackage{array}
\usepackage{algorithm}
\usepackage{algorithmicx}
\usepackage{algpseudocode}
\usepackage{listings}
\usepackage{bm}
\usepackage{xspace}
\usepackage{tikz}
\usepackage[edges]{forest}
\usepackage{tcolorbox}
\usepackage{titlesec}
\usepackage{enumitem}
\usepackage[section]{placeins}
\usepackage{caption}
\usepackage{fancyhdr}
\usepackage{lastpage}
\usepackage[numbers,sort&compress]{natbib}
\usepackage{xurl}
\usepackage{hyperref}
\usepackage[capitalise]{cleveref}
\usepackage{marvosym}
\usepackage{orcidlink}
\usepackage{fontawesome5}

\newcommand{\CoverTitleFont}{\sffamily\bfseries}
\newcommand{\CoverSubtitleFont}{\sffamily\bfseries}

\usetikzlibrary{arrows.meta,calc,positioning,shapes.geometric}

\definecolor{ReportNavy}{HTML}{0F2742}
\definecolor{ReportBlue}{HTML}{1A5F7A}
\definecolor{ReportTeal}{HTML}{159895}
\definecolor{ReportAccent}{HTML}{D95D39}
\definecolor{ReportSand}{HTML}{F6F2EA}
\definecolor{ReportInk}{HTML}{24303F}
\definecolor{ReportLine}{HTML}{D7DEE7}
\definecolor{ReportMuted}{HTML}{6C7A89}
\definecolor{ReportNight}{HTML}{081827}
\definecolor{ReportSky}{HTML}{74AFD6}
\definecolor{ReportMint}{HTML}{3BC3B2}
\definecolor{ReportCard}{HTML}{FFFDF8}
\definecolor{ReportMist}{HTML}{D8E3EE}

\definecolor{rowgray}{gray}{0.95}
\definecolor{geomBG}{HTML}{F2F9FF}
\definecolor{visBG}{HTML}{FFF9F2}
\definecolor{hidden-draw}{RGB}{128,128,128}
\definecolor{yellow}{RGB}{255,228,123}
\definecolor{purple}{RGB}{144,153,196}
\definecolor{hidden-blue}{RGB}{194,232,247}
\definecolor{green}{RGB}{34,139,34}
\definecolor{orange}{RGB}{255,165,0}
\definecolor{darkgray}{rgb}{0.66, 0.66, 0.66}
\definecolor{gray}{rgb}{0.5,0.5,0.5}
\definecolor{darkblue}{rgb}{0, 0.40, 0.75}

\hypersetup{
  colorlinks=true,
  citecolor=ReportBlue,
  linkcolor=ReportNavy,
  urlcolor=ReportBlue
}

\allowdisplaybreaks

\renewcommand{\headrulewidth}{0.5pt}

\renewcommand{\headrule}{\hbox to\headwidth{\color{ReportLine}\leaders\hrule height \headrulewidth\hfill}}
\fancypagestyle{plain}{%
  \fancyhf{}
  \fancyhead[L]{\sffamily\scriptsize\textcolor{ReportNavy}{Feed-Forward 3D Scene Modeling: A Problem-Driven Perspective}}
  \fancyhead[R]{\sffamily\footnotesize\textcolor{ReportMuted}{\nouppercase{\rightmark}}}
  \fancyfoot[L]{}
  \fancyfoot[R]{\sffamily\footnotesize\textcolor{ReportMuted}{\thepage\ /\ \pageref{LastPage}}}
  \renewcommand{\headrulewidth}{0.5pt}
  
}
\makeatletter
\def\fps@figure{!t}
\def\fps@table{!t}
\makeatother

\titleformat{\section}
  {\Large\sffamily\bfseries\color{ReportNavy}}
  {\colorbox{ReportAccent}{\makebox[2.1em][c]{\color{white}\thesection}}}
  {0.8em}
  {}
  [\vspace{0.35ex}\color{ReportLine}\titlerule]

\titleformat{name=\section,numberless}
  {\Large\sffamily\bfseries\color{ReportNavy}}
  {}
  {0pt}
  {}
  [\vspace{0.35ex}\color{ReportLine}\titlerule]

\titleformat{\subsection}
  {\large\sffamily\bfseries\color{ReportBlue}}
  {\textcolor{ReportAccent}{\thesubsection}}
  {0.6em}
  {}

\titleformat{\subsubsection}
  {\normalsize\sffamily\bfseries\color{ReportInk}}
  {\textcolor{ReportMuted}{\thesubsubsection}}
  {0.55em}
  {}

\titlespacing*{\section}{0pt}{2.0ex plus 0.5ex minus 0.2ex}{1.0ex}
\titlespacing*{\subsection}{0pt}{1.6ex plus 0.4ex minus 0.2ex}{0.7ex}
\titlespacing*{\subsubsection}{0pt}{1.2ex plus 0.3ex minus 0.2ex}{0.5ex}

\tcbset{
  reportbox/.style={
    colback=white,
    colframe=ReportBlue,
    boxrule=0.8pt,
    arc=4pt,
    left=10pt,
    right=10pt,
    top=8pt,
    bottom=8pt
  }
}

\crefname{figure}{Fig.}{Figs.}
\crefname{table}{Tab.}{Tabs.}
\crefname{section}{Sec.}{Secs.}
\crefname{subsection}{Sec.}{Secs.}
\crefname{equation}{Eq.}{Eqs.}

\setlist[itemize]{leftmargin=1.5em,itemsep=0.2em,topsep=0.4em}
\setlist[enumerate]{leftmargin=1.7em,itemsep=0.2em,topsep=0.4em}

\tikzset{
  my-box/.style={
    rectangle,
    draw=hidden-draw,
    rounded corners,
    text opacity=1,
    minimum height=1.5em,
    minimum width=5em,
    inner sep=2pt,
    align=center,
    fill opacity=.5,
  },
  level1-style/.style={
    my-box,
    fill=gray!5,
    text=black,
    font=\scriptsize,
    inner xsep=2pt,
    inner ysep=3pt,
  },
  level2-style/.style={
    my-box,
    fill=gray!10,
    text=black,
    font=\scriptsize,
    inner xsep=2pt,
    inner ysep=3pt,
  },
  level3-style/.style={
    my-box,
    fill=gray!15,
    text=black,
    font=\scriptsize,
    inner xsep=2pt,
    inner ysep=3pt,
  },
  leaf/.style={
    my-box,
    minimum height=1.5em,
    fill=yellow!32,
    text=black,
    align=left,
    font=\scriptsize,
    inner xsep=2pt,
    inner ysep=4pt,
  },
  leaf2/.style={
    my-box,
    minimum height=1.5em,
    fill=purple!27,
    text=black,
    align=left,
    font=\scriptsize,
    inner xsep=2pt,
    inner ysep=4pt,
  },
  leaf3/.style={
    my-box,
    minimum height=1.5em,
    fill=hidden-blue!57,
    text=black,
    align=left,
    font=\scriptsize,
    inner xsep=2pt,
    inner ysep=4pt,
  },
  leaf4/.style={
    my-box,
    minimum height=1.5em,
    fill=green!14,
    text=black,
    align=left,
    font=\scriptsize,
    inner xsep=2pt,
    inner ysep=4pt,
  },
  leaf5/.style={
    my-box,
    minimum height=1.5em,
    fill=orange!16,
    text=black,
    align=left,
    font=\scriptsize,
    inner xsep=2pt,
    inner ysep=4pt,
  }
}

\theoremstyle{plain}

\theoremstyle{remark}

\theoremstyle{definition}

\newcommand{\iconObj}{\textcolor{ReportAccent}{\faCube}}
\newcommand{\iconIn}{\textcolor{ReportTeal}{\faHome}}
\newcommand{\iconOut}{\textcolor{ReportBlue}{\faTree}}
\newcommand{\iconMix}{\iconIn\!+\!\iconOut}

\newcommand\rurl[1]{\href{https://#1}{\nolinkurl{#1}}}

\newcommand{\coverlogo}[2]{%
  \begin{minipage}[c][1.55cm][c]{#1}
    \centering
    \includegraphics[width=#1,height=1.12cm,keepaspectratio]{#2}
  \end{minipage}
}

\newcommand{\makecoverpage}{%
  \begin{titlepage}
    \thispagestyle{empty}
    \begin{tikzpicture}[remember picture,overlay,x=1cm,y=1cm]
      \begin{scope}[shift={(current page.north west)}]
        \fill[ReportNight] (0,0) rectangle (21,-29.7);
        \fill[ReportBlue!45!ReportNight] (11.8,0) -- (21,0) -- (21,-12.1) -- (16.9,-9.4) -- cycle;
        \fill[ReportTeal!58!ReportNight,opacity=0.55] (15.0,-2.9) circle (3.55);
        \fill[ReportAccent!72!ReportNight,opacity=0.34] (18.9,-22.0) circle (4.6);
        \fill[white,opacity=0.035] (11.2,-6.8) -- (21,-11.7) -- (21,-29.7) -- (13.4,-29.7) -- cycle;

        \foreach \x in {12.0,12.75,...,20.25} {
          \foreach \y in {-2.1,-2.85,...,-24.9} {
            \fill[white,opacity=0.05] (\x,\y) circle (0.028);
          }
        }

        \fill[white,opacity=0.09] (0.95,-1.0) -- (12.65,-1.0) -- (11.25,-13.95) -- (0.95,-15.1) -- cycle;
        \draw[white!16,line width=0.03cm] (0.95,-1.0) -- (12.65,-1.0) -- (11.25,-13.95) -- (0.95,-15.1) -- cycle;
        \fill[ReportAccent] (0.95,-1.0) rectangle (4.15,-1.32);
        \fill[ReportTeal] (4.3,-1.0) rectangle (6.85,-1.32);
        \draw[white!18,line width=0.04cm] (1.35,-12.1) -- (11.0,-13.35);

        \node[
          anchor=north west,
          inner sep=0pt,
          text width=10.8cm,
          align=left,
          text=white,
          font=\CoverTitleFont\fontsize{39}{41}\selectfont
        ] at (1.45,-2.65) {\raggedright\hyphenpenalty=10000\exhyphenpenalty=10000\ReportTitle\par};
        \node[
          anchor=north west,
          inner sep=0pt,
          text width=9.3cm,
          align=left,
          text=ReportCard,
          font=\CoverSubtitleFont\fontsize{18}{21}\selectfont
        ] at (1.45,-7.7) {\ReportSubtitle};

        \draw[white!14,line width=0.06cm] (15.85,-14.15) circle (4.35);
        \draw[white!16,dashed,line width=0.03cm,dash pattern=on 0.18cm off 0.18cm] (15.9,-14.15) ellipse (5.25cm and 2.95cm);

        \filldraw[fill=ReportBlue!22,draw=white!26,line width=0.04cm,opacity=0.5]
          (13.85,-11.5) -- (16.75,-10.35) -- (19.1,-12.35) -- (16.25,-13.5) -- cycle;
        \filldraw[fill=ReportTeal!18,draw=white!22,line width=0.04cm,opacity=0.55]
          (13.15,-14.15) -- (16.0,-13.0) -- (18.35,-15.0) -- (15.5,-16.15) -- cycle;
        \filldraw[fill=ReportAccent!17,draw=white!18,line width=0.03cm,opacity=0.42]
          (16.0,-13.0) -- (18.35,-15.0) -- (19.1,-12.35) -- (16.75,-10.35) -- cycle;
        \filldraw[fill=white,draw=white!14,line width=0.03cm,opacity=0.08]
          (13.15,-14.15) -- (15.5,-16.15) -- (16.25,-13.5) -- (13.85,-11.5) -- cycle;

        \filldraw[fill=ReportAccent!75,draw=white!14,line width=0.02cm,opacity=0.26]
          (14.2,-12.45) -- (16.65,-11.45) -- (18.45,-12.9) -- (16.0,-13.9) -- cycle;
        \filldraw[fill=ReportSky,draw=white!14,line width=0.02cm,opacity=0.22]
          (15.4,-11.2) -- (17.35,-12.0) -- (17.35,-15.5) -- (15.4,-14.7) -- cycle;
        \filldraw[fill=ReportMint,draw=white!14,line width=0.02cm,opacity=0.22]
          (14.1,-13.65) -- (16.5,-12.7) -- (18.25,-14.2) -- (15.85,-15.15) -- cycle;

        \draw[white!22,line width=0.03cm] (13.15,-14.15) -- (13.85,-11.5);
        \draw[white!22,line width=0.03cm] (16.0,-13.0) -- (16.75,-10.35);
        \draw[white!22,line width=0.03cm] (18.35,-15.0) -- (19.1,-12.35);
        \draw[white!22,line width=0.03cm] (15.5,-16.15) -- (16.25,-13.5);

        \fill[ReportAccent] (15.0,-12.35) circle (0.075);
        \fill[ReportSky] (16.1,-11.95) circle (0.075);
        \fill[ReportMint] (17.35,-12.8) circle (0.075);
        \fill[white] (15.85,-13.95) circle (0.075);
        \fill[ReportSky] (16.9,-14.45) circle (0.075);
        \fill[ReportMint] (14.9,-14.55) circle (0.075);
        \fill[ReportAccent] (17.85,-13.55) circle (0.075);
        \fill[white] (15.45,-15.2) circle (0.075);

        \draw[ReportSky!85,opacity=0.8,line width=0.03cm] (12.75,-9.95) -- (14.3,-12.45);
        \draw[ReportSky!85,opacity=0.8,line width=0.03cm] (12.75,-9.95) -- (14.9,-10.95);
        \draw[ReportSky!85,opacity=0.8,line width=0.03cm] (12.1,-10.15) rectangle (12.5,-9.75);
        \draw[ReportSky!85,opacity=0.8,line width=0.03cm] (12.5,-10.05) -- (12.78,-9.9) -- (12.78,-9.8) -- (12.5,-9.65);

        \draw[ReportMint!85,opacity=0.8,line width=0.03cm] (12.9,-17.75) -- (14.65,-15.2);
        \draw[ReportMint!85,opacity=0.8,line width=0.03cm] (12.9,-17.75) -- (15.2,-16.0);
        \draw[ReportMint!85,opacity=0.8,line width=0.03cm] (12.2,-17.95) rectangle (12.6,-17.55);
        \draw[ReportMint!85,opacity=0.8,line width=0.03cm] (12.6,-17.85) -- (12.88,-17.7) -- (12.88,-17.6) -- (12.6,-17.45);

        \draw[ReportAccent!88,opacity=0.82,line width=0.03cm] (19.45,-10.85) -- (17.8,-12.35);
        \draw[ReportAccent!88,opacity=0.82,line width=0.03cm] (19.45,-10.85) -- (17.4,-10.95);
        \draw[ReportAccent!88,opacity=0.82,line width=0.03cm] (19.5,-11.05) rectangle (19.9,-10.65);
        \draw[ReportAccent!88,opacity=0.82,line width=0.03cm] (19.5,-10.95) -- (19.22,-10.8) -- (19.22,-10.7) -- (19.5,-10.55);

        \draw[white!72,opacity=0.72,line width=0.03cm] (18.95,-18.75) -- (17.2,-15.45);
        \draw[white!72,opacity=0.72,line width=0.03cm] (18.95,-18.75) -- (16.45,-15.9);
        \draw[white!72,opacity=0.72,line width=0.03cm] (19.0,-18.95) rectangle (19.4,-18.55);
        \draw[white!72,opacity=0.72,line width=0.03cm] (19.0,-18.85) -- (18.72,-18.7) -- (18.72,-18.6) -- (19.0,-18.45);

        \draw[white!18,line width=0.025cm] (13.0,-9.2) .. controls (14.0,-8.35) and (16.1,-8.25) .. (18.2,-9.5);
        \draw[white!18,line width=0.025cm] (13.35,-19.2) .. controls (15.2,-20.2) and (17.25,-20.0) .. (19.2,-18.8);

        \fill[ReportMist!82!white,opacity=0.92] (0,-26.05) rectangle (21,-29.7);
        \draw[white!38,line width=0.03cm] (0,-26.05) -- (21,-26.05);
        \draw[ReportBlue!20!white,line width=0.02cm] (0,-26.45) -- (21,-26.45);
        \node[
          anchor=north west,
          inner sep=0pt,
          text width=19.1cm
        ] at (0.95,-26.95) {%
          \makebox[19.1cm][c]{%
            \coverlogo{2.85cm}{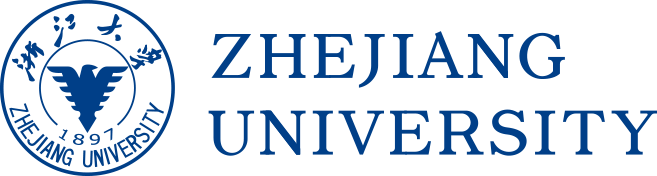}%
            \hspace{0.40cm}%
            \coverlogo{2.85cm}{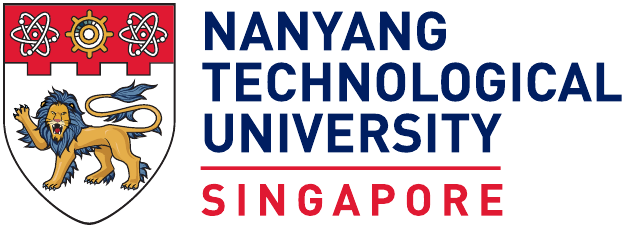}%
            \hspace{0.40cm}%
            \coverlogo{2.85cm}{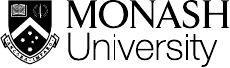}%
            \hspace{0.40cm}%
            \coverlogo{2.85cm}{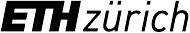}%
            \hspace{0.40cm}%
            \coverlogo{2.85cm}{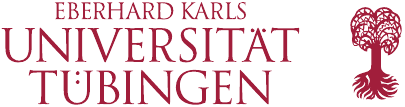}%
            \hspace{0.40cm}%
            \coverlogo{2.85cm}{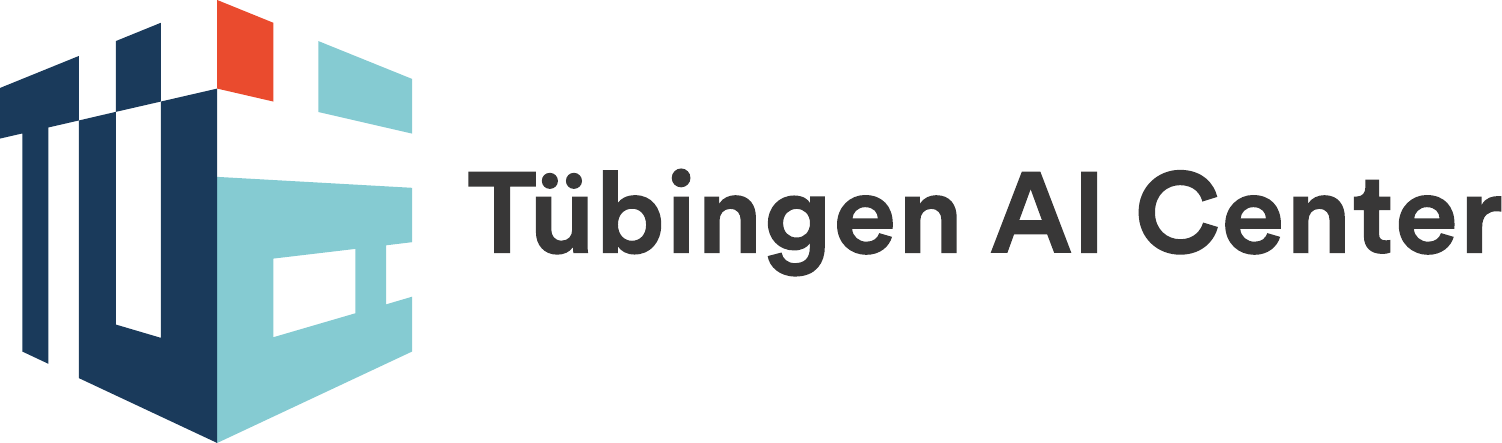}%
          }%
        };
      \end{scope}
    \end{tikzpicture}
    \null
  \end{titlepage}
}

\newcommand{\makefrontmatterpage}{%
  \clearpage
  \thispagestyle{plain}
  \section*{Abstract}
  \addcontentsline{toc}{section}{Abstract}
  \ReportAbstract

  \vspace{1.3em}
  {\sffamily\bfseries\color{ReportNavy} Keywords}\par
  \vspace{0.35em}
  \ReportKeywords

  \vspace{1.4em}
  {\sffamily\bfseries\color{ReportNavy} Authors}\par
  \vspace{0.35em}
  {\small \ReportAuthors\par}

  \vspace{1.1em}
  {\sffamily\bfseries\color{ReportNavy} Affiliations}\par
  \vspace{0.35em}
  {\small \ReportAffiliations\par}

  \vspace{1.1em}
  {\sffamily\bfseries\color{ReportNavy} Corresponding Authors}\par
  \vspace{0.35em}
  {\small \CorrespondingAuthors\par}
}

\newcommand{\makecontentspage}{%
  \clearpage
  \thispagestyle{plain}
  \pdfbookmark[1]{Contents}{contents}
  \tableofcontents
  \clearpage
}

\newcommand{\ReportType}{Survey Technical Report}
\newcommand{\ReportTitle}{Feed-Forward 3D Scene Modeling}
\newcommand{\ReportSubtitle}{A Problem-Driven Perspective}

\newcommand{\EqualContributionSymbol}{\ensuremath{\dagger}}
\newcommand{\ProjectLeadSymbol}{\ensuremath{\blacklozenge}}
\newcommand{\CorrespondenceSymbol}{\Letter}
\newcommand{\AuthorSup}[1]{\textsuperscript{#1}}
\newcommand{\AuthorOrcid}[1]{%
  \if\relax\detokenize{#1}\relax\else\,\orcidlink{#1}\fi
}
\newcommand{\AuthorName}[3]{\textbf{#1}\AuthorOrcid{#2}\AuthorSup{#3}}
\newcommand{\WeijieWangOrcid}{0009-0006-9088-1471}
\newcommand{\QihangCaoOrcid}{0009-0009-7554-724X}
\newcommand{\SensenGaoOrcid}{0009-0009-2282-491X}
\newcommand{\DonnyYChenOrcid}{0000-0003-0943-1512}
\newcommand{\HaofeiXuOrcid}{0000-0003-1313-3358}
\newcommand{\WenjingBianOrcid}{0000-0002-6672-3450}
\newcommand{\SongyouPengOrcid}{0009-0007-6085-8059}
\newcommand{\TatJenChamOrcid}{0000-0001-5264-2572}
\newcommand{\ChuanxiaZhengOrcid}{0000-0002-3584-9640}
\newcommand{\AndreasGeigerOrcid}{0000-0002-8151-3726}
\newcommand{\JianfeiCaiOrcid}{0000-0002-9444-3763}
\newcommand{\JiaWangBianOrcid}{0000-0003-2046-3363}
\newcommand{\BohanZhuangOrcid}{0000-0002-0074-0303}
\newcommand{\CorrespondenceOne}{jiawang.bian@gmail.com}
\newcommand{\CorrespondenceTwo}{bohan.zhuang@zju.edu.cn}
\newcommand{\CorrespondingAuthors}{%
  \AuthorName{Jia-Wang Bian}{\JiaWangBianOrcid}{\CorrespondenceSymbol}\ (\href{mailto:\CorrespondenceOne}{\CorrespondenceOne})\\
  \AuthorName{Bohan Zhuang}{\BohanZhuangOrcid}{\CorrespondenceSymbol}\ (\href{mailto:\CorrespondenceTwo}{\CorrespondenceTwo})
}
\newcommand{\ReportKeywords}{Feed-forward 3D, 3D Representations, 3DGS, NeRF, Pointmaps, Survey}
\newcommand{\ReportAuthors}{%
  \AuthorName{Weijie Wang}{\WeijieWangOrcid}{1,\EqualContributionSymbol},\ 
  \AuthorName{Qihang Cao}{\QihangCaoOrcid}{2,\EqualContributionSymbol},\ 
  \AuthorName{Sensen Gao}{\SensenGaoOrcid}{2,\EqualContributionSymbol},\ 
  \AuthorName{Donny Y. Chen}{\DonnyYChenOrcid}{3,\ProjectLeadSymbol},\\[0.35em]
  \AuthorName{Haofei Xu}{\HaofeiXuOrcid}{4,5},\ 
  \AuthorName{Wenjing Bian}{\WenjingBianOrcid}{5},\ 
  \AuthorName{Songyou Peng}{\SongyouPengOrcid}{4},\ 
  \AuthorName{Tat-Jen Cham}{\TatJenChamOrcid}{2},\ 
  \AuthorName{Chuanxia Zheng}{\ChuanxiaZhengOrcid}{2},\\[0.35em]
  \AuthorName{Andreas Geiger}{\AndreasGeigerOrcid}{5},\ 
  \AuthorName{Jianfei Cai}{\JianfeiCaiOrcid}{3},\ 
  \AuthorName{Jia-Wang Bian}{\JiaWangBianOrcid}{2,\CorrespondenceSymbol},\ 
  \AuthorName{Bohan Zhuang}{\BohanZhuangOrcid}{1,\CorrespondenceSymbol}
}
\newcommand{\ReportAffiliations}{%
  $^{1}$ Zhejiang University, China\\
  $^{2}$ Nanyang Technological University, Singapore\\
  $^{3}$ Monash University, Australia\\
  $^{4}$ ETH Zurich, Switzerland\\
  $^{5}$ University of T\"ubingen, T\"ubingen AI Center, Germany\\[0.35em]
  \AuthorSup{\EqualContributionSymbol}\ Equal contribution\\
  \AuthorSup{\CorrespondenceSymbol}\ Corresponding authors\\
  \AuthorSup{\ProjectLeadSymbol}\ Project lead
}
\newcommand{\ReportAbstract}{%
Reconstructing 3D representations from 2D inputs is a fundamental task in computer vision and graphics, serving as a cornerstone for understanding and interacting with the physical world. While traditional methods achieve high fidelity, they are limited by slow per-scene optimization or category-specific training, which hinders their practical deployment and scalability. Hence, \textbf{generalizable feed-forward 3D reconstruction} has witnessed rapid development in recent years. By learning a model that maps images directly to 3D representations in a single forward pass, these methods enable efficient reconstruction and robust cross-scene generalization. Our survey is motivated by a critical observation: despite the diverse geometric output representations, ranging from implicit fields to explicit primitives, existing feed-forward approaches share similar high-level architectural patterns, such as image feature extraction backbones, multi-view information fusion mechanisms, and geometry-aware design principles. Consequently, we abstract away from these representation differences and instead focus on model design, proposing \textbf{a novel taxonomy centered on model design strategies that are agnostic to the output format}. Our proposed taxonomy organizes the research directions into five key problems that drive recent research development: 1) feature enhancement for robust 2D-to-3D lifting; 2) geometry awareness to incorporate priors for sparse inputs; 3) model efficiency to reduce computation and memory; 4) augmentation strategies leveraging generative models; and 5) temporal-aware models for dynamic 4D reconstruction. To support this taxonomy with empirical grounding and standardized evaluation, we further comprehensively review related benchmarks and datasets, and extensively discuss and categorize real-world applications based on feed-forward 3D models. Finally, we outline future directions to address open challenges such as scalability, evaluation standards, and world modeling. More can be found on our \href{https://github.com/ziplab/Awesome-Feed-Forward-3D}{GitHub repository} and \href{https://ff3d-survey.github.io}{project page}.
}

\hypersetup{
  pdftitle={\ReportTitle: \ReportSubtitle},
  pdfauthor={Weijie Wang et al.},
  pdfsubject={\ReportType},
  pdfkeywords={\ReportKeywords}
}

\begin{document}

\makecoverpage
\makefrontmatterpage
\makecontentspage

\section{Introduction}\label{introduction}
\begin{figure*}[htbp]
  \centering
  \includegraphics[width=\textwidth]{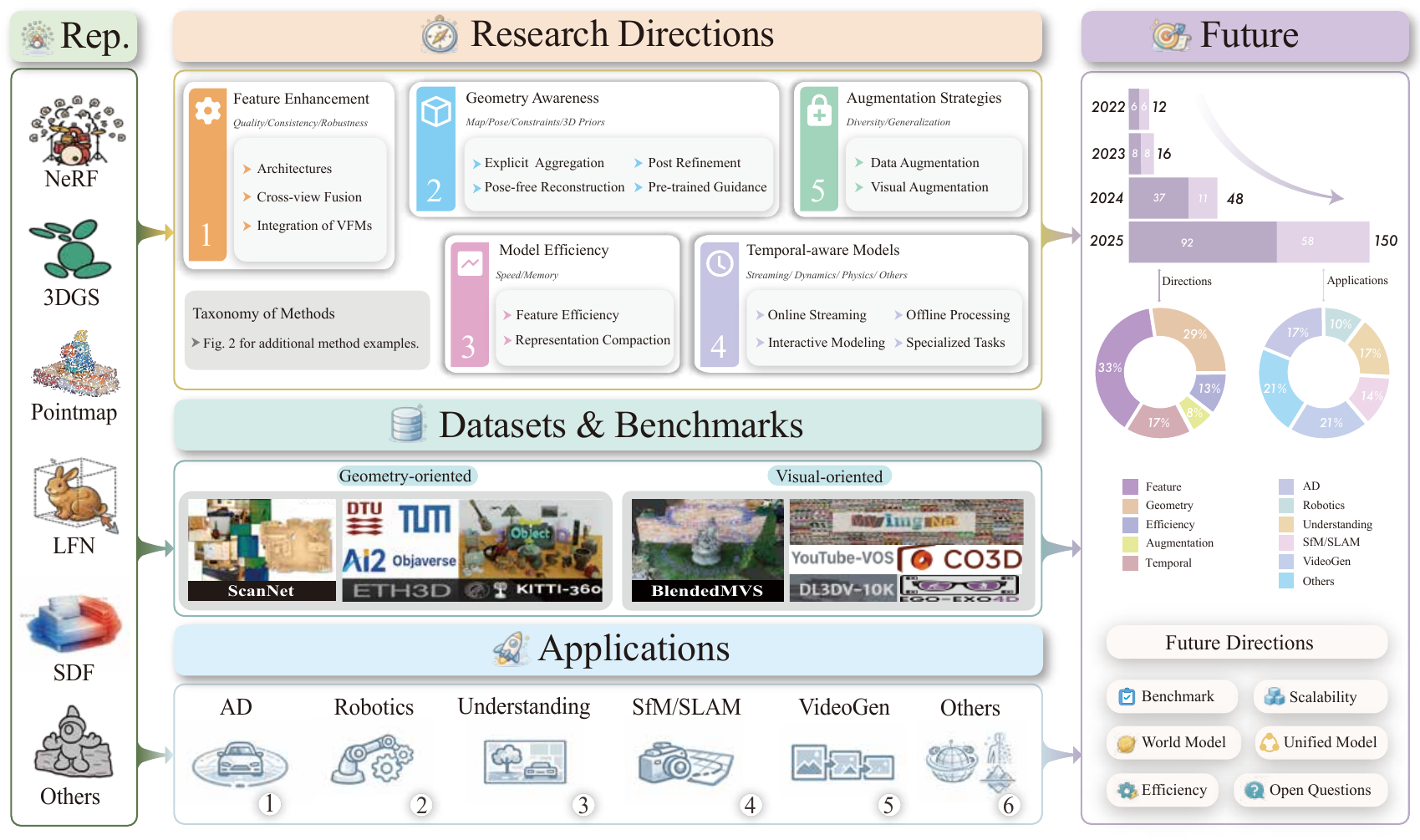}
  \caption{\textbf{Outline of the survey.} 
  The paper begins with an overview of core 3D representations (NeRF~\citep{mildenhall2020nerf}, 3D Gaussian Splatting~\citep{kerbl20233d}, pointmaps~\citep{wang2024dust3r}, and others~\citep{park2019deepsdf,mescheder2019occupancy,oechsle2019texture,oechsle2020learning,niemeyer2020differentiable}) and how feed-forward networks generate them.
  It then analyzes methods along five challenge-driven axes: feature enhancement, geometry awareness, model efficiency, data or visual augmentation, and temporal modeling, followed by reviews of datasets and benchmarks, and practical applications (\textit{e.g.}, autonomous driving and robotics). Finally, based on a statistical summary of the surveyed works, we outline recent trends and discuss future directions for feed-forward 3D scene modeling.
  }
  \label{fig:paper_trends}
\end{figure*}

Modeling 3D scenes from images or videos, including their geometry, appearance, motion, and interactions, is a fundamental problem in computer vision, with broad applications in robotics, AR/VR, digital heritage, and autonomous systems.
Classical methods,
such as Structure-from-Motion (SfM)~\citep{longuet1981computer},
Multi-View Stereo (MVS)~\citep{furukawa2015multi},
and 
more recent Neural Radiance Fields (NeRF)~\citep{mildenhall2020nerf} and 3D Gaussian Splatting (3DGS)~\citep{kerbl20233d},
have laid important groundwork, progressively advancing both geometric fidelity and photorealism.
However, these methods often rely on per-scene optimization with substantial computational costs, which limits their scalability and real-time use.
This creates a pressing need for more efficient and generalizable paradigms.

Feed-forward 3D modeling~\citep{wu2016learning,fan2017point} has recently emerged as an alternative paradigm in the field.
Instead of optimizing a scene representation at test time, feed-forward methods learn a mapping from input images, optionally incorporating auxiliary signals such as camera poses or depth priors~\citep{yu2021pixelnerf,sitzmann2021light,charatan2024pixelsplat,szymanowicz24splatter,chen2024mvsplat,xu2024depthsplat,wang2024dust3r,wang2025vggt}, to an explicit or implicit 3D representation within a single forward pass.
This design enables significantly faster inference, improved amortization across scenes, and seamless integration into end-to-end pipelines for downstream tasks.
However, this paradigm also presents new technical challenges, including multi-view feature fusion, preservation of geometric details, model efficiency, and temporal coherence for dynamic scenes.
Addressing these challenges continues to drive rapid progress and methodological innovation.



Motivated by the rapid development of feed-forward 3D reconstruction, this survey aims to systematically synthesize recent advances, particularly to clarify the key challenges.
\citet{zhang2025advances} provide a complementary representation-driven review spanning NeRF, pointmaps, 3DGS, other 3D representations, and 3D-free models, together with task, application, and evaluation discussions.
Alongside this representation-driven organization, our work adopts a problem-driven perspective and offers a structured, end-to-end panorama of the field, as illustrated in~\cref{fig:paper_trends}.
Our survey spans representations, five key research directions (with a detailed taxonomy in~\cref{fig:ff_3d_reconstruction_taxonomy}), datasets and benchmarks, a wide range of real-world applications, and future directions.

To organize the field in a way that reflects both technical progress and underlying challenges, we do not simply categorize prior work by its output 3D representations.
While representations,
such as meshes, SDFs, NeRF, 3DGS, and pointmaps,
provide a useful descriptive layer, they often obscure the distinct functional goals and design motivations that drive recent feed-forward methods.
In practice, methods built upon the same representation can target fundamentally different problems, such as feature robustness, geometric ambiguity, or computational efficiency, while approaches addressing similar challenges may adopt diverse representations.
Therefore, after briefly reviewing the commonly used representations in feed-forward 3D reconstruction, we adopt a problem-driven taxonomy that categorizes methods according to the core challenges they aim to address (see~\cref{fig:paper_trends} and~\cref{fig:ff_3d_reconstruction_taxonomy}).
Specifically, we identify five key research directions: (1) Feature Enhancement (\S\ref{sec:feature_enhance}), which seeks to improve the quality of implicit feature representations for more accurate decoding of 3D scenes; (2) Geometry Awareness (\S\ref{sec:geo_aware}), which targets robust and accurate inference of underlying scene geometry; (3) Model Efficiency (\S\ref{sec:model_efficiency}), which addresses computational and memory bottlenecks for real-time and resource-limited settings; (4) Augmentation Strategies (\S\ref{sec:data_visual_refine}), which enrich data distributions and visual representations to overcome sparse inputs and limited training diversity; and (5) Temporal-aware Models (\S\ref{sec:temporal_aware}), which capture geometry and motion consistency across frames for low-latency 4D scene modeling.
This perspective better captures the functional roles, design trade-offs, and developmental trends of existing approaches, and provides a clearer roadmap for understanding current progress and future directions in feed-forward 3D reconstruction.

Hence, we also re-evaluate datasets and benchmarks.
Moving beyond traditional dataset enumeration, we categorize datasets based on their core focus areas: geometry-oriented datasets (\textit{e.g.}, DTU~\citep{jensen2014large}, ScanNet~\citep{dai2017scannet}, Replica~\citep{straub2019replica}) that focus on point clouds, depth, and pose, and visual-oriented datasets (\textit{e.g.}, NeRF-Synthetic~\citep{mildenhall2020nerf}, RealEstate10K~\citep{zhou2018stereo}, DL3DV~\citep{ling2024dl3dv}) that prioritize photorealistic view synthesis.

To further illustrate how benchmark selection impacts research progress, we systematically compile the reported performance of representative methods across key datasets, revealing significant trends across different categories.
We derive several important data-driven takeaways, such as the need to establish standardized quantification methods for scene complexity and to report geometric diversity more clearly~\citep{zhou2018stereo,ling2024dl3dv}. These findings are explored in depth in our discussion of future directions (\S\ref{future}).

After summarizing methods and benchmarks, we turn to the expanding impact of feed-forward 3D reconstruction in real-world applications. This paradigm has evolved from a research concept into a practical technology, now driving progress in domains including autonomous driving (\S\ref{sec:autonomous_driving}), robotics (\S\ref{sec:robotics}), scene understanding (\S\ref{sec:perception}), SfM and SLAM (\S\ref{sec:slam}), video generation (\S\ref{sec:video_gen}), and other scenarios such as visual localization (\S\ref{sec:other_app}).
Together, these applications demonstrate how feed-forward 3D reconstruction is advancing fundamental tasks in computer vision, achieving unprecedented efficiency and significantly lowering the barrier to practical deployment.

While remarkable progress has been achieved, feed-forward 3D reconstruction remains an active frontier with many open challenges and opportunities ahead. We conclude by outlining several promising future directions, spanning benchmark rigor (\S\ref{sec:benchmark}), model efficiency (\S\ref{sec:efficiency}), scalable scene representations (\S\ref{sec:representation}), 
world models (\S\ref{sec:world_model}), unified perception and reconstruction (\S\ref{sec:understand}), and other critical open questions (\S\ref{sec:openq}).
All references and resources surveyed in this work are curated and continuously updated at \rurl{ff3d-survey.github.io}.

\section{Problem Formulation}\label{sec:preliminary}

The goal of generalizable feed-forward 3D models is to reconstruct a 3D scene from a set of input images in a single forward pass, without any per-scene optimization.
Existing approaches generally consist of an \emph{Encoder},
a \emph{Decoder}, and an optional \emph{Renderer} for novel view synthesis.
The encoder $\Phi_\mathrm{image}$ encodes the input images into implicit features, and the decoder $\Psi_\mathrm{pred}$ decodes the implicit features into 3D scene representations. If required, a renderer $\mathcal{R}$ then synthesizes images from the predicted 3D scene.

Given $K$ input images $\mathcal{I}=\{{\mathbf I}_{i}\}_{i=1}^K$ where ${\mathbf I}_i \in \mathbb{R}^{H \times W \times 3}$ and (optionally) their corresponding camera poses $\mathcal{P}^{\ast}=\{{\mathbf P}^{\ast}_{i}\}_{i=1}^K$,  the encoder extracts a set of implicit feature maps $\mathcal{F}=\{{\mathbf F}_{i}\}_{i=1}^K$:
\begin{align}
    \label{eq:feature_extraction}
    \mathcal{F}=\Phi_\mathrm{image}(\mathcal{I}, \mathcal{P}^{\ast}), 
    \quad \mathbf{F}_i \in\mathbb{R}^{\frac{H}{s}\times \frac{W}{s}\times C},
\end{align}
where $s$ is the spatial downsampling factor and $C$ is the feature dimension. In practice, these features can be further refined by additional feature enhancement modules, yielding enhanced feature maps $\mathcal{F}'$.
The enhanced implicit feature maps are then fed into the decoder to produce the 3D scene representations $\mathcal{G}$:

\begin{align}
\label{eq:decode}
    \mathcal{G} = \Psi_\mathrm{pred}(\mathcal{F}', \mathcal{P}^{\ast}). 
\end{align}
where $\Psi_\mathrm{pred}$ is the decoder that decodes the implicit feature maps into 3D scene representations $\mathcal{G}$. For example, when using 3D Gaussian Splatting, $\mathcal{G}$ is a set of Gaussian primitives,
\begin{align}
    \mathcal{G} = \{(\bm{\mu}_{j}, \bm{\Sigma}_{j}, \alpha_{j}, \bm{c}_{j})\}_{j=1}^{N},
\end{align}
where $\bm{\mu}_{j}\!\in\!\mathbb{R}^{3}$ is the mean position vector, $\bm{\Sigma}_{j}\!\in\!\mathbb{R}^{3\times 3}$ the covariance matrix, $\alpha_{j}\!\in\!\mathbb{R}$ the opacity scalar, and $\mathbf{c}_{j}\!\in\!\mathbb{R}^{3}$ the color vector of the $j$-th Gaussian, and $N$ is the total number of primitives.


\noindent\textbf{Training on large-scale data.}
Feed-forward 3D approaches do \emph{not} learn a separate model for each scene.
Instead, they are trained once on a large collection of scenes and then reused at test time.
Let $\mathcal{D}=\{(\mathcal{I}, \mathcal{P})\}$ denote a dataset of training scenes.
For each scene, the encoder and decoder produce a 3D representation $\mathcal{G}$,
such as depth maps, voxels, SDFs, meshes, pointmaps, NeRF, or 3DGS, depending on the specific method.
If the model includes a renderer, it takes $\mathcal{G}$ and novel camera poses $\mathcal{P}_\mathrm{novel}$ as input,
and the renderer $\mathcal{R}$ synthesizes images
\begin{align}
    \widehat{\mathcal{I}} = \mathcal{R}(\mathcal{G},\, \mathcal{P}_\mathrm{train/novel}).
\end{align}
The parameters of $\Phi_\mathrm{image}$, $\Psi_\mathrm{pred}$ and, if learnable, $\mathcal{R}$ are optimized jointly by minimizing a weighted sum of loss terms:
\begin{align}
    \mathcal{L}
    = \sum_{t\in\mathcal{T}} 
      \lambda_t\,\mathcal{L}_t(\mathcal{G},\,\mathcal{G}^{\ast},\,\widehat{\mathcal{I}},\,\mathcal{I},\,\mathcal{P}),
\end{align}
where $\mathcal{T}$ is the set of loss terms, $\lambda_t$ is the weight of $\mathcal{L}_t$, and $\mathcal{G}^{\ast}$ denotes ground-truth geometric annotations (\textit{e.g.}, depth maps, pointmaps, or normals) when available. Depending on the 3D representation and supervision signals, different methods activate different subsets of losses. In practice, $\mathcal{L}$ typically combines:
(i) geometric supervision losses that directly constrain $\mathcal{G}$ against $\mathcal{G}^{\ast}$, including pointmap regression~\citep{wang2024dust3r,leroy2024grounding}, depth supervision~\citep{xu2024depthsplat}, and normal consistency~\citep{chang2025meshsplat,shi2025pmloss}. Methods that directly regress geometry without a renderer (\textit{e.g.}, pointmap-based approaches) rely primarily on this term;
(ii) photometric or perceptual losses between rendered and ground-truth images~\citep{yu2021pixelnerf,charatan2024pixelsplat,chen2024mvsplat}, applicable when a differentiable renderer $\mathcal{R}$ is employed;
and (iii) regularization on the structure of $\mathcal{G}$, such as opacity sparsity and scale constraints for Gaussian splats~\citep{kerbl20233d,fan2024lightgaussian,lee2024compact}, or distortion losses~\citep{barron2022mipnerf360} and depth smoothness terms~\citep{niemeyer2022regnerf} for neural fields.
This large-scale, multi-scene training is essential for feed-forward models, as it amortizes reconstruction over the dataset so that a single set of weights can generalize to unseen scenes.

\noindent\textbf{Feed-forward inference.}
At inference time, feed-forward systems reconstruct a new scene in a \emph{single forward pass} through \cref{eq:feature_extraction,eq:decode} for a novel input $(\mathcal{I}, \mathcal{P})$, without any per-scene gradient-based optimization or test-time fine-tuning.
Some lightweight post-processing, such as confidence-based pruning and merging~\citep{ziwen2024long}, can be applied to further reduce memory consumption and improve rendering stability.
This defines the strict feed-forward setting. We also cover related extensions with recurrent prediction, gradient-guided scene refinement, and sequence-specific adaptation~\citep{xu2025resplat,chen2024g3r,chen2025ttt3r,wang2026tttlrm}; \S\ref{sec:visual_refine} distinguishes their test-time updates and computational requirements.

\section{Representations}\label{representation}

The choice of 3D scene representation is critical to the capabilities and performance of any 3D reconstruction system. The field has seen a progression from traditional explicit geometric models to neural rendering-based representations. In this section, we explore several prominent representations, including NeRF, 3DGS, and pointmaps, examining how feed-forward approaches improve these representations to better meet the demands of modern 3D reconstruction tasks.



\subsection{NeRF}
Neural Radiance Fields (NeRF)~\citep{mildenhall2020nerf} are neural rendering-based representations that model a 3D scene as a continuous function of position and viewing direction. The core mechanism involves a multi-layer perceptron (MLP) that maps a 5D coordinate (comprising a 3D spatial position $\mathbf{x} = (x, y, z)$ and a 2D viewing direction $\mathbf{d} = (\theta, \phi)$) to an RGB color ($\mathbf{c}$) and a volume density ($\sigma$):

\begin{align}
    MLP(\mathbf{x}, \mathbf{d}) = (\mathbf{c}, \sigma).
\end{align}


To render an image from a novel viewpoint, camera rays $\mathbf{r}(t) = \mathbf{o} + t\mathbf{d}$ with camera center $\mathbf{o}$ and viewing direction $\mathbf{d}$ are cast through each pixel of the virtual camera. Points are sampled along each ray, and for each sampled point, the MLP is queried to obtain its color and density. 
These values are then composited along the ray via differentiable volume rendering~\citep{max2002optical,mildenhall2020nerf,tagliasacchi2022volume} to determine the final pixel color,
\begin{align}
    \mathbf{C} = \sum_{i=1}^{K} T_i\,(1-\exp(-\sigma_i\delta_i))\,\mathbf{c}_i, \\ T_i = \exp\!\Bigl(-\sum_{j=1}^{i-1}\sigma_j\delta_j\Bigr),
\end{align}
where $\mathbf{c}_i$ and $\sigma_i$ are the color and density of the $i$-th sampled point on the ray, $T_i$ is the accumulated transmittance, $\delta_i$ is the distance between adjacent samples, and $K$ is the total number of sampled points.

The strengths of NeRF lie in its ability to produce high-fidelity, photorealistic novel views, effectively handling complex occlusions and view-dependent effects such as specular reflections.
However, the original NeRF framework has several limitations, such as relying on lengthy per-scene optimization, requiring dense MLP queries for rendering, and necessitating extensive calibration of viewpoints.



Follow-up works substantially improve the quality and robustness across challenging scenarios. For example, Mip-NeRF~\citep{barron2021mipnerf} and Mip-NeRF 360~\citep{barron2022mipnerf360} address aliasing artifacts and extend NeRF's reach to unbounded scenes through multi-scale representations and advanced regularization. Ref-NeRF~\citep{verbin2022refnerf} adapts the volumetric formulation to better capture reflective surfaces. DS-NeRF~\citep{deng2022depth} and NerfingMVS~\citep{wei2021nerfingmvs} incorporate geometric priors to accelerate convergence and improve scene accuracy, even with sparser supervision. On the efficiency front, InstantNGP~\citep{muller2022instant} leverages hash-based encodings to dramatically speed up training and inference.


Despite these advances, most variants still rely on per-scene optimization. Feed-forward NeRF-style reconstruction aims to generalize across scenes without test-time retraining.
pixelNeRF~\citep{yu2021pixelnerf} is an early example that conditions the radiance field on image features extracted from one or more input views.
An image encoder processes the input image(s) to produce a feature volume. When rendering a ray, features are sampled at the projected locations of the 3D query points and fed into the MLP along with the position and viewing direction. This allows the network to predict the radiance field in a single forward pass, conditioned on the visual input.

\subsection{3D Gaussian Splatting}

3D Gaussian Splatting (3DGS)~\citep{kerbl20233d} is an explicit, primitive-based representation that has significantly advanced real-time neural rendering. 
A scene is modeled as a set of anisotropic 3D Gaussians, where each Gaussian is parameterized by $(\mathbf{\mu}, \mathbf{\Sigma}, \alpha, \mathbf{c})$, where $\mathbf{\mu} \in \mathbb{R}^3$ is the 3D position, $\mathbf{\Sigma} \in \mathbb{R}^{3 \times 3}$ is the covariance matrix, $\alpha \in [0, 1]$ is the opacity value, and $\mathbf{c}$ is the view-dependent color. Rendering takes place through a differentiable, visibility-aware splatting process that projects Gaussians onto the target image plane, and then combines their contributions based on opacity as they overlap in screen space. A differentiable forward rasterizer effectively computes visibility and compositing. This method greatly improves efficiency and scalability for real-time rendering.


3DGS offers notable advantages in comparison with NeRF. Training typically converges within minutes rather than hours, and rendering is achieved in real time. The explicit primitive-based representation produces high-fidelity results suited for interactive applications such as AR and VR. However, these strengths come with trade-offs. Standard 3DGS pipelines are highly dependent on precise camera positions and SfM-derived sparse point clouds for initialization. Furthermore, achieving high-quality reconstructions often requires millions of Gaussians, leading to substantial memory usage and bandwidth demands, sometimes even exceeding those of NeRF MLPs.

To tackle these issues, recent research has taken up the challenge in two main ways. First, a series of quality-oriented methods, like FreGS~\citep{zhang2024fregs}, Scaffold-GS~\citep{lu2024scaffold}, and Gaussian Opacity Fields~\citep{yu2024gaussian}, introduce new regularization strategies and advanced shading models to further improve stability, mitigate artifacts, and enhance the overall visual fidelity of rendered scenes. Other specialized techniques, including those for reflection~\citep{jiang2024gaussianshader, yang2024spec, meng2024mirror} and motion blur~\citep{peng2024bags, zhao2024bad}, further expand the expressive power of 3DGS in highly challenging scenarios. Second, efficiency-oriented methods try to solve the problems entailed in the representation and manipulation of 3DGS by making its formal representation more compact. Methods like Compact-3DGS~\citep{lee2024compact}, Reducing-3DGS~\citep{papantonakis2024reducing}, and LightGaussian~\citep{fan2024lightgaussian} directly reduce the number of Gaussian primitives, while works such as Compressed-3DGS~\citep{niedermayr2024compressed} and HAC~\citep{chen2024hac} focus on compressing the attributes of each Gaussian, making the storage and streaming of large-scale scenes far more practical.

Despite these advances, vanilla 3DGS and most extensions still follow a per-scene optimization paradigm initialized from SfM. Feed-forward 3DGS methods aim to bypass this requirement by predicting Gaussian primitives directly from images. For example, pixelSplat~\citep{charatan2024pixelsplat} adopts an encoder--decoder architecture where multi-scale features are extracted from one or more input views and decoded into pixel-aligned 3D Gaussian parameters in a single forward pass. Such approaches move 3DGS toward the feed-forward regime, enabling faster, more generalizable, and more deployment-friendly Gaussian-based 3D reconstruction.

\subsection{Pointmap}

A pointmap represents a scene with 3D points and associated features. In feed-forward methods such as DUSt3R~\citep{wang2024dust3r}, a pointmap is formally defined as a dense 2D field of 3D points, denoted as $X \in \mathbb{R}^{H \times W \times 3}$. When associated with its corresponding RGB image $I$ of resolution $H\times W$, a pointmap establishes a direct, one-to-one mapping between image pixels and 3D scene points. Notably, when the network processes a pair of images $I_1$ and $I_2$, it outputs two corresponding pointmaps $X_{1,1}$ and $X_{2,1}$. Here, $X_{n, m}$ denotes the pointmap $X_{n}$ from camera $n$ expressed in camera $m$'s coordinate frame:
\begin{align}
    X_{n, m} = \mathbf{P}_m \mathbf{P}_n^{-1} h(X_{n}),
\end{align}
where $\mathbf{P}_m$ and $\mathbf{P}_n$ are the world-to-camera poses for cameras $m$ and $n$, respectively. $h: (x, y, z) \rightarrow (x, y, z, 1)$ is the homogeneous mapping.

Pointmaps prove invaluable for visual localization applications. Both scene-specific optimization methods~\citep{brachmann2017dsac, brachmann2018learning, brachmann2021visual} and scene-agnostic inference approaches that generalize across scenes~\citep{revaud2024sacreg,tang2021learning,yang2019sanet} have successfully employed this representation. The concept extends beyond localization: storing 3D geometry through 2D views has become fundamental in both single-image 3D reconstruction~\citep{lin2018learning,shin2018pixels,tatarchenko2016multi} and view synthesis~\citep{wiles2020synsin}. By operating in image space and utilizing perspective camera geometry for rendering, these methods effectively process and manipulate the underlying 3D structure.

Feed-forward reconstruction of pointmaps is exemplified by DUSt3R~\citep{wang2024dust3r}. A transformer-based ViT~\citep{weinzaepfel2022croco} is used to process multiple input images. The model directly regresses a pointmap for each input view in a single forward pass. When processing a pair of images, it simultaneously predicts their relative camera poses and consistent pointmaps, where the 3D points from one view are expressed in the coordinate frame of the other. The core operation is learning to map pixel-aligned image features directly to 3D coordinates, effectively unifying depth estimation and camera pose inference.

\begin{figure*}[!htbp]
\newcommand{\TaxSecondary}{\textsuperscript{$\dagger$}}
    \centering
    \resizebox{\textwidth}{!}{
        \begin{forest}
            leaf1/.style={fill=orange!8!white, draw=black!35},
            leaf2/.style={fill=cyan!8!white,   draw=black!35},
            leaf3/.style={fill=violet!7!white, draw=black!35},
            leaf4/.style={fill=pink!7!white,   draw=black!35},
            leaf5/.style={fill=lime!9!white,   draw=black!35},
            forked edges,
            for tree={
                grow=east,
                reversed=true,
                anchor=base west,
                parent anchor=east,
                child anchor=west,
                base=left,
                font=\small,
                rectangle,
                draw=hidden-draw,
                rounded corners,
                align=left,
                minimum width=4em,
                edge+={darkgray, line width=1pt},
                s sep=3pt,
                inner xsep=2pt,
                inner ysep=3pt,
                ver/.style={rotate=90, child anchor=north, parent anchor=south, anchor=center},
            },
            where level=1{text width=10.0em, font=\scriptsize}{},
            where level=2{text width=10.5em, font=\scriptsize}{},
            [
                \textbf{Research Directions of Feed-Forward 3D Reconstruction}, ver
                [
                    Feature Enhancement \\ ~(\S\ref{sec:feature_enhance})
                    [
                        Architectures ~(\S\ref{sec:encode_arch})
                        [
                            {\parbox[t]{28em}{\hyphenpenalty=10000\exhyphenpenalty=10000\relax
                            \textit{e.g.,}
                            pixelNeRF~\citep{yu2021pixelnerfneuralradiancefields}{,}
                            IBRNet~\citep{wang2021ibrnetlearningmultiviewimagebased}{,}
                            Occupancy Networks~\citep{mescheder2019occupancy}{,}
                            Splatter Image~\citep{szymanowicz24splatter}{,}
                            LRM~\citep{honglrm}{,}
                            LGM~\citep{tang2024lgm}{,}
                            GS-LRM~\citep{zhang2024gs}{,}
                            CATSplat~\citep{roh2024catsplat}{,}
                            Gamba~\citep{shen2025gamba}{,}
                            GNT~\citep{t2023attentionnerfneeds}{,}
                            Long-LRM\TaxSecondary~\citep{ziwen2024long}{,}
                            MeshLRM~\citep{wei2024meshlrm}{,}
                            Light Field Networks~\citep{sitzmann2021light}{,}
                            SRT~\citep{sajjadi2022scenerepresentationtransformergeometryfree}{,}
                            VisionNeRF~\citep{lin2023vision}{,}
                            TripoSR~\citep{tochilkin2024triposr}{,}
                            LVSM~\citep{jin2025lvsm}{,}
                            Depth Anything 3~\citep{depthanything3}{,}
                            Learned Initializations~\citep{tancik2021learnedinitializationsoptimizingcoordinatebased}{,}
                            C3-GS~\citep{hu2025c3}{,}
                            Instant3D~\citep{li2024instant3d}{,}
                            Flex3D~\citep{han2024flex3d}{,}
                            InfiniSplat~\citep{wang2026infinisplat}{,}
                            TokenGS~\citep{ren2026tokengs}{,}
                            UniqueSplat~\citep{song2026uniquesplat}{,}
                            ViewSplat~\citep{jeong2026viewsplat}{,}
                            LangFlash~\citep{liu2026langflash}{,}
                            2Xplat~\citep{jeong20262xplat}{,}
                            MVGamba~\citep{yi2024mvgamba}
                            }}
                            , leaf1, text width=28em
                        ]
                    ]
                    [
                        Cross-View Fusion \\ ~(\S\ref{sec:view_fusion})
                        [
                            {\parbox[t]{28em}{\hyphenpenalty=10000\exhyphenpenalty=10000\relax
                            \textit{e.g.,}
                            MASt3R~\citep{leroy2024grounding}{,}
                            MV-DUSt3R~\citep{tang2024mv}{,}
                            MUSt3R~\citep{cabon2025must3r}{,}
                            Spann3R~\citep{wang20243d}{,}
                            TTT3R~\citep{chen2025ttt3r}{,}
                            tttLRM~\citep{wang2026tttlrm}{,}
                            CUT3R~\citep{wang2025continuous}{,}
                            DUSt3R\TaxSecondary~\citep{wang2024dust3r}{,}
                            3D-R2N2~\citep{choy20163d}{,}
                            Fast3R~\citep{yang2025fast3r}{,}
                            Dens3r~\citep{fang2025dens3r}{,}
                            DAGE~\citep{ngo2026dage}{,}
                            FF3R~\citep{zhou2026ff3r}{,}
                            RobustVGGT\TaxSecondary~\citep{han2025outlierrejection}{,}
                            MVGamba\TaxSecondary~\citep{yi2024mvgamba}
                            }}
                            , leaf1, text width=28em
                        ]
                    ]
                    [
                        Integration of Visual \\ Foundation Models \\ ~(\S\ref{sec:prior_foundation})
                        [
                            {\parbox[t]{28em}{\hyphenpenalty=10000\exhyphenpenalty=10000\relax
                            \textit{e.g.,}
                            Mono3R~\citep{li2025mono3r}{,}
                            VGGT~\citep{wang2025vggt}{,}
                            Feat2GS~\citep{chen2025feat2gs}{,}
                            VGGT-$\Omega$~\citep{wang2026vggtomega}{,}
                            G2VLM~\citep{hu2025g2vlm}{,}
                            Fisheye3R~\citep{duan2026fisheye3r}{,}
                            LangFlash\TaxSecondary~\citep{liu2026langflash}{,}
                            FF3R\TaxSecondary~\citep{zhou2026ff3r}
                            }}
                            , leaf1, text width=28em
                        ]
                    ]
                ]
                [
                    Geometry Awareness \\ ~(\S\ref{sec:geo_aware})
                    [
                        Explicit Geometric \\ Aggregation ~(\S\ref{mvs_methods})
                        [
                            {\parbox[t]{28em}{\hyphenpenalty=10000\exhyphenpenalty=10000\relax
                            \textit{e.g.,}
                            MVSNeRF~\citep{chen2021mvsnerffastgeneralizableradiance}{,}
                            GeoNeRF~\citep{johari2022geonerfgeneralizingnerfgeometry}{,}
                            MuRF~\citep{xu2024murfmultibaselineradiancefields}{,}
                            GTA~\citep{Miyato2024GTA}{,}
                            pixelSplat~\citep{charatan2024pixelsplat}{,}
                            MVSplat~\citep{chen2024mvsplat}{,}
                            MVSGaussian~\citep{liu2025mvsgaussian}{,}
                            MatchNeRF~\citep{chen2023explicit}{,}
                            TranSplat~\citep{zhang2025transplat}{,}
                            MuGS~\citep{lou2025mugs}{,}
                            SparseNeuS~\citep{long2022sparseneus}{,}
                            VolRecon~\citep{ren2023volrecon}{,}
                            BoostMVSNeRFs~\citep{su2024boostmvsnerfs}{,}
                            UFORecon~\citep{na2024uforecon}{,}
                            MAC-Splat~\citep{yang2026macsplat}{,}
                            Splat-SAP~\citep{zhou2025splatsap}{,}
                            VGD~\citep{lin2025vgd}{,}
                            SurfaceSplat~\citep{gao2025surfacesplat}{,}
                            JointSplat\TaxSecondary~\citep{xiao2025jointsplat}{,}
                            GPOcc\TaxSecondary~\citep{zhou2026generalizingvisualge}{,}
                            DGS-LRM\TaxSecondary~\citep{lin2025dgs}{,}
                            VGGT-Motion\TaxSecondary~\citep{xiong2026vggtmotion}{,}
                            PLANA3R\TaxSecondary~\citep{liu2025plana3r}
                            }}
                            , leaf2, text width=28em
                        ]
                    ]
                    [
                        Post Refinement ~(\S\ref{sec:refine_3d})
                        [
                            {\parbox[t]{28em}{\hyphenpenalty=10000\exhyphenpenalty=10000\relax
                            \textit{e.g.,}
                            HiSplat~\citep{tang2024hisplat}{,}
                            PixelGaussian\TaxSecondary~\citep{fei2024pixelgaussian}{,}
                            ReSplat~\citep{xu2025resplat}{,}
                            G3R~\citep{chen2024g3r}{,}
                            GGN\TaxSecondary~\citep{zhang2024gaussian}{,}
                            FreeSplat++\TaxSecondary~\citep{wang2025freesplat++}{,}
                            Trust3R~\citep{zhu2026trustitor}{,}
                            SurfaceSplat\TaxSecondary~\citep{gao2025surfacesplat}{,}
                            FreeSplat~\citep{wang2024freesplat}{,}
                            RobustVGGT~\citep{han2025outlierrejection}
                            }}
                            , leaf2, text width=28em
                        ]
                    ]
                    [
                        Pose-free Reconstruction \\ ~(\S\ref{sec:pose_free})
                        [
                            {\parbox[t]{28em}{\hyphenpenalty=10000\exhyphenpenalty=10000\relax
                            \textit{e.g.,}
                            DUSt3R~\citep{wang2024dust3r}{,}
                            Splatt3R~\citep{smart2024splatt3r}{,}
                            NoPoSplat~\citep{ye2024no}{,}
                            FLARE~\citep{zhang2025flare}{,}
                            Pow3R~\citep{jang2025pow3r}{,}
                            SPFSplat~\citep{huang2025no}{,}
                            AnySplat~\citep{jiang2025anysplat}{,}
                            $\pi^{3}$~\citep{wang2025pi}{,}
                            PF3plat~\citep{hong2024pf3plat}{,}
                            YoNoSplat~\citep{ye2025yonosplat}{,}
                            WorldMirror\TaxSecondary~\citep{liu2025worldmirror}{,}
                            FlexSplat~\citep{sabbaghziarani2026flexsplat}{,}
                            Fisheye3R\TaxSecondary~\citep{duan2026fisheye3r}{,}
                            2Xplat\TaxSecondary~\citep{jeong20262xplat}{,}
                            PLANA3R~\citep{liu2025plana3r}{,}
                            UFO-4D\TaxSecondary~\citep{hur2026ufo4d}
                            }}
                            , leaf2, text width=28em
                        ]
                    ]
                    [
                        Pre-trained Geometric \\ Guidance ~(\S\ref{sec:pretrained_geo})
                        [
                            {\parbox[t]{28em}{\hyphenpenalty=10000\exhyphenpenalty=10000\relax
                            \textit{e.g.,}
                            Flash3D~\citep{szymanowicz2024flash3d}{,}
                            PM-Loss~\citep{shi2025pmloss}{,}
                            DepthSplat~\citep{xu2024depthsplat}{,}
                            Niagara~\citep{wu2025niagara}{,}
                            WorldMirror~\citep{liu2025worldmirror}{,}
                            JointSplat~\citep{xiao2025jointsplat}{,}
                            GPOcc~\citep{zhou2026generalizingvisualge}
                            }}
                            , leaf2, text width=28em
                        ]
                    ]
                ]
                [
                    Model Efficiency ~(\S\ref{sec:model_efficiency})
                    [
                    Feature Efficiency\\(\S\ref{sec:feat_eff})
                    [
                        {\parbox[t]{28em}{\hyphenpenalty=10000\exhyphenpenalty=10000\relax
                        \textit{e.g.,}
                        ENeRF~\citep{lin2022efficient}{,}
                        iLRM~\citep{kang2025ilrm}{,}
                        Long-LRM~\citep{ziwen2024long}{,}
                        ZPressor~\citep{wang2025zpressor}{,}
                        FastVGGT~\citep{shen2025fastvggttrainingfreeaccelerationvisual}{,}
                        LiteVGGT~\citep{shu2025litevggt}{,}
                        ProNeRF~\citep{bello2024pronerf}{,}
                        TinySplat~\citep{song2025tinysplat}{,}
                        HeSS~\citep{kim2026hess}{,}
                        InfiniteVGGT~\citep{yuan2026infinitevggt}{,}
                        StreamSplat (Song et al.)\TaxSecondary~\citep{song2026streamsplat}{,}
                        QuantVGGT~\citep{feng2025quantized}{,}
                        Diversity-aware View Partitioning~\citep{park2026viewpartition}{,}
                        Long-LRM++~\citep{ziwen2025longlrm}{,}
                        MuSASplat~\citep{xu2025musasplat}{,}
                        STAC~\citep{wang2026stac}{,}
                        Z-Order Transformer~\citep{wang2026zordertransformer}{,}
                        LingBot-Map\TaxSecondary~\citep{chen2026lingbotmap}{,}
                        TTSA3R\TaxSecondary~\citep{zheng2026ttsa3r}
                        }}
                        ,leaf3, text width=28em]],[
                        Representation\\ Compaction(\S\ref{sec:rep_compact})
                        [
                        {\parbox[t]{28em}{\hyphenpenalty=10000\exhyphenpenalty=10000\relax
                        \textit{e.g.,}
                        GGN~\citep{zhang2024gaussian}{,}
                        PixelGaussian~\citep{fei2024pixelgaussian}{,}
                        FreeSplat++~\citep{wang2025freesplat++}{,}
                        LongSplat~\citep{huang2025longsplat}{,}
                        AdaptiveSplat~\citep{singhal2026adaptivesplat}{,}
                        Off The Grid~\citep{moreau2025offgrid}{,}
                        SparseSplat~\citep{zhang2026sparsesplat}{,}
                        SubSplat~\citep{lee2026subsplat}{,}
                        FreeSplat\TaxSecondary~\citep{wang2024freesplat}
                        }}
                        ,leaf3, text width=28em
                    ]]
                ]
                [
                    Augmentation Strategies \\ ~(\S\ref{sec:data_visual_refine})
                    [
                    Data Augmentation\\(\S\ref{sec:data_refine})
                    [
                        {\parbox[t]{28em}{\hyphenpenalty=10000\exhyphenpenalty=10000\relax
                        \textit{e.g.,}
                        Puzzles~\citep{ma2025puzzles}{,}
                        MegaSynth~\citep{jiang2025megasynth}{,}
                        Aug3D~\citep{rauniyar2025aug3d}{,}
                        MVBoost~\citep{liu2025mvboost}{,}
                        Wild3R~\citep{furutani2026wild3r}{,}
                        Wat3R~\citep{ren2026wat3r}{,}
                        AREA3D~\citep{xu2025area3d}{,}
                        E-RayZer~\citep{zhao2025erayzer}
                        }}
                        , leaf4, text width=28em
                    ]],
                    [
                    Visual Augmentation\\(\S\ref{sec:visual_refine})
                    [
                        {\parbox[t]{28em}{\hyphenpenalty=10000\exhyphenpenalty=10000\relax
                        \textit{e.g.,}
                        MVSplat360~\citep{chen2024mvsplat360}{,}
                        ProSplat~\citep{lu2025prosplat}{,}
                        LatentSplat~\citep{wewer2024latentsplat}{,}
                        DIFIX3D+~\citep{wu2025difix3d+}{,}
                        One2Scene~\citep{wang2026one2scene}{,}
                        Pano3DComposer~\citep{qiu2026pano3dcomposer}{,}
                        SRSplat~\citep{hu2025srsplat}{,}
                        StereoSplat+~\citep{liu2026stereosplat}
                        }}
                        , leaf4, text width=28em
                    ]
                    ]
                ]
                [
                    Temporal-aware Models \\ ~(\S\ref{sec:temporal_aware})
                    [
                        Online Streaming \\ ~(\S\ref{sec:temporal_online})
                        [
                            {\parbox[t]{28em}{\hyphenpenalty=10000\exhyphenpenalty=10000\relax
                            \textit{e.g.,}
                            StreamSplat (Wu et al.)~\citep{wu2025streamsplat}{,}
                            CUT3R\TaxSecondary~\citep{wang2025continuous}{,}
                            StreamSplat (Song et al.)~\citep{song2026streamsplat}{,}
                            LASER~\citep{ding2025laser}{,}
                            OnlineSplatter~\citep{huang2025onlinesplatter}{,}
                            Event3R\TaxSecondary~\citep{huang2026event3r}{,}
                            DGS-LRM~\citep{lin2025dgs}{,}
                            LingBot-Map~\citep{chen2026lingbotmap}{,}
                            TTSA3R~\citep{zheng2026ttsa3r}{,}
                            VGGT-Motion~\citep{xiong2026vggtmotion}
                            }}
                            , leaf5, text width=28em
                        ]
                    ]
                    [
                        Offline Processing \\ ~(\S\ref{sec:temporal_offline})
                        [
                            {\parbox[t]{28em}{\hyphenpenalty=10000\exhyphenpenalty=10000\relax
                            \textit{e.g.,}
                            L4GM~\citep{ren2024l4gm}{,}
                            MonST3R~\citep{zhang2024monst3r}{,}
                            BTimer~\citep{liang2024feed}{,}
                            4D-LRM~\citep{ma20254d}{,}
                            4DGT~\citep{xu20254dgt}{,}
                            MoVieS~\citep{lin2025movies}{,}
                            Any4D~\citep{karhade2025any4d}{,}
                            OmniX~\citep{jiang2026omnix}{,}
                            PAGE-4D~\citep{zhou2025page4d}{,}
                            VGGT4D~\citep{hu2025vggt4d}{,}
                            4D-VGGT~\citep{wang20254dvggt}{,}
                            MoRe~\citep{fang2026more}{,}
                            TrajVG~\citep{miao2026trajvg}{,}
                            UniSplat~\citep{shi2025unisplat}{,}
                            V-DPM~\citep{sucar2026vdpm}{,}
                            DeGO~\citep{gao2026deformablegaussianoc}{,}
                            SPAR~\citep{cai2026dynamicrobustphotome}{,}
                            VGGT-World~\citep{sun2026vggtworld}{,}
                            WristWorld~\citep{qian2025wristworld}{,}
                            UFO-4D~\citep{hur2026ufo4d}
                            }}
                            , leaf5, text width=28em
                        ]
                    ]
                    [
                        Interactive Modeling \\ ~(\S\ref{sec:temporal_interactive})
                        [
                            {\parbox[t]{28em}{\hyphenpenalty=10000\exhyphenpenalty=10000\relax
                            \textit{e.g.,}
                            PIXIE~\citep{le2025pixie}{,}
                            PhysGM~\citep{lv2025physgm}
                            }}
                            , leaf5, text width=28em
                        ]
                    ]
                    [
                        Specialized Tasks \\ ~(\S\ref{sec:temporal_others})
                        [
                            {\parbox[t]{28em}{\hyphenpenalty=10000\exhyphenpenalty=10000\relax
                            \textit{e.g.,}
                            DAS3R~\citep{xu2024das3r}{,}
                            St4RTrack~\citep{st4rtrack2025}{,}
                            Event3R~\citep{huang2026event3r}{,}
                            DeGO\TaxSecondary~\citep{gao2026deformablegaussianoc}{,}
                            SPAR\TaxSecondary~\citep{cai2026dynamicrobustphotome}{,}
                            VGGT-World\TaxSecondary~\citep{sun2026vggtworld}{,}
                            WristWorld\TaxSecondary~\citep{qian2025wristworld}
                            }}
                            , leaf5, text width=28em
                        ]
                    ]
                ]
            ]
        \end{forest}
    }
    \vspace{-1.8em}
    \caption{\textbf{A taxonomy of feed-forward 3D reconstruction methods.} Representative methods spanning NeRF, 3DGS, pointmaps, voxels, meshes, occupancy fields, and signed-distance fields are grouped by five problem-driven directions. Unmarked entries indicate primary categories; $\dagger$ marks secondary memberships for methods appearing in multiple branches. The complete taxonomy is available at \rurl{ff3d-survey.github.io}.}
    \label{fig:ff_3d_reconstruction_taxonomy}
\end{figure*}

\subsection{Others}

Beyond NeRF, 3DGS, and pointmaps, several alternative representations have been explored within the feed-forward paradigm. Many of these predate the neural rendering era and, importantly, were among the first to demonstrate feed-forward 3D reconstruction from learned representations. These representations can be broadly categorized by whether they model \emph{geometry only} (\textit{e.g.}, occupancy, SDF, mesh) or \emph{geometry and appearance jointly} (\textit{e.g.}, light fields, texture fields, triplane-based radiance).

\noindent\textbf{Neural implicit representations for geometry.}
The methods in this group focus on recovering \emph{geometric structure} and do not, by themselves, model view-dependent appearance.
Occupancy Networks~\citep{mescheder2019occupancy} represent 3D shapes as continuous decision boundaries of deep classifiers, learning implicit occupancy functions that map 3D coordinates to binary occupancy probabilities, enabling high-resolution shape reconstruction without discretization artifacts. Concurrent works such as DeepSDF~\citep{park2019deepsdf}, which regresses continuous signed distance functions, and IM-NET~\citep{chen2019learning}, which learns implicit field decoders conditioned on shape embeddings, independently established the same core idea. Together, these works laid the conceptual foundation for feed-forward 3D reconstruction via neural implicit functions, and their influence extends to subsequent methods in this survey. Building upon Occupancy Networks~\citep{mescheder2019occupancy}, Convolutional Occupancy Networks~\citep{Peng2020ECCV} impose structured geometric reasoning through convolutional encoders and local implicit decoders, improving scalability to large scenes while maintaining robust reconstruction from noisy inputs.

\noindent\textbf{Joint geometry and appearance.}
Extending neural implicit representations beyond pure geometry, Texture Fields~\citep{oechsle2019texture} attach a learned texture function to implicit surfaces, enabling joint shape and appearance prediction in a feed-forward manner. Implicit Surface Light Fields~\citep{oechsle2020learning} further unify surface and radiance representations. Differentiable Volumetric Rendering~\citep{niemeyer2020differentiable} connects implicit surface representations with differentiable rendering, enabling feed-forward prediction of both geometry and novel views from images—bridging the gap between reconstruction and synthesis that NeRF would later popularize through volumetric density fields. These works are direct precursors to the NeRF-based and 3DGS-based pipelines that dominate the current landscape.

\noindent\textbf{Light fields and SDF-based methods.}
Light field networks~\citep{sitzmann2021light} parameterize scenes as 4D light fields, achieving real-time rendering without volumetric ray marching at the cost of limited extrapolation beyond observed rays and difficult explicit surface extraction. SDF-based methods~\citep{long2022sparseneus, ren2023volrecon, liang2023ReTR, 10378140, na2024uforecon, gao2025surfacesplat} excel at producing clean, watertight surfaces suitable for downstream geometry processing, but the inherent smoothness prior can cause under-reconstruction of fine details and thin structures.

\noindent\textbf{Explicit and hybrid representations.}
Mesh-based models~\citep{wei2024meshlrm, liu2024meshformer, zeng2025renderformer} offer direct compatibility with standard graphics and simulation pipelines, yet they face difficulties in representing topologically complex or semi-transparent scenes. 2DGS~\citep{chen2024lara, chang2025meshsplat} replaces volumetric Gaussians with oriented 2D disks for tighter surface coupling and more geometrically consistent reconstruction, trading off the capacity to represent volumetric phenomena such as fog and transparency. Triplane representations~\citep{honglrm, li2024instant3d, xu2024agg, zou2024triplane} provide a compact, memory-efficient intermediate form amenable to 2D convolutional processing, though axis-aligned factorization can introduce anisotropic artifacts for geometry misaligned with the canonical planes.

\noindent\textbf{Other specialized representations.}
3D-free approaches~\citep{suhail2022generalizablepatchbasedneuralrendering, jin2025lvsm} bypass explicit 3D modeling entirely with purely data-driven architectures, benefiting from architectural simplicity but potentially suffering from geometric inconsistencies across large viewpoint changes and the inability to recover explicit 3D structure. Specialized representations such as Pl\"ucker line fields~\citep{bahrami2025pluckerf} and planar primitives~\citep{liu2025plana3r} demonstrate that task-specific geometric priors can yield substantial gains in targeted domains like thin-structure recovery and indoor reconstruction.

\section{Research Directions}\label{model}


While~\S\ref{representation} focused on what to reconstruct, this section examines the key research directions that the community has pursued to push feed-forward models toward greater robustness, accuracy, and efficiency. 
Despite converging on a unified paradigm (~\S\ref{sec:preliminary}), feed-forward reconstruction still presents a number of open challenges that have attracted substantial research attention. For brevity, only representative methods are included.

As illustrated in ~\cref{fig:ff_3d_reconstruction_taxonomy}, we organize these methods into five directions: 1) feature enhancement (\S\ref{sec:feature_enhance}), which improves the quality of implicit representations through better architectures, cross-view fusion, or integration of visual foundation models; 2) geometry awareness (\S\ref{sec:geo_aware}), which incorporates geometric priors to resolve depth ambiguity and handle sparse or pose-free inputs; 3) model efficiency (\S\ref{sec:model_efficiency}), which reduces computation and memory overhead for practical deployment; 4) augmentation strategies (\S\ref{sec:data_visual_refine}), which leverage data or visual augmentation to improve generalization; and 5) temporal-aware models (\S\ref{sec:temporal_aware}), which extend the paradigm to dynamic scenes and streaming settings. This problem-driven taxonomy cuts across output representations. Methods built upon NeRF, 3DGS, or Pointmap may appear in any of the five directions, because the core challenge they address determines their categorization. For brevity, only representative methods are included. The same problem axes also include the geometry-only lineage established by Occupancy Networks, Convolutional Occupancy Networks, DeepSDF, and IM-NET~\citep{mescheder2019occupancy,Peng2020ECCV,park2019deepsdf,chen2019learning}, and extended by SparseNeuS, VolRecon, ReTR, UFORecon, SurfaceSplat, MeshLRM, MeshFormer, RenderFormer, LaRa, MeshSplat, and PLANA3R~\citep{long2022sparseneus,ren2023volrecon,liang2023ReTR,na2024uforecon,gao2025surfacesplat,wei2024meshlrm,liu2024meshformer,zeng2025renderformer,chen2024lara,chang2025meshsplat,liu2025plana3r}.

\subsection{Feature Enhancement}
\label{sec:feature_enhance}

The implicit feature maps form the key of the entire network in feed-forward neural rendering models. Their quality directly impacts the decoding of 3D scenes, and thus, enhancing these features is crucial for improving rendering accuracy and model generalization. A lot of work has been devoted to enhancing the features in the feed-forward model, which we can summarize into the following directions: 1) Architectures (\S\ref{sec:encode_arch}), which evolve the feature extractor from early CNN-based conditioning to transformers and state-space models for richer global context; (2) Cross-View Fusion (\S\ref{sec:view_fusion}), which aggregates multi-view features into a geometrically consistent representation; and (3) Integration of Visual Foundation Models (\S\ref{sec:prior_foundation}), which injects pre-trained geometric and semantic priors rather than learning all representations from 3D data alone.

\subsubsection{Architectures}
\label{sec:encode_arch}

\begin{figure}[t]
  \centering
  \includegraphics[width=\columnwidth]{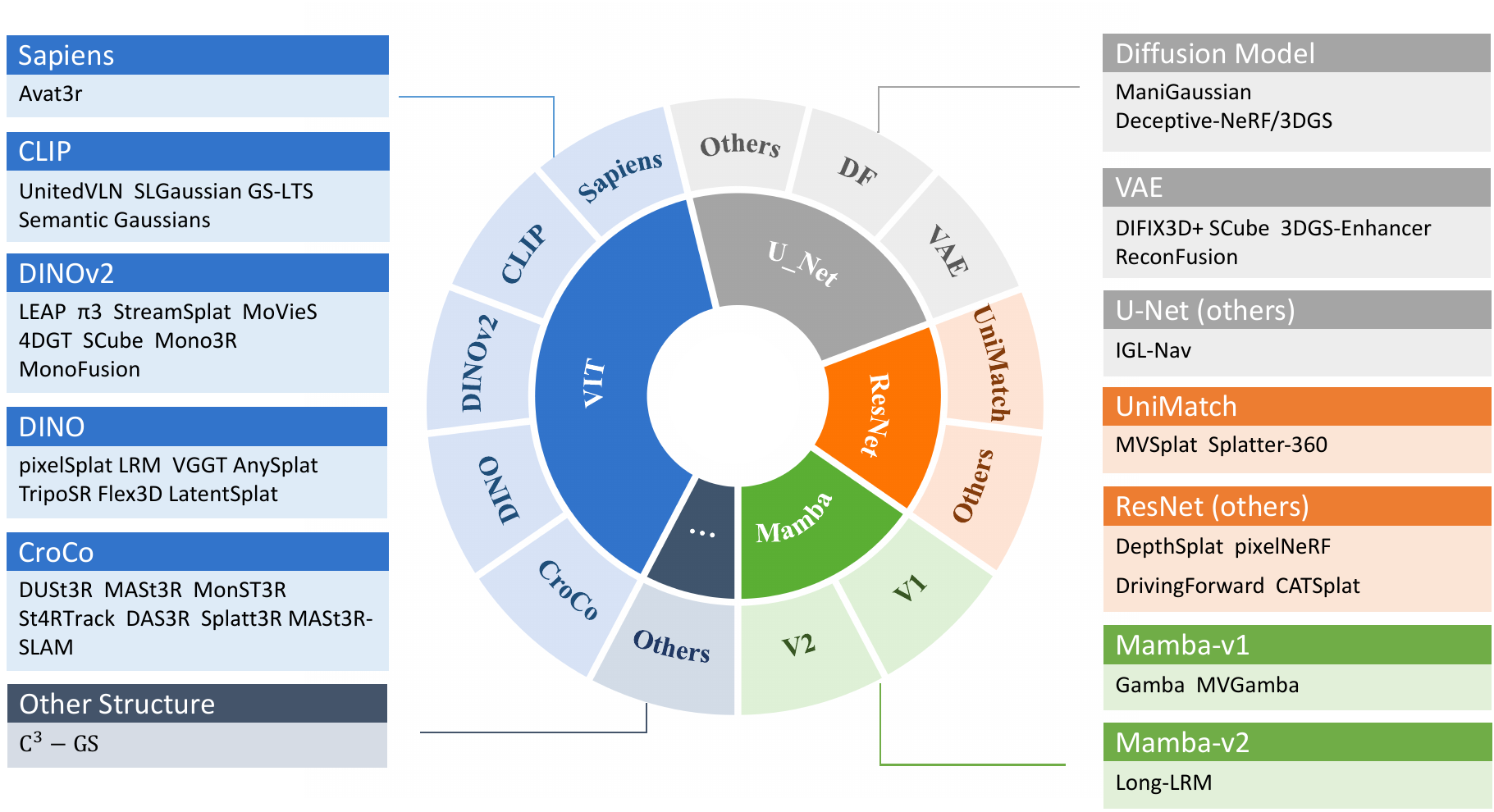}
  \caption{Encoder taxonomy in recent feed-forward 3D reconstruction models. Common backbones include ViT\citep{dosovitskiy2020vit}, ResNet\citep{he2015deepresiduallearningimage}, U-Net\citep{ronneberger2015u}, and Mamba\citep{mamba}/Mamba2\citep{mamba2}, often initialized with or augmented by large-scale pre-trained priors (\textit{e.g.}, CroCo\citep{NEURIPS2022_16e71d1a}, DINO\citep{zhang2022dino}/DINOv2\citep{oquab2023dinov2}, CLIP\citep{radford2021learning}, UniMatch\citep{xu2023unifying}, diffusion models~\citep{Rombach_2022_CVPR}, or VAEs\citep{kingma2013auto}) to inject visual/geometric knowledge learned from 2D data. Representative model instances are discussed in \S\ref{sec:prior_foundation}. }
  \label{fig:encoder-taxonomy}
\end{figure}

The design of the feature extraction architecture serves as the foundation for the entire reconstruction pipeline. As overviewed in Fig. 3, the community has explored a spectrum of encoder backbones, evolving from early ResNets~\citep{he2015deepresiduallearningimage} to ViTs~\citep{dosovitskiy2020vit} to inject rich visual and geometric knowledge.

Early feed-forward (\textit{a.k.a.} generalizable) neural rendering methods condition radiance evaluation on image-aligned features queried at projected locations. pixelNeRF~\citep{yu2021pixelnerfneuralradiancefields} pioneers fully-convolutional conditioning to predict NeRF from one or few views without per-scene optimization, making the per-ray MLP a function of local image features aggregated from nearby source views. By extracting features from a target viewpoint $\mathbf{x}$ with view direction $\mathbf{d}$, the image feature is passed into the decoder to compute color and density. IBRNet~\citep{wang2021ibrnetlearningmultiviewimagebased} takes a step further by employing a ray transformer, learning a continuous view-interpolation function that jointly handles density, occlusion and color blending along each query ray, improving generalization to novel scenes. Splatter Image~\citep{szymanowicz24splatter} uses U-Net as a backbone to predict pixel-aligned 3D Gaussians from a single image. For other representations, Convolutional Occupancy Networks~\citep{Peng2020ECCV} combine the expressive power of convolutional encoders with implicit occupancy decoders. LFN~\citep{sitzmann2021light} proposes to represent the scene as a 4D light field parameterized by a neural implicit function.

Subsequent research addresses specific architectural limitations. 
Inconsistent image features across views can lead to artifacts in 3D scene representations. NeuRay~\citep{liu2022neuralraysocclusionawareimagebased} mitigates this by predicting the visibility of 3D points from input views, thus enhancing feature consistency by focusing on visible points. 
${C}^{3}$-GS~\citep{hu2025c3} proposes a context-aware encoder that mixes information across spatial dimensions and scales, together with cross-dimension feature mixing that preserves local geometry cues for Gaussian prediction.

The introduction of transformers marks a significant evolution in encoding architecture. 
SRT~\citep{sajjadi2022scenerepresentationtransformergeometryfree} introduces a transformer-based decoder that processes input images into latent features for scene representations, enabling novel view generation by minimizing the novel-view reconstruction error. 
A major challenge for SRT in large-scale scenes is its handling of camera poses. The method's arbitrary reference camera selection can result in flickering artifacts, which necessitates model invariance to various parametrizations.
RePAST~\citep{safin2023repast} addresses this limitation by incorporating pairwise relative camera pose information into the attention mechanism, making the model invariant to global reference frame choices. Further refining this idea, GNT~\citep{t2023attentionnerfneeds} proposes a unified two-stage transformer framework for real-time novel view synthesis. In the first stage, a view transformer aggregates information from epipolar lines across neighboring views to produce coordinate-aligned features. In the second, a ray transformer employs attention-based decoding along sampled points during ray marching to render high-quality novel views directly from these features. In parallel to transformer decoders that directly query scene latents for rendering,
VisionNeRF~\citep{lin2023vision} incorporates transformer blocks primarily as a feature encoder, utilizing a ViT to extract global context via self-attention over image tokens and fusing it with CNN-based local features to condition NeRF-style volumetric rendering.

Another series of transformer-based models represents a significant shift. Large Reconstruction Model (LRM)~\citep{honglrm} and Instant3D~\citep{li2024instant3d} establish the core feature-level pipeline. They begin by using a ViT to encode each single 2D input image into a set of feature tokens. Then the transformer-based decoder takes these compact 2D features and expands them into a 3D representation, effectively unprojecting the features into a triplane feature grid. Pre-training on massive datasets teaches the decoder how to infer a complete and plausible 3D feature volume from the sparse cues present in the features of a single 2D image. Building upon the LRM network architecture, TripoSR~\citep{tochilkin2024triposr} integrates substantial improvements in data processing, model design, and training techniques.
GRM~\citep{xu2024grm} and GS-LRM~\citep{zhang2024gs} adapt the foundational LRM~\citep{honglrm} pipeline to improve efficiency. Instead of decoding features into an intermediate triplane that then requires further processing, these methods are trained to directly output the parameters for 3D Gaussian Splatting, streamlining the reconstruction process.
MeshLRM~\citep{wei2024meshlrm} retargets LRM~\citep{honglrm} from NeRFs to meshes by integrating differentiable marching cubes and rasterization into the model. MeshFormer~\citep{liu2024meshformer} stores features in 3D sparse voxels and blends transformers with 3D convolutions. Unlike pixel-aligned 3DGS or pointmap decoders, these geometry-only architectures must retain an inside--outside boundary, signed distance, or mesh connectivity throughout feature decoding. This stronger geometric bias yields explicit surfaces that can enter graphics and simulation pipelines directly, but topology changes and thin structures are harder to represent than with unconstrained points or Gaussians.
Flex3D~\citep{han2024flex3d} extends these methods to accommodate different numbers of input views and viewpoints.
Instead of relying on explicit 3D representations, LVSM~\citep{jin2025lvsm} employs a fully data-driven approach based on Transformers to tackle novel view synthesis.
VGGT~\citep{wang2025vggt} is a foundational model that enhances features through extensive pre-training on diverse visual geometry tasks. It learns an extremely rich and general-purpose visual-geometric feature space by being trained to predict multiple outputs such as camera poses, depth maps, and point clouds.
Depth Anything 3~\citep{depthanything3} achieved impressive results using a single plain transformer~\citep{oquab2023dinov2} without any additional special design.
Recent designs also reconsider the decoder interface rather than only enlarging the backbone: UniQueR~\citep{peng2026uniquer} uses sparse global 3D queries to predict both visible and occluded structure instead of dense pixel-aligned 2.5D outputs, while 2Xplat~\citep{jeong20262xplat} separates geometric and camera inference from Gaussian appearance generation through two specialized experts.
The decoder interface now forms a clearer design axis. Dense dual-stream prediction retains image detail, while token-based and view-conditioned decoders trade strict pixel alignment for global coherence and controllable primitive counts~\citep{ngo2026dage,ren2026tokengs,song2026uniquesplat,wang2026infinisplat,jeong2026viewsplat}. These designs converge on separating geometry from appearance but disagree on whether the scene state should be globally shared or specialized for each query view.

Unlike transformer-based models, Gamba~\citep{shen2025gamba} introduces a Mamba-based~\citep{mamba, mamba2} GambaFormer network to model single-image-to-3DGS reconstruction as sequential prediction with complexity that scales linearly with token length. Subsequently, MVGamba~\citep{yi2024mvgamba} extends it to multiple views with cross-view self-refinement. Long-LRM~\citep{ziwen2024long} tackles the challenge of reconstructing large scenes from long image sequences with a state-space model. This allows the model to maintain long-range consistency and build wide-coverage scenes without the quadratic complexity that makes a pure transformer approach infeasible for long sequences.

\textbf{Takeaway.} Global attention enables flexible cross-view interaction, but its cost grows with input length~\citep{zhang2024gs}. Hybrid state-space backbones reduce this burden~\citep{ziwen2024long}, making the balance between detailed interactions and scalable context more informative than a universal ranking of backbone families.


\subsubsection{Cross-View Fusion}
\label{sec:view_fusion}
A critical aspect of enhancing implicit representations lies in fusing information across multiple viewpoints to form a coherent and geometrically consistent 3D scene. Achieving this requires establishing robust cross-view feature correspondences that effectively capture spatial relationships between input images. AttnRend~\citep{du2023learningrendernovelviews} proposes using a multi-view ViT encoder to extract features from the input images while leveraging attention across views and their corresponding camera poses. This well-designed architecture extracts more effective features from the input images with great consistency between different views. 
For more effective and consistent features across different views, eFreeSplat~\citep{min2024epipolarfree} uses a self-supervised ViT with cross-completion pre-training and introduces a Gaussian alignment module to iteratively refine the Gaussian parameters predicted by each view through a 2D U-Net. This method outperforms approaches that utilize epipolar lines, achieving improvement solely through feature enhancement. 
Following the LRM series, LGM~\citep{tang2024lgm} presents an asymmetric U-Net as a high-throughput backbone operating on multi-view images and directly regressing the parameters for the 3DGS representation. 
iLRM~\citep{kang2025ilrm} introduces an iterative cross-view refinement loop to enhance feature quality. Unlike a single-pass LRM, iLRM feeds its features back into the network over multiple iterations and decomposes fully attentional multi-view interactions into a two-stage attention scheme to reduce computational costs. By effectively utilizing more input views, iLRM provides significantly higher reconstruction quality at comparable computational cost.

Beyond feature-level alignment, a geometry-first line of work establishes a stronger foundation by directly regressing 3D coordinates and camera geometry. DUSt3R~\citep{wang2024dust3r} establishes a new foundation for geometric 3D vision by reformulating pairwise and multi-view stereo as a direct, dense coordinate regression problem. The model simultaneously predicts dense pointmaps and relative camera poses from unposed image pairs. 
MASt3R~\citep{leroy2024grounding} augments DUSt3R with a dense local feature head and a novel matching loss, enabling robust and accurate matching capabilities. By introducing a fast reciprocal matching scheme, MASt3R achieves both theoretical efficiency and strong empirical performance.
To overcome the pairwise matching bottleneck, MV-DUSt3R~\citep{tang2024mv} incorporates multi-view decoder blocks that facilitate cross-view information exchange. Its enhanced variant, MV-DUSt3R+~\citep{tang2024mv}, further integrates multiple reference views and introduces a Gaussian prediction head for direct rendering supervision. This turns feature enhancement into a scalable multi-view fusion framework capable of reconstructing large scenes within seconds. 
MUSt3R~\citep{cabon2025must3r} enhances multi-view features by introducing a symmetric and memory-efficient architecture. Its symmetric design ensures consistency regardless of the reference view choice, while a memory mechanism allows it to scale to a large number of views. The enhancement lies in its robust fusion process, which effectively aggregates features from multiple stereo pairs into a unified and coherent 3D representation. 
When processing long sequences, reconstruction quality and real-time performance often cannot be achieved simultaneously. WinT3R~\citep{li2025wint3rwindowbasedstreamingreconstruction} relaxes this constraint via a sliding-window Transformer that reasons over a short, fixed-lag window, trading minimal latency for stronger global consistency. All of the above methods are designed for pixel-aligned predictions, yet they remain constrained by 2D feature matching and averaged density. Dens3R~\citep{fang2025dens3r} is a unified geometry foundation model that jointly predicts correlated dense geometric quantities such as pointmaps, depth, and normals within a shared representation. MoRE~\citep{gao2025more} introduces a mixture-of-experts architecture for dense 3D visual geometry reconstruction, improving scalability and specialization across diverse geometric tasks. Flow3r~\citep{cong2026flow3r} learns scalable visual geometry from unlabeled monocular videos by using factored flow prediction to disentangle geometry and camera motion supervision. NOVA3R~\citep{chennova3r} departs from pixel-aligned prediction by learning a global scene representation and decoding complete amodal geometry from unposed images. Gen3R~\citep{huang2026gen3r} bridges feed-forward reconstruction and video diffusion by aligning geometric and appearance latents for scene-level 3D generation. Uni3R~\citep{sun2025uni3r} jointly reconstructs semantic 3D Gaussian primitives from unposed multi-view images for unified rendering, depth prediction, and open-vocabulary 3D understanding.
VolSplat~\citep{wang2025volsplat} directly regresses Gaussians from 3D features based on a voxel-aligned prediction strategy, thereby resolving these limitations.
Recent fusion mechanisms differ in where they enforce alignment. PRoPE encodes complete camera frusta within attention, while TokenSplat combines token-aligned prediction, camera tokens, and fusion confidence to align unposed views and suppress redundant Gaussians~\citep{li2025prope,li2026tokensplat}. FF3R instead couples semantic and geometric prediction under RGB and feature rendering supervision~\citep{zhou2026ff3r}. Geometry-aware position encoding offers explicit camera control but depends on pose quality. Confidence gating is more robust to redundant observations but may discard useful uncertain evidence, whereas semantic fusion supplies broader correspondence cues at additional representation cost. This line has converged on alignment inside the predictor rather than post-processing, but no design removes the trade-off between explicit geometric control and learned robustness.

When moving to long sequences, the main challenge shifts to maintaining global consistency. PreF3R~\citep{chen2024pref3rposefreefeedforward3d} introduces a spatial memory network that acts as a persistent global scene representation. As each new image from a sequence is processed, its features are used to update this memory. This incremental fusion allows the model to aggregate temporal information and progressively refine the implicit features, leading to a more complete and coherent reconstruction over time, even with variable-length inputs. 
Similar to PreF3R~\citep{chen2024pref3rposefreefeedforward3d}, Spann3R~\citep{wang20243d} enhances features by using a persistent memory, but it formalizes this as an external 3D spatial memory that stores point-wise features. For each new frame, features are extracted and then ``written'' into this memory based on their projected 3D locations. This mechanism allows for the continuous accumulation and refinement of features over long periods, effectively enhancing the scene representation by integrating information across time and space.
Concurrently, CUT3R~\citep{wang2025continuous} also establishes a global state for processing streams of images through a compressed state representation, and is not limited to capturing the observed scene content~\citep{wang20243d}, but can also infer unobserved structures. Finally, unlike the methods that only work for static scenes, it can also seamlessly reconstruct dynamic scenes. 
Furthermore, G-CUT3R~\citep{khafizov2025g} enhances CUT3R by integrating priors from pre-trained camera pose and monocular depth estimators. 
TTT3R~\citep{chen2025ttt3r} adopts a test-time training perspective, aiming to derive a closed-form learning rate for memory updates from memory states and new observations, thereby enhancing length generalization. Point3R~\citep{wu2025point3rstreaming3dreconstruction} explores the complementary, point-centric extreme, prioritizing minimalist causal reconstruction and tracking for strong real-time behavior.
IncVGGT~\citep{fangincvggt} is a training-free incremental variant of VGGT that enables memory-bounded long-range reconstruction without full-sequence processing. ZipMap~\citep{jin2026zipmap} compresses an image collection into a compact hidden scene state with test-time training, achieving linear-time bidirectional reconstruction. LoGeR~\citep{zhang2026loger} processes long videos in chunks and combines test-time-training memory with sliding-window attention for globally consistent long-context reconstruction. tttLRM~\citep{wang2026tttlrm} uses a test-time-training layer to support long-context autoregressive 3D reconstruction with linear complexity and explicit Gaussian-splat decoding. VGG-T$^3$~\citep{elflein2026vgg} distills the varying-length scene representation into a fixed-size MLP via test-time training to scale offline feed-forward reconstruction linearly with the number of views. 

These sequence-specific updates differ in mechanism. TTT3R adjusts recurrent memory with a confidence-dependent rule, whereas tttLRM updates fast weights through a token-based learning objective~\citep{chen2025ttt3r,wang2026tttlrm}. We distinguish their adaptation mechanisms and costs from direct prediction and scene refinement in \S\ref{sec:visual_refine}.

\textbf{Takeaway.} Cross-view fusion must establish a shared geometric frame as well as aggregate features~\citep{ye2024no,leroy2024grounding}. Extending the context alone does not resolve inconsistent correspondences; long sequences additionally require a retention mechanism that balances local alignment with historical evidence~\citep{zhang2026loger}, as discussed in \S\ref{sec:temporal_online}.

\subsubsection{Integration of Visual Foundation Models}
\label{sec:prior_foundation}
Leveraging large-scale pre-trained foundation models to inject powerful visual and geometric priors into 3D reconstruction pipelines is a significant paradigm. These models enhance implicit representations by transferring knowledge learned from vast and diverse 2D datasets, significantly improving generalization and data efficiency.

DUSt3R~\citep{wang2024dust3r} is among the first to incorporate a pre-trained model, CroCo~\citep{NEURIPS2022_16e71d1a}, to establish robust feature correspondences between images in a feed-forward reconstruction framework. 
Building on this direction, Mono3R~\citep{li2025mono3r} integrates strong single-view priors~\citep{wang2025moge, oquab2023dinov2} into feed-forward architectures. It introduces a mono-guided refinement module that fuses multi-view stereo features with representations from a monocular feature branch, enriching the implicit representation with geometrically plausible details. This design enhances reconstruction quality in regions where multi-view correspondence is weak or unreliable.

To further extend the use of visual foundation models, Feat2GS~\citep{chen2025feat2gs} demonstrates that high-quality 3D reconstruction can be achieved by probing pre-trained 2D models, such as those from segmentation~\citep{kirillov2023segment}, diffusion~\citep{Rombach_2022_CVPR}, recognition~\citep{He_2022_CVPR, Ranzinger_2024_CVPR, pmlr_radford21a}, and representation learning~\citep{oquab2023dinov2, caron2021emerging, 9178977, wang2024dust3r, leroy2024grounding}, without the need for retraining large 3D networks. It employs a lightweight decoder to directly map them into the parameters of a 3DGS representation. This approach effectively reveals that pre-trained 2D features already encode rich geometric priors for high-fidelity 3D reconstruction.

Extending VGGT's visual-geometry foundation-model paradigm, VGGT-$\Omega$~\citep{wang2026vggtomega} scales feed-forward reconstruction through larger models, substantially expanded supervised data, and self-supervised learning from unlabeled videos. Its scene registers provide a compact and transferable scene representation; their alignment with language and utility for vision-language-action models suggest that reconstruction pre-training can support broader spatial understanding.
Geometry and language are also becoming joint outputs rather than one-way priors. G2VLM and LangFlash align reconstructed 3D structure with semantic tokens, which improves spatial querying but forces the geometry branch to share capacity with reasoning and language supervision~\citep{hu2025g2vlm,liu2026langflash}.

Complementary to visual foundation integration, CATSplat~\citep{roh2024catsplat} leverages textual guidance from vision-language models to augment single-image 3D reconstruction. By conditioning the reconstruction on language-derived semantics, CATSplat compensates for missing visual information and enhances structural plausibility when visual cues are limited.

\textbf{Takeaway.} Pretrained priors are most useful when they complement missing observations, through monocular geometry or semantic completion~\citep{li2025mono3r,roh2024catsplat}. Their role should therefore be matched to the ambiguity in the input, while plausible completion must still be checked against the observed geometry.

\subsection{Geometry Awareness}
\label{sec:geo_aware}

\begin{figure}
    \centering
    \includegraphics[width=0.9\linewidth]{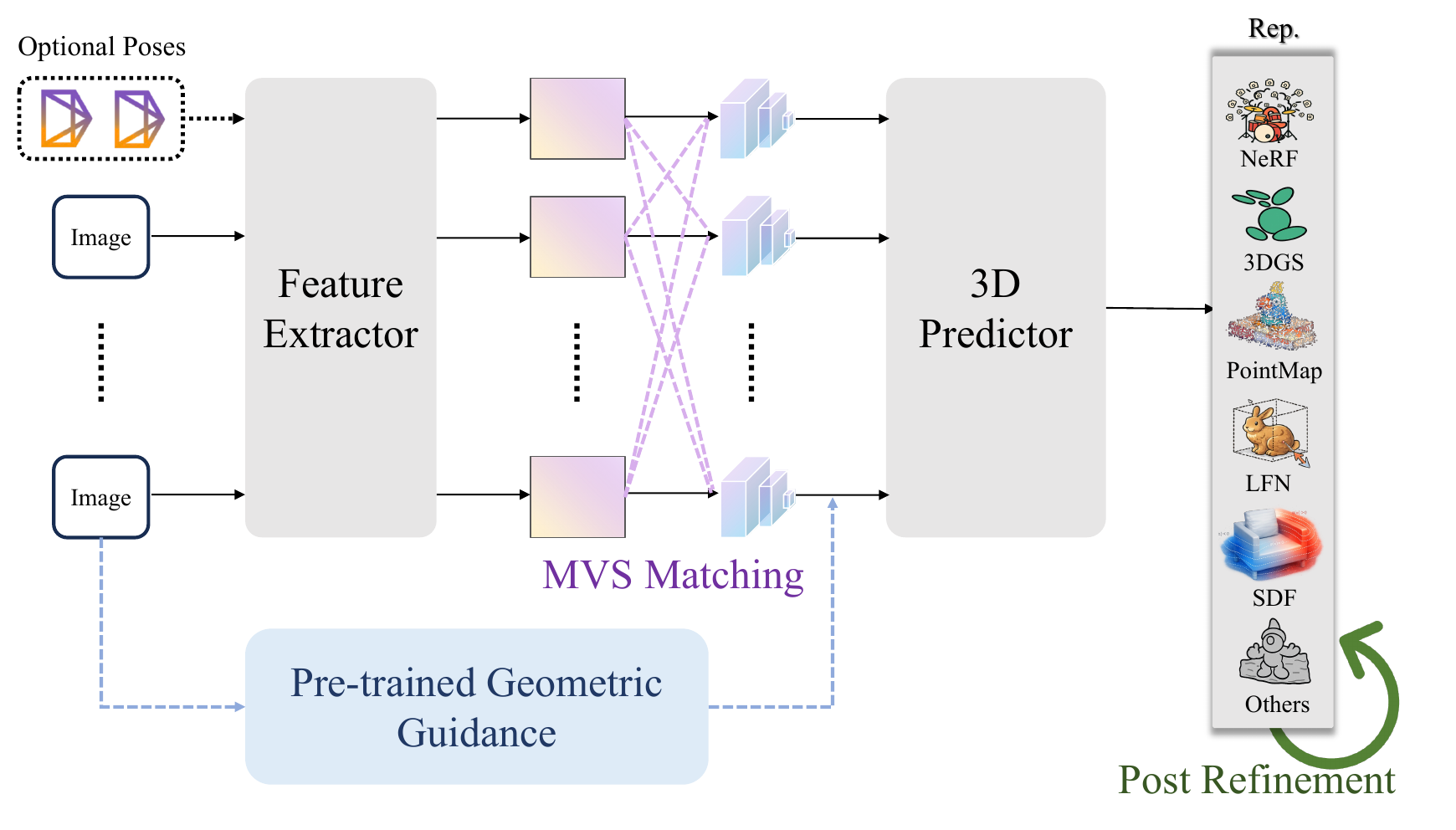}
    \caption{\textbf{Overview of research directions for geometry-aware improvement.}}
    \label{fig:geo_aware_pipeline}
\end{figure}


A central challenge in feed-forward 3D reconstruction lies in robust, accurate inference of underlying scene geometry. As illustrated in \cref{fig:geo_aware_pipeline}, the geometry-aware pipeline generally takes images and optional poses as input through a feature extractor and 3D predictor to produce various representations, while different research directions improve this pipeline from complementary angles. The fidelity of the reconstructed shape is paramount, as it directly dictates the photorealism and multi-view consistency of the final output, preventing artifacts such as floaters or distorted surfaces. Consequently, research in this area focuses on developing architectures that incorporate stronger geometric reasoning. This pursuit yields a diverse array of strategies, which can be broadly categorized as follows: 1) Explicit Geometric Aggregation (\S\ref{mvs_methods}), which incorporates structures such as cost volumes and epipolar constraints to explicitly encode multi-view relationships; 2) Post Refinement (\S\ref{sec:refine_3d}), which improves the generated primitives to better capture complex geometry, either through learned coarse-to-fine stages inside a single forward pass or through iterative updates at test time; 3) Pose-free Reconstruction (\S\ref{sec:pose_free}), which removes the dependency on known camera parameters by jointly inferring geometry and poses; and 4) Pre-trained Geometric Guidance (\S\ref{sec:pretrained_geo}), which transfers rich geometric priors from foundation models to enhance reconstruction quality.
Mechanistic probing complements these architectural categories: an analysis of DUSt3R-, VGGT-, and Depth-Anything-3-style models~\citep{bratulic2025geometrypriors} finds that their intermediate attention correspondences encode emergent epipolar geometry, providing evidence that learned priors can internalize geometric constraints even without an explicit geometric module.

\subsubsection{Explicit Geometric Aggregation}
\label{mvs_methods}

Relying solely on 2D image features can lead to geometric ambiguities. To address this issue, recent methods introduce explicit geometric aggregation mechanisms that encode geometric relationships across multiple views. These methods differ mainly in how such geometric evidence is constructed and propagated, ranging from cost volumes and correspondence constraints to surface-aware modeling and geometry-guided Gaussian prediction.

A representative line of work relies on cost-volume construction to explicitly aggregate geometric evidence across multiple views. MVSNeRF~\citep{chen2021mvsnerffastgeneralizableradiance} first explores this direction and uses image features with plane-sweep transformation to build a cost volume. Then, 3D convolution is used to aggregate them to form a feature volume. The feature volume is then decoded into a color field and a density field. Among them, the cost volume, as a strong geometric prior, guides the network to achieve accurate 3D structure estimation and consistent novel-view synthesis. On this basis, GeoNeRF~\citep{johari2022geonerfgeneralizingnerfgeometry} proposes an explicit two-stage design for modeling geometry and occlusion. It first builds cascaded cost volumes through a dedicated geometry reasoner, and then uses these cost volumes to guide a Transformer-based renderer, thereby achieving higher rendering fidelity in scenes with complex geometric structures. To enhance robustness in large-scale scenarios, BoostMVSNeRFs~\citep{su2024boostmvsnerfs} introduces scale-aware priors and adaptive mechanisms based on MVSNeRF, enabling it to achieve high-quality reconstruction results at the urban scale and in open environments. MuRF~\citep{xu2024murfmultibaselineradiancefields}, on the other hand, discretizes space into multiple planes aligned with the target camera to construct a target-view frustum volume. This strategy can effectively aggregate cross-view features and capture contextual information through 3D convolutions, thereby achieving clearer geometric structures and finer spatial details.


Beyond dense cost-volume construction, another line of work turns to correspondence-based geometric reasoning. SRF~\citep{chibane2021stereoradiancefieldssrf} introduces a neural stereo framework. It establishes stereo correspondences between image pairs to infer 3D point attributes, avoiding the construction of a full cost volume. This method provides robust geometric signals by modeling the similarity between paired features, enabling the model to maintain strong generalization ability even with sparse views or large baselines. Following a correspondence-based strategy, GPNR~\citep{suhail2022generalizablepatchbasedneuralrendering} uses epipolar geometry from the reference view to extract local patches along the epipolar lines. Subsequently, the information of these geometrically constrained patches is aggregated through a Transformer to predict the color of the target ray, and it demonstrates high robustness under large-baseline configurations without relying on volume rendering.

The concept of using feature similarity as a geometric proxy is formalized in MatchNeRF~\citep{chen2023explicit}, and the research shows that the cosine similarity between features projected onto 2D positions corresponding to 3D points can serve as an effective geometric prior. By combining cross-attention to optimize feature matching, MatchNeRF establishes a strong correlation between feature similarity and volume density, thereby guiding a more realistic reconstruction. Based on this perspective, GTA~\citep{Miyato2024GTA} generalizes the attention formula by explicitly encoding queries and geometric transformations between key-value tokens. This embeds the relative 3D structure into the attention mechanism, enhancing the representation efficiency and geometric awareness.

Beyond volumetric and correspondence-based reasoning, several methods impose stronger geometric structure through \textbf{surface-aware representations}. SparseNeuS~\citep{long2022sparseneus} combines SDF-based surfaces with volume rendering, ensuring the stability of geometry with limited views through customized sampling or regularization. VolRecon~\citep{ren2023volrecon} introduces SRDFs together with a ray-based rendering formulation, retaining the advantages of SDF-based surfaces while improving supervision quality and generalization in few-view input scenarios. ReTR~\citep{liang2023ReTR} reimagines the rendering process using transformers, explicitly modeling depth distributions and surface reasoning via meta-ray tokens and cross-attention, thereby significantly improving zero-shot neural surface reconstruction performance. C2F2NeUS~\citep{10378140} constructs per-view cost frusta, fuses them in a cascade manner, and subsequently reverts to an implicit SDF representation. This coarse-to-fine frustum fusion captures both global and local structures, enabling high-fidelity surface reconstruction with strong generalization. For arbitrary and unfavorable view sets, UFORecon~\citep{na2024uforecon} removes the dependency on predefined ``superior'' view combinations, enabling robust performance under highly sparse, misaligned, or biased input conditions. SurfaceSplat~\citep{gao2025surfacesplat} proposes a hybrid approach, integrating SDF for coarse geometry to enhance 3DGS-based rendering with 3DGS-rendered images that refine SDF details to achieve more accurate surface reconstruction.
RenderFormer~\citep{zeng2025renderformer} demonstrates a pipeline for rendering directly from a triangle mesh. It uses a Transformer to learn global illumination effects, with a positional encoding based on the 3D spatial position of triangles, making the model inherently aware of the explicit mesh geometry.

More recently, explicit geometric aggregation has been extended beyond classical radiance-field pipelines to Gaussian-based and hybrid representations. AGG~\citep{xu2024agg} proposes a cascading pipeline that first generates rough representations of positions and triplanes and then performs upsampling through the 3D Gaussian high-resolution module. TGS~\citep{zou2024triplane} also introduces a hybrid triplane-Gaussian intermediate representation, which is decoded through the transformer network for single-view 3D reconstruction. LaRa~\citep{chen2024lara} is an efficient feed-forward radiance field model for large baselines, representing the scene as Gaussian volumes and unifying local and global reasoning through group-attention in the transformer layer. It achieves robust reconstruction through 2D Gaussian Splatting and is trained rapidly on moderate computational resources. MeshSplat~\citep{chang2025meshsplat} predicts pixel-aligned 2D Gaussians for each view to supervise the geometry, without the need for 3D ground-truth data, and achieves accurate sparse-view mesh extraction through weighted-Chamfer depth regularization and normal alignment. These surface-aware methods differ from pointmap coordinate regression and photometric 3DGS training at both the supervision and rendering stages: occupancy and SDF predictors learn a continuous decision surface from occupancy, signed-distance, depth, or normal constraints, meshes expose connectivity to differentiable rasterization, and hybrid methods such as SurfaceSplat and MeshSplat couple surface constraints to efficient splatting. Their explicit geometry is consequently assessed with Chamfer distance and normal consistency rather than PSNR alone; conversely, smoothness priors or a fixed mesh topology can erase thin structures or produce an incorrect surface even when novel views appear plausible.


A parallel research thread integrates MVS principles with 3DGS. 
pixelSplat~\citep{charatan2024pixelsplat} learns to predict a dense probabilistic distribution over 3D space, from which Gaussian means are differentiably sampled. 
MVSplat~\citep{chen2024mvsplat} revisits plane-sweep cost volumes to accurately localize Gaussian centers, leveraging cross-view feature similarity as a powerful geometric cue. The result is a lightweight yet highly accurate reconstruction model. 
MVSGaussian~\citep{liu2025mvsgaussian} further fuses MVS-derived point clouds with 3D Gaussian optimization, using the former as high-quality geometric initialization to combine the precision of MVS with the efficiency of 3DGS rendering. 
TranSplat~\citep{zhang2025transplat} identifies failure modes of prior feed-forward approaches~\citep{chen2024mvsplat} under limited view overlap and unreliable matching. It mitigates these issues using two key strategies: a learned depth-confidence map that guides local feature matching, and monocular depth priors that fill in regions lacking correspondence.

Finally, several recent methods pursue hybrid architectures that balance geometric consistency with computational efficiency. 
H3R~\citep{jia2025h3r} employs a compact latent volume to enforce 3D consistency, while a camera-aware Transformer conditioned on Plücker coordinates refines correspondence where stereo cues are weak, thereby achieving faster convergence and improved robustness in textureless areas. 
MuGS~\citep{lou2025mugs} unifies MVS-based volumetric evidence and monocular depth cues through a projection-sampling depth-consistency module and a probabilistic volume regularizer for Gaussian prediction. A reference-view loss further stabilizes appearance, leading to high-fidelity and geometrically consistent reconstructions.
Recent systems also use compact geometric intermediates to bridge sparse evidence and dense prediction. AMB3R~\citep{wang2025amb3r} fuses pointmap features in a sparse volumetric backend to recover metric-scale geometry that transfers to visual odometry and SfM, whereas GGPT~\citep{chen2026ggpt} obtains reliable sparse geometry with lightweight SfM and propagates it into dense pointmaps through a 3D point Transformer.
Recent Gaussian systems increasingly treat intermediate geometry as a shared scaffold rather than only a rendering aid. VGD, Splat-SAP, and MAC-Splat enforce metric, scale-aware, or multi-attribute agreement before appearance decoding~\citep{lin2025vgd,zhou2025splatsap,yang2026macsplat}. AnchorSplat, AirSplat, and GPOcc instead structure the prediction around geometric anchors, reliability estimates, or sparse occupancy~\citep{zhang2026anchorsplat,bui2026airsplat,zhou2026generalizingvisualge}. The first route preserves a dense image-to-Gaussian path, while the second exposes a compact geometric state that is easier to validate but more dependent on the quality of the prior.

\textbf{Takeaway.} MVSplat constrains correspondence through cost-volume depth hypotheses, whereas GS-LRM uses pure attention to learn cross-view geometry without constructing such volumes~\citep{chen2024mvsplat,zhang2024gs}. The trade-off is between a structured matching prior that depends on reliable camera geometry and flexible learned interactions. MVSplat also uses cross-view attention, showing that the two mechanisms can be complementary rather than mutually exclusive~\citep{chen2024mvsplat}.

\subsubsection{Post Refinement}
\label{sec:refine_3d}
Post-refinement methods improve an initial prediction through different update mechanisms. Learned stages can refine geometry without computing test-time gradients. HiSplat~\citep{tang2024hisplat} allocates coarse and fine Gaussians through error-aware hierarchical prediction; FreeSplat~\citep{wang2024freesplat} aligns and fuses Gaussian triplets to suppress overlap. PixelGaussian~\citep{fei2024pixelgaussian} adapts Gaussian distribution and count, while GGN~\citep{zhang2024gaussian} pools a graph of Gaussian groups into a compact representation. GD~\citep{nam2025generative} learns to densify an initial representation in a forward pass. These designs differ from ReSplat~\citep{xu2025resplat}, which repeatedly uses rendering-error feedback to predict Gaussian updates with a fixed network and no explicit gradient computation. Its recurrent inference has a selectable iteration budget rather than a single refinement step.

G3R~\citep{chen2024g3r} instead initializes neural Gaussians from an approximate geometry scaffold and repeatedly applies a learned update network conditioned on rendering-loss gradients. The reconstruction network remains fixed, but each scene requires gradient computation and iterative updates. We detail this inference regime in \S\ref{sec:visual_refine}.
RobustVGGT~\citep{han2025outlierrejection} instead refines the input set without retraining: it identifies an outlier-suppressing layer that emerges inside VGGT and uses its response to reject non-overlapping distractor views before they corrupt the reconstruction.
Robustness can also be represented rather than filtered. Trust3R predicts calibrated geometric uncertainty and refines its mean estimate in one pass, whereas RobustVGGT removes suspect observations before reconstruction~\citep{zhu2026trustitor,han2025outlierrejection}. This distinction separates uncertainty-aware output from input-set selection.

\textbf{Takeaway.} Learned refinement can use scene-specific feedback without explicit gradients, as in ReSplat, while G3R requires rendering-loss gradients~\citep{xu2025resplat,chen2024g3r}. Thus, fixed network weights do not imply single-step reconstruction; the update signal and iteration budget determine the practical inference cost.

\subsubsection{Pose-Free Reconstruction}
\label{sec:pose_free}
A major leap for feed-forward methods is the move towards reconstruction from uncalibrated images, where camera poses are unknown. This requires the model to infer geometry and camera parameters simultaneously. 

LEAP~\citep{jiangleap} is a foundational work in this area for radiance fields, discarding explicit camera poses entirely. Instead, it aggregates 2D image features into a shared neural volume based on feature similarity, learning geometry directly from the data. This paradigm is powerfully realized by DUSt3R~\citep{wang2024dust3r}, which formulates pairwise reconstruction as the regression of dense pointmaps. This unified approach learns strong geometric priors, enabling direct 3D inference without known camera parameters. Recognizing that some priors may be available, Pow3R~\citep{jang2025pow3r} extends the DUSt3R architecture with a lightweight conditioning mechanism, allowing it to incorporate auxiliary information such as intrinsics or relative poses at test time to improve accuracy. Building on pointmaps, $\pi^{3}$~\citep{wang2025pi} introduces a fully permutation-equivariant architecture that predicts affine-invariant camera poses and scale-invariant pointmaps, making the model robust to input order and highly scalable.

The 3DGS community has also taken up the challenge of pose-free reconstruction. Splatt3R~\citep{smart2024splatt3r} directly adapts the geometric framework of MASt3R~\citep{leroy2024grounding} to predict the attributes required for Gaussian splatting from uncalibrated image pairs. A simpler but surprisingly effective approach is taken by NoPoSplat~\citep{ye2024no}, which reconstructs a scene by predicting all Gaussians in the local coordinate system of a single canonical input view, cleverly sidestepping the need for pose estimation and global alignment. For more complex scenarios, PF3plat~\citep{hong2024pf3plat} develops a pipeline that uses pre-trained monocular depth and correspondence models to achieve a coarse geometric alignment, which is then refined by lightweight learnable modules. FreeSplatter~\citep{xu2024freesplatter} employs a streamlined transformer architecture to directly decode multi-view image tokens from uncalibrated images into 3D Gaussian primitives within a unified reference frame. In contrast, FLARE~\citep{zhang2025flare} proposes a cascaded learning paradigm that first estimates camera poses and then uses these estimates to condition the subsequent learning of a global geometric structure, which initializes the Gaussian centers. For cases where local reconstructions are easier to obtain, RegGS~\citep{cc2025_reggs} introduces a registration-based framework that aligns locally generated 3D Gaussians into a globally consistent scene using a novel metric for Gaussian mixture models. Addressing the practical challenge of difficult inputs, UFV-Splatter~\citep{fujimura2025ufv} develops an adaptation framework that enables pretrained pose-free models to handle unfavorable views by leveraging geometric priors learned from more favorable images. AnySplat~\citep{jiang2025anysplat} couples a geometry transformer for unposed inputs with a decoder that predicts Gaussian parameters, enabling zero-annotation pipelines that still match pose-aware baselines in quality. It integrates a differentiable pose estimation module into its architecture, allowing end-to-end training. Similarly, SPFSplat~\citep{huang2025no} remarkably eliminates the need for pose supervision by jointly predicting camera extrinsics and Gaussian primitives inside a single feed-forward network, while a reprojection loss anchors the learned canonical geometry and a rendering loss enforces photometric fidelity. SPFSplatV2~\citep{huang2025spfsplatv2} introduces a unified architecture with a masked attention mechanism and a learnable pose token, improving the accuracy of camera pose estimation and overall efficiency. For metric 3D reconstruction of indoor scenes, PLANA3R~\citep{liu2025plana3r} introduces planar 3D primitives and learns planar structures without explicit plane supervision. YoNoSplat~\citep{ye2025yonosplat} uses a single feed-forward model to reconstruct high-quality 3D Gaussian Splatting from an arbitrary number of posed or unposed images.
The pose-free setting is also expanding beyond pinhole imagery. Wid3R~\citep{jung2026wid3r} combines spherical-harmonic ray encoding with camera-model tokens to unify pinhole, fisheye, and spherical inputs, while CAM3R~\citep{guruprasad2026cam3r} decouples per-pixel ray prediction from radial-distance inference and aligns them globally for calibration-free wide-angle reconstruction.
Lens robustness exposes a tradeoff between camera-model conditioning for broad coverage and targeted adaptation for accuracy within one lens family~\citep{jung2026wid3r,guruprasad2026cam3r,duan2026fisheye3r}. A related choice appears in pose-free Gaussian prediction. WildSplat couples appearance conditioning to joint pose estimation, whereas FlexSplat removes point-cloud correspondence and predicts cameras, depth, and compact Gaussians together~\citep{zhang2026wildsplat,sabbaghziarani2026flexsplat}.

\textbf{Takeaway.} Removing input calibration shifts the burden to coordinate-frame selection and cross-view alignment rather than eliminating it~\citep{ye2024no,cc2025_reggs}. Joint and permutation-equivariant prediction reduce dependence on a privileged input ordering, but scale and frame conventions remain essential to interpreting the output~\citep{wang2025pi}.

\subsubsection{Pre-trained Geometric Guidance}
\label{sec:pretrained_geo}
A promising strategy to enhance reconstruction fidelity is to directly inject geometric cues derived from powerful pre-trained models. Modern monocular estimators have achieved remarkable robustness in predicting depth, normals, and optical flow. Integrating these off-the-shelf priors allows the model to bypass the difficult cold-start problem of geometry learning. 

The crucial role of depth as a geometric prior is explored directly by DepthSplat~\citep{xu2024depthsplat}, which establishes a powerful synergistic link between multi-view depth estimation and 3DGS. By using a robust depth model that leverages pretrained monocular features, it produces high-quality feed-forward reconstructions. The work also demonstrates that the differentiable 3DGS module can serve as an unsupervised objective for training depth models. The quality of this geometric prior is paramount, as highlighted in PM-Loss~\citep{shi2025pmloss}, which introduces a regularization loss based on a feed-forward pointmap model. While it is potentially less accurate than a depth map, it enforces geometric smoothness at object boundaries, leading to cleaner depth priors and fewer artifacts in the final 3DGS output. AnySplat~\citep{jiang2025anysplat} uses VGGT~\citep{wang2025vggt} weights to initialize the geometry transformer, thereby obtaining better geometric representations. Fin3R~\citep{ren2025fin3r} enriches the encoder with fine geometric details distilled from a strong monocular teacher model~\citep{wang2025moge} using a custom and lightweight LoRA adapter.
A complementary trend allows geometric priors to be supplied only when available rather than fixing a single input protocol. WorldMirror~\citep{liu2025worldmirror} accepts arbitrary combinations of pose, intrinsic, and depth prompts and jointly predicts cameras, depth, pointmaps, normals, and 3D Gaussians, while OmniVGGT~\citep{peng2025omnivggt} injects variable numbers of these modalities into VGGT through a GeoAdapter and randomized modality combinations.

Other works have also explored the direct use of pretrained monocular depth estimation models~\citep{Ranftl_2021_ICCV,depth_anything_v1,depth_anything_v2,Piccinelli_2024_CVPR} for single-image 3D reconstruction. Flash3D~\citep{szymanowicz2024flash3d} achieves this by extending a monocular depth estimation model to a full 3D reconstructor. It first predicts an initial layer of 3D Gaussians at the estimated depth and then intelligently adds subsequent layers to reconstruct occluded parts of the scene. Similarly, Niagara~\citep{wu2025niagara} enhances single-view reconstruction by integrating both monocular depth and normal estimates as inputs using a geometric affine field and 3D self-attention to capture finer geometric details. Furthermore, JointSplat~\citep{xiao2025jointsplat} proposes a probabilistic joint optimization that fuses pixel-level optical flow and depth information to improve feed-forward 3DGS.

\textbf{Takeaway.} Monocular priors complement weak correspondence evidence, but their usefulness depends on how they are reconciled with the observed views~\citep{xu2024depthsplat}. Lightweight distillation can transfer geometric detail~\citep{ren2025fin3r}; the resulting predictor still inherits the limitations of the teacher where image evidence cannot correct them.

\subsection{Model Efficiency}
\label{sec:model_efficiency}

\begin{figure*}
    \centering
    \includegraphics[width=\linewidth]{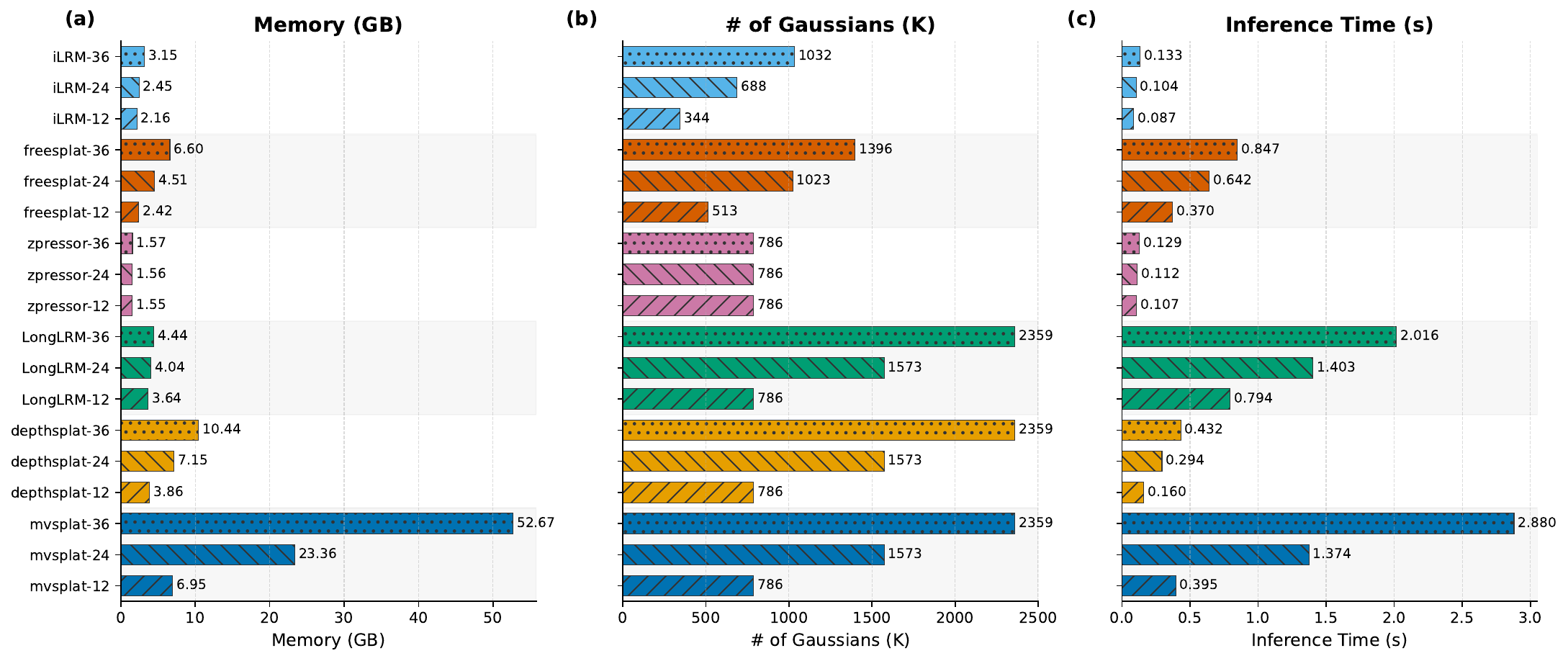}
    \caption{
Efficiency comparison of different novel view synthesis (NVS) methods across varying numbers of input views (12, 24, and 36).
Subplots (a)--(c) illustrate memory consumption, the number of Gaussians, and inference time, respectively.
Color encodes the reconstruction method, while hatch patterns indicate the number of input views.
All values were measured by the survey authors on RealEstate10K under a unified profiling setting on a single NVIDIA A100 GPU, with input images resized to $256\times256$, using the publicly released code and checkpoints for every method.
}
\label{fig:eff_comp}
\end{figure*}

Existing 3D reconstruction methods either depend on slow per-scene optimization or require heavy, generalizable models, making them unsuitable for real-time applications and memory-limited settings. \cref{fig:eff_comp} compares the efficiency of representative NVS methods under different input-view settings. Recent research addresses these bottlenecks from two directions: improving feature efficiency and compacting the explicit 3D representation.

\subsubsection{Feature Efficiency}
\label{sec:feat_eff}
Along the feature axis, recent methods pay more attention to learning where and how to aggregate multi-view information in order to minimize unnecessary computations. ENeRF~\citep{lin2022efficient} introduces a learned depth-guided sampler, significantly reducing the number of per-ray queries for interactive free-viewpoint video. ProNeRF~\citep{bello2024pronerf} further proposes project-aware sampling on this basis, predicting a small number of informative ray samples through a dedicated head and dynamically adjusting opacities. During its training process, the exploration-exploitation schedule is adopted to achieve a balance between global scene coverage and local detail. TinySplat~\citep{song2025tinysplat} focuses on training-free perceptual and spatial compression pipelines, enabling lightweight networks to run on compacted input. ZPressor~\citep{wang2025zpressor} introduces IB-guided anchor-support cross-attention, fusing dense view features into a compact latent Z, and it can be decoded by any 3DGS head. Long-LRM~\citep{ziwen2024long} combines token merging with a hybrid Mamba2~\citep{dao2024transformers}--Transformer~\citep{vaswani2017attention} backbone to handle approximately 250k tokens from 32 views, then decodes to obtain per-pixel Gaussians. However, iLRM~\citep{kang2025ilrm} completely decouples images from representations. The low-resolution viewpoint tokens are iteratively updated through per-view cross-attention and global self-attention, and finally decoded into Gaussians. In other words, the model learns an update rule for a compact state.

Recent work focuses on the VGGT~\citep{wang2025vggt} series, improving feature efficiency through token merging, post-training quantization, block-sparse attention, or KV-cache budgeting, thereby enabling real-time, memory-constrained deployment.
Beyond its foundation-model role (\S\ref{sec:prior_foundation}), VGGT-$\Omega$~\citep{wang2026vggtomega} introduces register attention, which restricts selected cross-frame interactions to compact scene registers. Together with its simplified dense prediction head, these architectural changes reduce training GPU memory to approximately 30\% of that required by VGGT, enabling substantially larger-scale training. More specifically, in selected global blocks only learned scene registers perform cross-frame self-attention; subsequent frame-wise attention returns global information to every frame while retaining all patch tokens. The authors attribute about 23\% of backbone FLOPs and 16\% of backbone training-memory savings to this register-attention change. Importantly, the approximately 70\% full-model training-memory reduction should not be ascribed to register attention alone: it also depends on the unified dense head and efficient MLP/pixel-shuffle decoding.
To solve the bottleneck from a geometric angle, FastVGGT~\citep{shen2025fastvggttrainingfreeaccelerationvisual} implements training-free token merging in the global attention of VGGT, reducing redundant inter-frame interactions by retaining only the first frame or dominant tokens. QuantVGGT~\citep{feng2025quantized} applies post-training quantization, compressing feed-forward VGGT, reducing memory and latency in real-time resource-constrained deployments with minimal accuracy loss. Sparse VGGT~\citep{wang2025faster} uses adaptive block-sparse kernels instead of dense global attention and takes advantage of cross-view sparsity to accelerate while retaining accuracy and improving scalability. Evict3R~\citep{mahdi2025evict3r} improves StreamVGGT~\citep{zhuo2025streaming4dvisualgeometry} with a training-free KV cache eviction strategy. It enforces the memory budget limit for each layer according to attention importance while still retaining the long-term context. LiteVGGT~\citep{shu2025litevggt} boosts vanilla VGGT with geometry-aware cached token merging, substantially reducing runtime and memory on long image sequences. Speed3R~\citep{ren2026speed3r} replaces dense attention with a sparse dual-branch design that focuses computation on informative tokens for faster large-scale reconstruction. SR3R~\citep{feng2026sr3r} reformulates 3D super-resolution as a feed-forward mapping from sparse low-resolution views to high-resolution 3D Gaussian Splatting.

Other recent systems differ in whether they remove redundancy from inputs, attention, repeated depth, or numeric precision. Diversity-aware view partitioning reduces repeated visual evidence before inference, whereas FlashVGGT and SAF3R compress attention through descriptors or adaptive sparsity~\citep{park2026viewpartition,wang2025flashvggt,deng2026saf3r}. Déjà View reuses a single Transformer block and makes iteration count a compute control, while QVGGT combines mixed precision with token filtering and task-aware scale search~\citep{burzio2026dejaview,pan2026qvggt}. Input pruning requires no internal model redesign, but its gains depend on scene coverage. Attention and precision methods preserve more evidence yet add selection policies whose errors can accumulate. This line has therefore converged on adaptive allocation rather than uniform compression, with no single mechanism dominating across view redundancy, scene difficulty, and hardware.

Sequence-specific state compression provides another efficiency axis: ZipMap~\citep{jin2026zipmap} stores an image collection in a compact test-time-trained hidden state, while VGG-T$^3$~\citep{elflein2026vgg} distills a variable-length scene into a fixed-size MLP; both scale linearly with the number of views but belong to the test-time-adaptation regime discussed in \S\ref{sec:visual_refine}.
The recent VGGT efficiency line splits into token reduction and state management. AVGGT, HTTM, and HeSS alter where attention is spent by restructuring global blocks, merging head-wise temporal tokens, or redistributing sparsity across heads~\citep{sun2025avggt,wang2025httm,kim2026hess}. STAC and InfiniteVGGT instead bound the temporal cache, which preserves causal coverage but makes retention policy a source of reconstruction error~\citep{wang2026stac,yuan2026infinitevggt}. Long-LRM++ preserves fine detail over many high-resolution views through a semi-explicit state, showing that efficient scaling must retain local information rather than only reduce tokens~\citep{ziwen2025longlrm}.

\textbf{Takeaway.} Efficiency gains depend on which information can be compressed without losing geometry. Token reduction and compact cross-frame registers target different sources of redundancy~\citep{shen2025fastvggttrainingfreeaccelerationvisual,wang2026vggtomega}; their usefulness depends on view overlap, input resolution, and context length, so latency alone is an incomplete comparison.


\subsubsection{Representation Compaction}
\label{sec:rep_compact}
On the representation side, a parallel line of research focuses on explicit Gaussian compaction. GGN~\citep{zhang2024gaussian} performs message passing over a learned Gaussian graph and pools groups to merge and prune splats. PixelGaussian~\citep{fei2024pixelgaussian} adapts both distribution and count: a cascade adapter involving keypoint scoring and deformable attention guides pruning and splitting, while an image--Gaussian refiner refines the retained primitives. FreeSplat++~\citep{wang2025freesplat++} targets whole-scene inputs with pixel-wise triplet fusion to deduplicate overlap and a weighted floater-removal mechanism that adjusts opacities from multi-view depth consistency; LongSplat~\citep{huang2025longsplat} adds identity-aware redundancy compression in GIR space to bound counts online. Notably, `feature-efficient' encoders (\textit{e.g.}, Long-LRM and iLRM) also apply basic count controls, such as opacity regularization, confidence masks, or post-hoc pruning, to keep Gaussian numbers bounded. Geometry-only representations move rather than remove this bottleneck: occupancy and SDF models store a compact continuous field but incur dense 3D queries and surface extraction, whereas direct mesh predictors such as MeshLRM and MeshFormer deliver graphics-ready assets whose rendering and simulation costs scale with triangle count, at the cost of topology construction and high-resolution detail. A fair efficiency comparison must therefore report field-query or extraction time, triangle or primitive count, memory, and input-view count together instead of treating FPS or PSNR as representation-neutral measures.
Primitive allocation now spans both feature-side compression and structural decoding. The Z-Order Transformer and MuSASplat reduce computation before dense primitive generation through sparse ordering or lightweight multi-scale adaptation~\citep{wang2026zordertransformer,xu2025musasplat}. Off The Grid, SparseSplat, AdaptiveSplat, and SubSplat instead change where primitives are instantiated by using off-grid detection, structure-adaptive maps, texture-aware allocation, or sub-pixel reparameterization~\citep{moreau2025offgrid,zhang2026sparsesplat,singhal2026adaptivesplat,lee2026subsplat}. Globally sparse layouts reduce redundancy, while local densification preserves high-frequency detail at a higher primitive count.

\textbf{Takeaway.} Compaction must distinguish redundant overlap from primitives needed for geometric detail~\citep{zhang2024gaussian,fei2024pixelgaussian,wang2025freesplat++}. Lower primitive counts reduce storage and rendering work, but aggressive merging can discard thin structures or boundaries; feature compression and output compaction therefore address different computational costs.


\subsection{Augmentation Strategies}
\label{sec:data_visual_refine}
\begin{figure}
    \centering
    \includegraphics[width=0.95\linewidth]{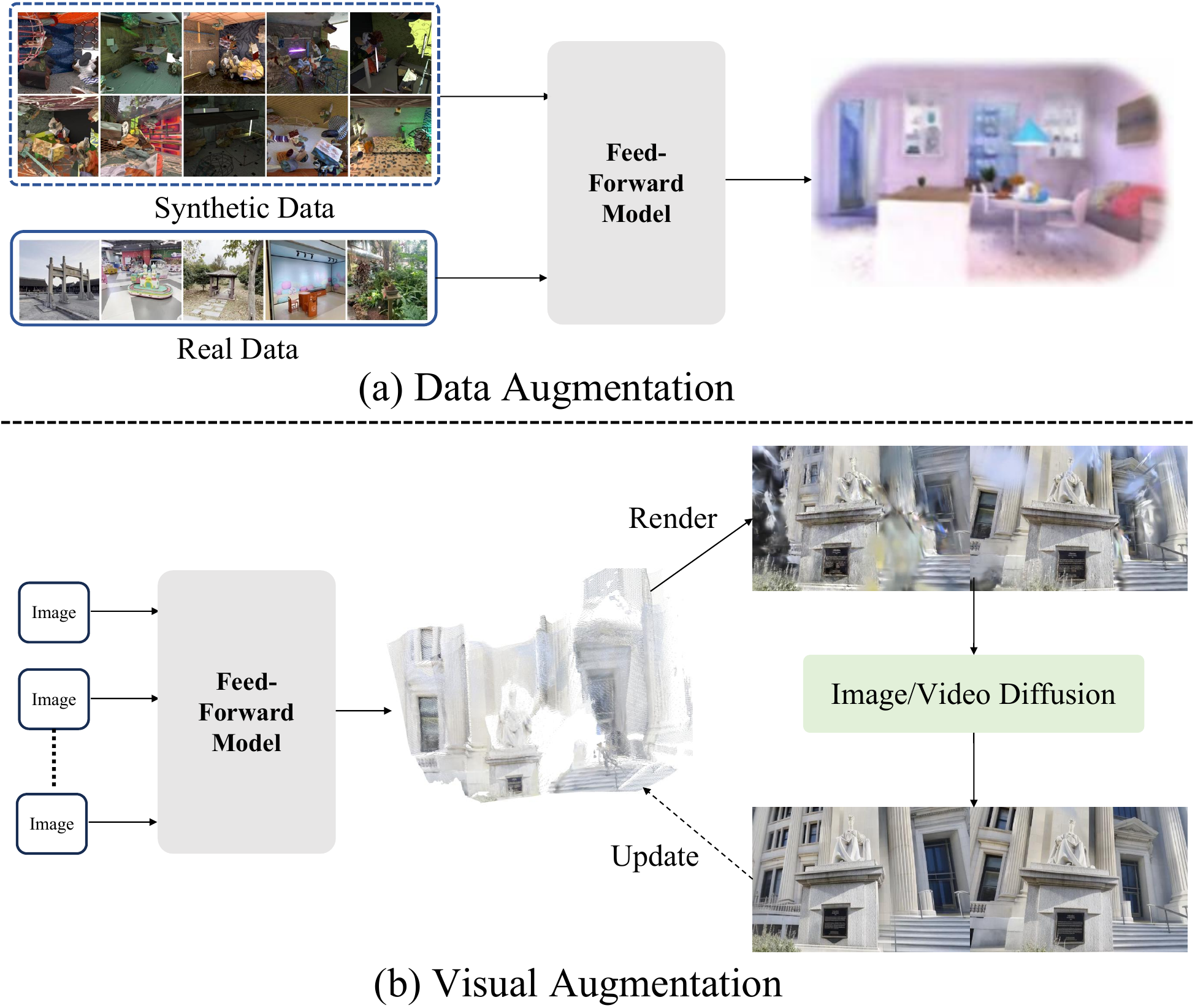}
    \caption{\textbf{Comparison between data and visual augmentation.} (a) Data Augmentation: Combines synthetic and real data to train the feature extractor and 3D predictor. (b) Visual Augmentation: Refines rendered scenes using image or video diffusion to improve visual realism. Notably, modifying 3D parameters is optional.}
    \label{fig:data_visual_augment}
\end{figure}

Neural rendering methods such as NeRF and 3DGS have achieved remarkable progress in 3D reconstruction and novel view synthesis, but they remain limited by sparse inputs, inaccurate poses, and insufficient training diversity~\citep{chen2023dbarf, lin2021barf, meuleman2023progressively, park2023camp, truong2023sparf, wang2021nerf, turki2023suds, martin2021nerf}. To address these challenges, recent research has increasingly focused on augmentation strategies that enrich either the data distribution or the visual representation (see Fig.~\ref{fig:data_visual_augment}). Data augmentation expands training corpora with synthetic scenes, novel views, or pseudo-ground-truth signals, improving model generalization. Visual augmentation, in contrast, leverages generative priors to enhance rendered outputs, suppress artifacts, and hallucinate plausible details. Together, these complementary directions provide a stronger foundation for building robust and scalable neural rendering systems. Section~\ref{sec:visual_refine} also distinguishes inference and adaptation regimes across the survey. This comparison includes scene refinement and memory adaptation, which are not themselves forms of visual augmentation. For geometry-only outputs, however, useful augmentation must preserve metric structure: occupancy, SDF, and mesh predictors benefit from watertight shapes and pseudo-labels for occupancy, distance, depth, or normals, whereas radiance fields and 3DGS can additionally exploit generated images and pseudo-views without explicit 3D ground truth. The latter route scales more easily but may improve PSNR while hallucinating an inconsistent surface, so geometry-oriented augmentation also requires Chamfer distance, normal consistency, and topology-sensitive validation.

\subsubsection{Data Augmentation}
\label{sec:data_refine}
Data augmentation addresses limited training coverage rather than test-time ambiguity, and the representative lines of work differ mainly in whether they prioritize quantity or quality. Puzzles~\citep{ma2025puzzles} synthesizes unlimited posed video-depth data from a single image or clip through simulated camera motions and geometric variations. MegaSynth~\citep{jiang2025megasynth} scales this further by procedurally generating hundreds of thousands of non-semantic 3D scenes, showing that low-level geometric diversity alone can already provide robust supervision. Aug3D~\citep{rauniyar2025aug3d} instead targets quality, augmenting outdoor datasets with structure-from-motion-based novel views so that training samples are better conditioned, while MVBoost~\citep{liu2025mvboost} filters the outputs of multi-view generative models through reconstruction-consistency checks to build reliable pseudo-ground truth at scale. The contrast is informative: purely synthetic scale broadens coverage but leaves a domain gap, whereas geometric validation is what keeps the added supervision consistent.
The newest augmentation strategies differ in whether they reduce annotation requirements, broaden environmental variation, or transfer across domains. Reliev3R replaces multi-view geometry annotations and SfM preprocessing with zero-shot monocular relative depth and sparse correspondences~\citep{chen2026reliev3r}. Wild3R broadens capture conditions through the WildCity corpus, while Wat3R combines teacher-student learning with cross-view consistency to adapt an aerial-domain reconstructor to unlabeled underwater video~\citep{furutani2026wild3r,ren2026wat3r}. Weak supervision reduces annotation cost but inherits errors from frozen monocular predictors. Corpus diversity improves robustness to illumination and transient objects without guaranteeing cross-domain geometry, whereas teacher-student consistency can reinforce teacher bias. These methods converge on replacing expensive ground truth with structured consistency signals, although transfer quality remains bounded by the reliability of their priors.
Supervision can also be generated by changing what the learner observes. E-RayZer learns geometry from unlabeled images through self-supervision, while AREA3D uses an active agent to choose views that reduce reconstruction uncertainty~\citep{zhao2025erayzer,xu2025area3d}. These approaches trade annotation cost against the risk that pretext objectives or acquisition policies emphasize geometry that is easy to predict rather than geometry needed downstream.

\textbf{Takeaway.} Synthetic diversity expands geometric coverage, while reconstruction-based filtering improves the reliability of generated supervision~\citep{jiang2025megasynth,liu2025mvboost}. These address different limitations: adding samples does not by itself close the domain gap, and consistency filtering remains dependent on the reconstructor used to judge them.

\subsubsection{Visual Enhancement and Test-Time Adaptation}
\label{sec:visual_refine}

Visual enhancement uses generative priors to complete appearance where observations leave reconstruction ambiguous. Sparse views and inaccurate poses limit scene fitting~\citep{chen2023dbarf,lin2021barf,meuleman2023progressively,park2023camp,truong2023sparf,wang2021nerf,turki2023suds,martin2021nerf}; geometric regularization constrains shape~\citep{niemeyer2022regnerf,somraj2023simplenerf,yang2023freenerf,deng2022depth,roessle2022dense,wang2023sparsenerf,zhu2024fsgs,yu2022monosdf}, while image-conditioned predictors transfer learned scene priors~\citep{zhou2023nerflix,chen2021mvsnerf,lu2024infinicube,ren2024scube,yu2021pixelnerf}. Generative enhancement adds plausible content, whose consistency with the observed scene must still be evaluated.

The cost depends on how the prior is used. GANeRF optimizes a per-scene model~\citep{roessle2023ganerf}; diffusion methods generate views~\citep{gao2024cat3d,guo2023animatediff,yu2024viewcrafter,zhang2025recapture} or guide scene optimization~\citep{gu2023nerfdiff,liu2023zero,warburg2023nerfbusters,wu2024reconfusion,zhou2023sparsefusion}. Deceptive-NeRF and 3DGS-Enhancer instead use enhanced pseudo-observations to refine the scene~\citep{liu2024deceptive,liu20243dgs}. Generating such supervision and fitting it are separate costs from rendering the final representation.

MVSplat360~\citep{chen2024mvsplat360} conditions video diffusion on Gaussian-rendered features, LatentSplat~\citep{wewer2024latentsplat} decodes variational Gaussian features through a generative image network, and ProSplat~\citep{lu2025prosplat} uses one-step diffusion with reference-view and epipolar conditioning. Their enhancement networks need not be fine-tuned on each scene, although diffusion sampling can still require multiple steps. DIFIX3D+~\citep{wu2025difix3d+} illustrates the distinction between output enhancement and scene refinement: its single-step enhancer also supplies cleaned pseudo-views for subsequent scene-specific fitting.
One-step and progressive priors define a second continuum. One-Shot Refiner and Z3D invoke diffusion once to restore appearance or infer hidden depth, keeping cost bounded but limiting correction depth~\citep{dong2026oneshotrefiner,akola2026zeroshotnovel}. SRSplat and StereoSplat+ reuse super-resolved or generated views across stages, gaining detail at a higher and input-dependent cost~\citep{hu2025srsplat,liu2026stereosplat}. One2Scene and Pano3DComposer use compositional generation when a single image leaves most geometry unobserved, which increases explorable coverage but weakens guarantees of fidelity to the captured scene~\citep{wang2026one2scene,qiu2026pano3dcomposer}.

\textbf{Inference and adaptation regimes.} Across these and the reconstruction methods above, the relevant distinctions are the variables updated at test time, the update signal, and the computation required. A fixed pretrained backbone can support direct prediction, recurrent scene updates, or internal memory adaptation; these do not imply the same inference budget.

\emph{Direct and learned recurrent prediction.} GS-LRM~\citep{zhang2024gs} predicts Gaussians directly, whereas iLRM~\citep{kang2025ilrm} iterates over latent features and ReSplat~\citep{xu2025resplat} predicts Gaussian updates from rendering-error feedback. These recurrent predictors keep network parameters fixed and do not require explicit scene-loss gradients, but their cost depends on the number of updates.

\emph{Gradient-guided scene refinement.} G3R~\citep{chen2024g3r} initializes neural Gaussians from a geometric scaffold and computes rendering-loss gradients with respect to the scene representation. A fixed, learned update network uses this feedback to refine the scene iteratively. The adapted variables belong to the scene, not the reconstruction network; both gradient evaluation and update steps contribute to inference cost.

\emph{Model fine-tuning.} CogNVS~\citep{chen2025reconstruct} renders covisible content from a reconstruction and inpaints hidden content with video diffusion, fine-tuning the diffusion model on the test video. Here model parameters are adapted, unlike G3R's scene variables. This adds a video-specific training stage before novel-view generation.

\emph{Memory and fast-weight adaptation.} TTT3R~\citep{chen2025ttt3r} interprets recurrent memory as fast weights and derives confidence-dependent, closed-form state updates while keeping the pretrained backbone frozen. It is a training-free intervention rather than backbone fine-tuning. tttLRM~\citep{wang2026tttlrm} instead updates fast weights through a token-based learning objective and queries them to decode Gaussians. ZipMap and LoGeR also use test-time-trained memory, while VGG-T$^3$ distills scene features into an MLP at inference~\citep{jin2026zipmap,zhang2026loger,elflein2026vgg}. These methods adapt an internal scene memory rather than directly fitting explicit geometry. Comparisons should report the adapted variables, update counts, input-view budget, resolution, total latency, and peak memory.

\textbf{Takeaway.} Generative enhancement can fill missing appearance, but visually plausible outputs do not alone establish fidelity to the captured scene~\citep{chen2024mvsplat360,wu2025difix3d+}. Comparisons should identify whether improvement comes from rendering, scene refinement, or model adaptation, and report the corresponding test-time computation.

\begin{figure*}[t]
    \centering
    \includegraphics[width=0.95\linewidth]{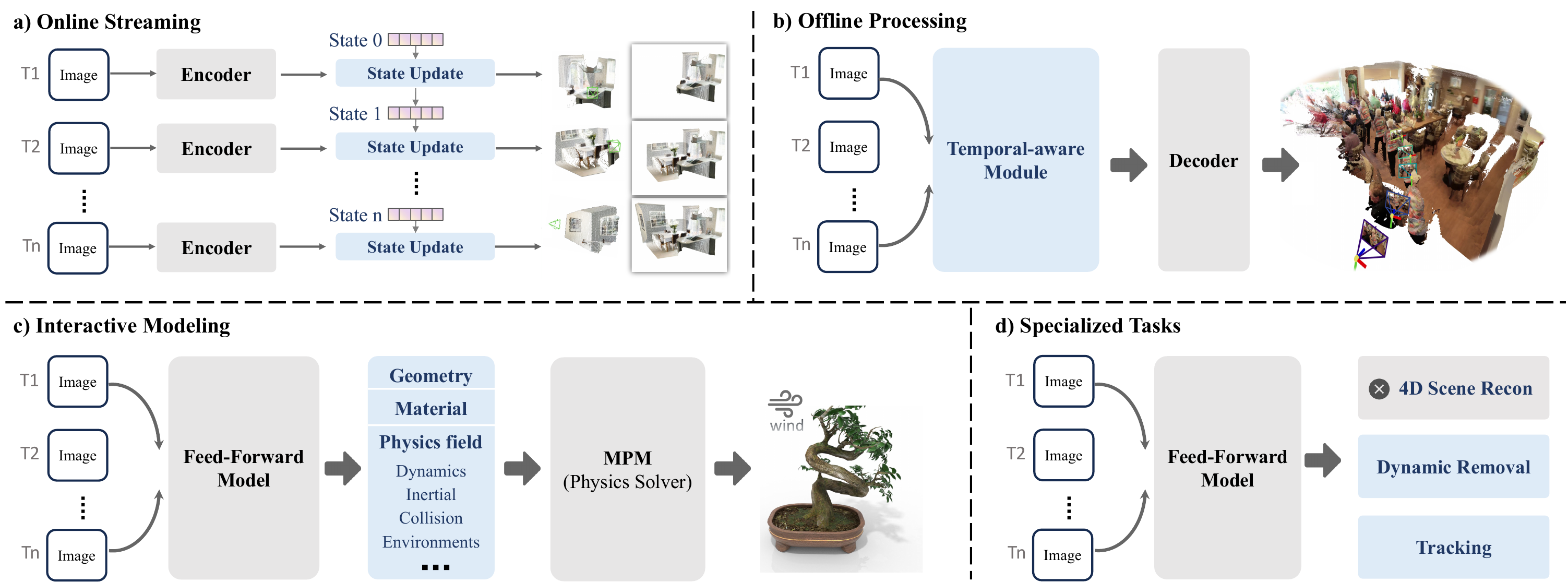}
    \caption{\textbf{Comparison of temporal-aware models.}
    (a) \emph{Online Streaming} models consume streaming observations and maintain a persistent global state for causal, step-by-step reasoning over time. 
    (b) \emph{Offline Processing} models jointly use a clip or window, allowing information from later frames to inform earlier predictions.
    (c) \emph{Interactive Modeling} models predict not only geometry but also material and physics properties to support interactions. 
    (d) \emph{Specialized Tasks} models (\textit{e.g.}, dynamic removal or joint reconstruct-and-track) build on feed-forward backbones but optimize for specific objectives rather than generic full-scene 4D reconstruction.}
    \label{fig:temporal_model}
\end{figure*}

\subsection{Temporal-aware Models}
\label{sec:temporal_aware}

In feed-forward 3D, temporal-aware models enable low-latency 4D scene reconstruction by capturing geometry and motion consistency across frames. As shown in Figure~\ref{fig:temporal_model}, these approaches can be grouped by how they handle time: online streaming models update the scene state per frame for real-time streaming inputs; offline processing models process entire sequences or windows at once to produce globally consistent 4D reconstructions, favoring fidelity over speed; and interactive models build on online backbones with user controls for real-time physics or editing feedback. Other approaches focus on specialized tasks, such as dynamic object removal or multi-view fusion, within a feed-forward pipeline. The output representation changes the temporal constraint: deformable 3DGS attaches motion to renderable primitives, pointmap models must maintain correspondences to avoid identity drift, and occupancy, SDF, or mesh outputs must additionally preserve a coherent surface and topology. Although such explicit surfaces connect directly to collision handling and physical simulation, independent frame-wise extraction can introduce connectivity flicker, self-intersections, or disappearing thin structures. Their evaluation must therefore combine image quality with temporal correspondence, Chamfer distance, and normal consistency; the scarcity of feed-forward temporal surface models remains an open direction.

\subsubsection{Online Streaming}
\label{sec:temporal_online}
Online temporal-aware models process frames sequentially and update the scene representation incrementally, enabling real-time, open-ended 3D and 4D reconstruction. Wu et al.'s StreamSplat~\citep{wu2025streamsplat} estimates per-Gaussian velocities and small deformations from short temporal cues, then updates both motion and appearance in a single pass, requiring no external calibration or test-time optimization. DGS-LRM~\citep{lin2025dgs} pushes this to more deformable content by encoding time with Plücker-ray queries and a spatiotemporal Transformer, predicting deformable Gaussians and 3D scene flow from monocular video at interactive rates. Complementing these Gaussian-motion pipelines, CUT3R~\citep{wang2025continuous} keeps a persistent recurrent state that carries geometry across frames, improving long-horizon consistency and reducing drift and flicker as the scene evolves. Stream3R~\citep{lan2025stream3r} reformulates pointmap prediction as a decoder-only causal Transformer, enabling scalable sequential 3D reconstruction from long image streams. LongStream~\citep{cheng2026longstream} introduces gauge-decoupled streaming visual geometry with keyframe-relative poses and cache-consistent training for stable metric-scale reconstruction over very long sequences.
Song et al.'s distinct StreamSplat~\citep{song2026streamsplat}, a streaming feed-forward 3DGS system, organizes a persistent voxel-aligned causal cache whose cost grows with explored geometry rather than sequence length, and incrementally decodes the cached state into Gaussians.
Long-horizon systems increasingly separate local prediction from global retention. LingBot-Map and TTSA3R use compact anchors or adaptive persistent state, while LASER aligns layer-wise scale and VGGT-Motion adds motion-aware SLAM constraints~\citep{chen2026lingbotmap,zheng2026ttsa3r,ding2025laser,xiong2026vggtmotion}. These choices bound memory differently, and the retained state determines whether failures appear as drift, forgotten geometry, or inconsistent scale. OnlineSplatter and Event3R show that the same causal principle extends to moving objects and asynchronous sensors, but their specialized inputs narrow transfer beyond their capture domains~\citep{huang2025onlinesplatter,huang2026event3r}.

\textbf{Takeaway.} Window-based modeling preserves recent detail, while memory-based modeling carries longer history through a potentially lossy state. Hybrid designs combine local alignment with global retention~\citep{zhang2026loger}. WinT3R also couples online windows with an expanding camera-token pool~\citep{li2025wint3rwindowbasedstreamingreconstruction}, so windowing implies neither offline processing nor constant total memory.

\subsubsection{Offline Processing}
\label{sec:temporal_offline}
Offline temporal-aware models aggregate clips or windows and can use future observations to resolve ambiguity. This access to non-causal context trades buffering and latency for temporal coherence; the complete reconstruction pipeline may still include alignment or refinement after prediction.

L4GM~\citep{ren2024l4gm} extends the idea of LRM~\citep{honglrm} to the temporal domain: given a monocular video, it uses generative priors for the first frame and temporal self-attention to rapidly generate per-frame Gaussians. Furthermore, 4D-LRM~\citep{ma20254d} jointly models space and time, taking a short sequence as input and outputting a 4D Gaussian field that can be queried at any viewpoint and timestamp. Meanwhile, the DUSt3R-style pointmap route has also been extended to dynamic scenes. MonST3R~\citep{zhang2024monst3r} is explicitly built on DUSt3R~\citep{wang2024dust3r} through fine-tuning on dynamic videos. Combined with the stabilization of SEA-RAFT flow~\citep{wang2024sea}, the pairwise pointmaps are aggregated to achieve clip-level reconstruction. Easi3R~\citep{chen2025easi3r} proposes a training-free variant of DUSt3R, which separates static and dynamic content through disentangled attention, restoring both cameras and dense 4D pointmaps simultaneously within the same offline window.

With Gaussians becoming the mainstream dynamic representation, 4DGT~\citep{xu20254dgt} uses a rolling temporal window to directly predict a set of consistent 4D Gaussians, and prunes primitives through density control to enhance efficiency. 4Real-Video-V2~\citep{wang20254real} adopts a similar holistic design, introducing a Gaussian head and dynamic layers on the basis of VGGT. This enables multi-view spatial cues and temporal cues to be integrated in a single pass. To further expand the application scope, MoVieS~\citep{lin2025movies} jointly models appearance, depth, and motion through a shared encoder and three independent heads, populating a Gaussian grid. This method still adopts a window-based rather than streaming approach.

In addition, several task-driven offline methods have been studied for practical application constraints. MonoFusion~\citep{wang2025monofusion} is oriented to sparse-view capture, independently reconstructing the monocular 4D representation for each camera and performing fusion at each time step. EgoMono4D~\citep{yuan2024self} focuses on egocentric video and proposes a self-supervised feed-forward loop to jointly estimate intrinsics, poses, and dense depth at the sequence level. BTimer~\citep{liang2024feed} reconstructs the full 3D snapshot from a single casual video at a specified timestamp, thereby supporting the ``bullet-time'' effect.
Recent offline 4D methods differ in what flexibility they expose at inference. Any4D accepts optional RGB-D, IMU, or radar to recover metric geometry, whereas UFO-4D infers dynamic Gaussians, motion, cameras, and appearance from only two unposed images~\citep{karhade2025any4d,hur2026ufo4d}. 4RC and NeoVerse instead broaden the output space through a reusable spatiotemporal latent or pose-free monocular reconstruction, supporting arbitrary time queries or novel trajectories~\citep{luo20264rc,yang2026neoverse}. Richer sensing reduces geometric ambiguity but complicates capture. Reusable latent states amortize encoding across queries, yet they must preserve motion beyond observed frames. This line has converged on queryable 4D scene states, while sparse-input systems still trade geometric certainty for deployment flexibility.
Recent 4D models differ less in whether they use a Transformer than in what they make persistent. C4D and V-DPM preserve explicit correspondences or dynamic pointmaps~\citep{wang2025c4d,sucar2026vdpm}. PAGE-4D, 4D-VGGT, VGGT4D, MoRe, and OmniX encode motion in learned spatiotemporal states~\citep{zhou2025page4d,wang20254dvggt,hu2025vggt4d,fang2026more,jiang2026omnix}. TrajVG and UniSplat expose trajectories or latent scaffolds as intermediate structure~\citep{miao2026trajvg,shi2025unisplat}. Correspondence-centered states are easier to inspect and track but vulnerable to occlusion, whereas latent states absorb longer context at the cost of weaker geometric interpretability.

\textbf{Takeaway.} Access to future frames helps resolve temporal ambiguity, but requires buffering and delays output~\citep{ren2024l4gm,zhang2024monst3r}. Offline processing is a context choice, not a guarantee of single-pass inference; consistency beyond the processed clip still depends on how separate reconstructions are aligned.

\subsubsection{Interactive Modeling}
\label{sec:temporal_interactive}
Online models update the scene frame by frame in a causal manner; offline models perform batch processing on entire sequences to obtain globally consistent output. Furthermore, interactive models allow users to apply forces, edit content, or adjust materials, and receive immediate, physically plausible feedback through real-time simulation. PIXIE~\citep{le2025pixie} conforms to this paradigm. It first reconstructs geometry and dense visual features (NeRF+CLIP) offline. Subsequently, the material field is predicted in a single feed-forward pass, enabling the physics solver to animate and simulate the scene on demand, thereby achieving rapid interaction on the fixed per-scene geometry. PhysGM~\citep{lv2025physgm} represents a fully amortized method that simultaneously predicts the 3D Gaussian scene and its physical attributes from a single image or sparse views. This approach relies on a single forward pass to completely eliminate the need for per-scene optimization. Because the generated parameters are compatible with physical simulations, the system supports low-latency editing and animation tasks. The framework utilizes a two-stage training process involving supervision and DPO~\citep{rafailov2023direct} to achieve a strong balance between visual realism and computational efficiency.

\textbf{Takeaway.} Interactive reconstruction must provide physical attributes as well as renderable geometry~\citep{le2025pixie,lv2025physgm}. Fast material prediction does not remove the cost of geometry preparation or simulation, so responsiveness should be assessed across the full pipeline and alongside the physical assumptions of the predicted scene.

\begin{table*}[t]
\centering
\small
\caption{\textbf{Representative datasets for feed-forward 3D reconstruction.} These datasets are categorized by their primary purpose as Visual-Oriented, Geometry-Oriented, or Mixed. Each entry includes statistics such as dataset scale, source type, and scene category, with representative methods listed for training or evaluation. Within each section, datasets are ordered by category and release year. Symbols: \iconObj = Objects, \iconIn = Indoor Scenes, \iconOut = Outdoor Scenes, \iconMix = Mixed Scenes. Source: R = Real, S = Synthetic.
}
\label{tab:reorganized_datasets}
\resizebox{\textwidth}{!}{%
\begin{tabular}{l c c c l l}
\toprule
\textbf{Datasets} & \textbf{\#Scenes} & \textbf{Type} & \textbf{Source} & \textbf{Train} & \textbf{Test} \\
\midrule

\multicolumn{6}{c}{\cellcolor{gray!20}\textbf{Geometry-Oriented}} \\

DTU~\citep{jensen2014large} & 124 & \iconObj & R & MVSNeRF, GeoNeRF & DUSt3R, MASt3R, VGGT \\ 
GSO~\citep{downs2022google} & 1{,}030 & \iconObj & R & IBRNet, GS\text{-}LRM, NeuRay & LRM, GRM, Gamma \\ 
ABO~\citep{collins2022abo} & 50K & \iconObj & R & -- & GS\text{-}LRM, LVSM \\ 
OmniObject3D~\citep{wu2023omniobject3d} & 6{,}000 & \iconObj & S & AGG & MeshFormer \\
Objaverse~\citep{deitke2023objaverse} & 818K & \iconObj & S & VGGT, LGM, LRM & LRM \\
WildRGBD~\citep{xia2024rgbd} & 23{,}049 & \iconObj & R & VGGT, AnySplat & -- \\ 

NYUv2~\citep{Silberman:ECCV12} & 464 & \iconIn & R & -- & CATSplat, Flash3D, WorldMirror \\ 
TUM RGBD~\citep{sturm2012evaluating} & 39 & \iconIn & R & -- & FLARE, LoRA3D, VGGT-SLAM \\ 
7-Scenes~\citep{shotton2013scene} & 7 & \iconIn & R & -- & DUSt3R, VGGT, Fast3R \\ 
ScanNet~\citep{dai2017scannet} & 1{,}513 & \iconIn & R & VGGT & DepthSplat, Uni3R \\ 
Matterport3D~\citep{chang2017matterport3d} & 90 & \iconIn & R & -- & Convolutional Occupancy Networks \\ 
Replica~\citep{straub2019replica} & 18 & \iconIn & R & VGGT, SAIL-Recon & MeshSplat, LoRA3D \\ 
Habitat~\citep{savva2019habitat} & 211 & \iconIn & S & DUSt3R, MASt3R, VGGT & -- \\
Hypersim~\citep{roberts2021hypersim} & 461 & \iconIn & S & VGGT, AnySplat & -- \\
ARKitScenes~\citep{baruch2021arkitscenes} & 1{,}661 & \iconIn & R & DUSt3R, MASt3R & PreF3R \\ 
ScanNet++~\citep{yeshwanth2023scannet++} & 1{,}006 & \iconIn & R & DUSt3R, MASt3R & PreF3R, Uni3R \\ 
Hot3D~\citep{banerjee2025hot3d} & 1.5M & \iconIn & R & 4DGT & 4DGT \\ 

Static Scenes 3D~\citep{mayer2016large} & 41K & \iconOut & R & Mono3R & -- \\ 
Waymo~\citep{sun2020scalability} & 110{,}384 & \iconOut & R & DUSt3R, MASt3R & ARTDECO, LoRA3D \\
Virtual KITTI2~\citep{cabon2020virtual} & 5 & \iconOut & S & VGGT & CATSplat, Flash3D \\
TartanAir~\citep{wang2020tartanair} & 1{,}037 & \iconOut & S & RAFT, SEA\text{-}RAFT & MapAnything \\
Spring~\citep{mehl2023spring} & 47 & \iconOut & S & SEA\text{-}RAFT & SEA\text{-}RAFT \\
PointOdyssey~\citep{zheng2023pointodyssey} & 159 & \iconOut & S & VGGT & DGS\text{-}LRM \\
nuScenes~\citep{nuscenes} & 1{,}000 & \iconOut & R & Omni\text{-}Scene, Driv3R & Omni\text{-}Scene, Driv3R \\
Tanks\&Temples~\citep{knapitsch2017tanks} & 21 & \iconMix & R & -- & Long\text{-}LRM, Mono3R \\
ETH3D~\citep{schops2017multi} & 28 & \iconMix & R & -- & DUSt3R, VGGT, Fast3R \\

\midrule
\multicolumn{6}{c}{\cellcolor{gray!20}\textbf{Visual-Oriented}} \\

CelebA~\citep{liu2015faceattributes} & 202{,}000 & \iconObj & R & MetaNeRF & MetaNeRF \\
ShapeNet~\citep{chang2015shapenet} & 51{,}300 & \iconObj & S & -- & pixelNeRF \\
NMR~\citep{kato2018neural} & 44{,}000 & \iconObj & S & -- & SplatterImage, SRT \\
NeRF\text{-}Synthetic~\citep{mildenhall2020nerf} & 8 & \iconObj & S & -- & NeuRay, GNT \\
CO3D~\citep{reizenstein2021common} & 18{,}619 & \iconObj & R & DUSt3R, MASt3R, VGGT & SplatterImage, TripoSR, LaRa \\
MultiShapeNet~\citep{sajjadi2022scene} & 1M & \iconObj & S & RePAST & RePAST, SRT \\
MVImgNet~\citep{yu2023mvimgnet} & 219{,}188 & \iconObj & R & LRM, 4Real\text{-}Video\text{-}V2 & LRM \\
Consistent4D~\citep{jiang2024consistentd} & 7 & \iconObj & S & -- & 4D\text{-}LRM \\
MegaDepth~\citep{li2018megadepth} & 196 & \iconOut & R & DUSt3R, MASt3R, VGGT & -- \\
ACID~\citep{liu2021infinite} & 13{,}047 & \iconOut & R & pixelSplat, DepthSplat & pixelSplat, DepthSplat \\
ENeRF\text{-}Outdoor~\citep{lin2022efficient} & 3 & \iconOut & R & FreeTimeGS & FreeTimeGS \\

DAVIS~\citep{perazzi2016benchmark} & 50 & \iconMix & R & CogNVS & -- \\
RealEstate10K~\citep{zhou2018stereo} & 74{,}766 & \iconMix & R & pixelSplat, DepthSplat & pixelSplat, DepthSplat \\
YouTube\text{-}VOS~\citep{xu2018youtube} & 4{,}519 & \iconMix & R & CogNVS & -- \\
LLFF~\citep{mildenhall2019llff} & 40 & \iconMix & R & IBRNet & NeuRay, WorldForge \\
BlendedMVS~\citep{yao2020blendedmvs} & 113 & \iconMix & S & DUSt3R, MASt3R, VGGT & SparseNeuS, ReTR \\
DyCheck~\citep{gao2022monocular} & 14 & \iconMix & R & -- & DGS\text{-}LRM, Easi3R \\
Neural3DV~\citep{li2022neural} & 6 & \iconMix & R & FreeTimeGS & FreeTimeGS \\
MipNeRF360~\citep{barron2022mip} & -- & \iconMix & R & -- & Feat2GS, WorldForge \\
EgoExo4D~\citep{grauman2024ego} & 5{,}035 & \iconMix & R & 4DGT & 4DGT, MonoFusion \\
DL3DV\text{-}10K~\citep{ling2024dl3dv} & 10{,}510 & \iconMix & R & pixelSplat, DepthSplat & pixelSplat, DepthSplat \\
MiraData~\citep{ju2024miradata} & 330K & \iconMix & S & NutWorld & -- \\

\bottomrule
\end{tabular}
}
\end{table*}

\subsubsection{Specialized Tasks}
\label{sec:temporal_others}
Methods in this category target specific applications rather than universal 4D reconstruction. DAS3R~\citep{xu2024das3r} is oriented towards dynamic removal in static mapping by learning a ``static'' attribute for each Gaussian. It adopts dynamics-aware training to suppress moving objects at rendering time. Thus, a clean and complete static background is reconstructed from dynamic videos. St4RTrack~\citep{st4rtrack2025} integrates reconstruction and tracking into a unified model by predicting world-space pointmaps alongside persistent temporal correspondences. This architecture utilizes a single feed-forward pass to establish geometry and motion without requiring an independent tracking module. By processing these elements together, the framework ensures that point identities remain consistent across a video sequence.
The boundary of temporal reconstruction is also moving toward predictive scene state. DeGO and dynamic-robust photometric-semantic reconstruction separate deforming or transient content from persistent geometry~\citep{gao2026deformablegaussianoc,cai2026dynamicrobustphotome}. SparseWorld-TC, VGGT-World, and WristWorld move from recovering observed frames to forecasting occupancy, geometry tokens, or task-conditioned future views~\citep{du2025sparseworldtc,sun2026vggtworld,qian2025wristworld}. This increases action relevance but mixes reconstruction fidelity with predictive plausibility, so the two outcomes should be evaluated separately.


\textbf{Takeaway.} The treatment of motion follows the task objective: static mapping suppresses moving content, whereas joint reconstruction and tracking preserves its identity~\citep{xu2024das3r,st4rtrack2025}. Their success therefore requires task-specific evaluation of background recovery or correspondence, rather than a shared image-quality ranking.

\section{Datasets and Benchmarks}
\label{benchmark}
\begin{figure*}[t]
    \centering
    \includegraphics[width=0.9\linewidth]{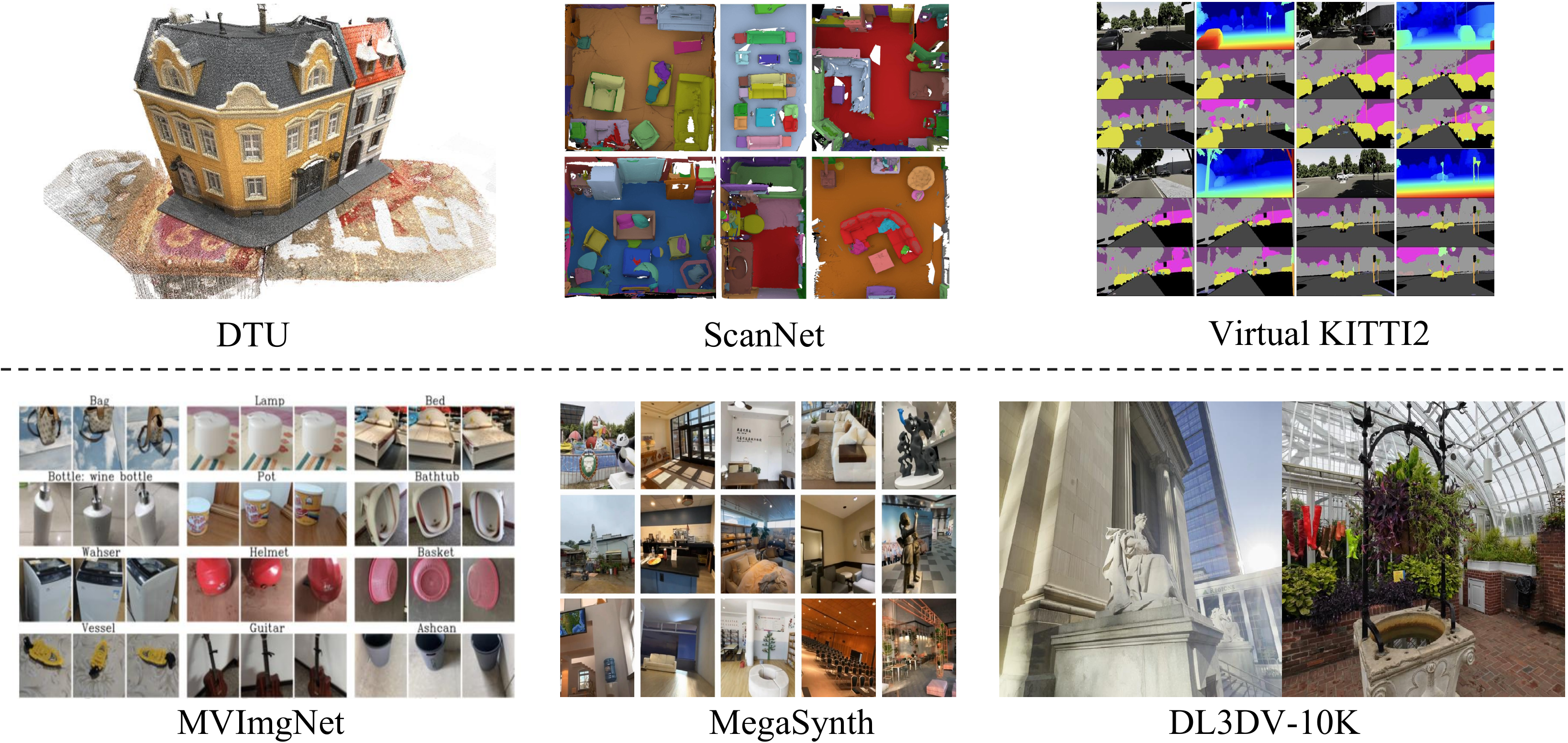}
    \caption{\textbf{Illustration of dataset types.} The upper part presents geometry-oriented datasets, covering objects, indoor scenes, and outdoor scenes. The lower part presents visual-oriented datasets, including objects, synthetic indoor scenes, and mixed indoor and outdoor datasets.}
    \label{fig:vis_datasets}
\end{figure*}

Datasets form the foundation of feed-forward 3D reconstruction and view synthesis. For a comprehensive overview, Table~\ref{tab:reorganized_datasets} summarizes dataset scale, scene categories, real or synthetic sources, and training and test splits. In addition, compared with previous surveys, we introduce a new perspective for categorizing datasets based on whether they are geometry-oriented or visual-oriented, as shown in Fig.~\ref{fig:vis_datasets}. Geometry-oriented datasets provide reliable ground truth 3D representations~\citep{jensen2014large,dai2017scannet,liao2022kitti} such as point clouds, depth, and camera poses, rather than geometry reconstructed from images. These datasets are therefore particularly suitable for tasks in which accurate geometric information is essential. In contrast, visual-oriented datasets are typically sourced from in-the-wild or curated videos and are more appropriate for applications such as novel view synthesis and photorealistic rendering. Incorporating this distinction provides a significant conceptual contribution to the 3D reconstruction community.

\definecolor{best}{RGB}{217,239,212}
\definecolor{second}{RGB}{237,248,233}
\definecolor{third}{RGB}{247,252,245}

\begin{table}[t]
\centering
\caption{\textbf{Results on the DTU 3-view NVS Benchmark.}
Feed-forward 3D reconstruction methods evaluated on the benchmark are listed along with their reported results under the small-baseline setting.
}
\label{tab:dtu_3view}
\setlength{\tabcolsep}{1pt}
\rowcolors{2}{rowgray}{white}
\begin{tabular}{l|ccc}
\toprule
\textbf{Method} & \textbf{PSNR$\uparrow$} & \textbf{SSIM$\uparrow$} & \textbf{LPIPS$\downarrow$} \\
\midrule

pixelNeRF~\citep{yu2021pixelnerfneuralradiancefields} & 19.31 & 0.789 & 0.382 \\
SRF~\citep{chibane2021stereoradiancefieldssrf} & 22.12 & 0.845 & 0.292 \\
IBRNet~\citep{wang2021ibrnetlearningmultiviewimagebased} & 26.04 & 0.917 & 0.190 \\
MVSNeRF~\citep{chen2021mvsnerffastgeneralizableradiance} & 26.63 & 0.931 & 0.168 \\
NeuRay~\citep{liu2022neuralraysocclusionawareimagebased} & 26.47 & 0.875 & 0.158 \\
GeoNeRF~\citep{johari2022geonerfgeneralizingnerfgeometry} & 26.76 & 0.893 & \cellcolor{third}0.150 \\
ENeRF~\citep{lin2022efficient} & \cellcolor{third}27.61 & \cellcolor{second}0.956 & \cellcolor{second}0.091 \\
GPNR~\citep{suhail2022generalizablepatchbasedneuralrendering} & \cellcolor{second}28.50 & 0.932 & 0.167 \\
MuRF~\citep{xu2024murfmultibaselineradiancefields} & \cellcolor{best}28.76 & \cellcolor{best}0.961 & \cellcolor{best}0.077 \\
MatchNeRF~\citep{chen2023explicit} & 26.91 & \cellcolor{third}0.934 & 0.159 \\

\bottomrule
\end{tabular}
\end{table}

\definecolor{best}{RGB}{217,239,212}
\definecolor{second}{RGB}{237,248,233}
\definecolor{third}{RGB}{247,252,245}

\begin{table}[t]
\centering
\caption{\textbf{Results on the RealEstate10K 2-view NVS Benchmark.}
Feed-forward 3D reconstruction methods evaluated on the benchmark are listed along with their reported results under the small-baseline setting.
}
\label{tab:re10k_GS}
\setlength{\tabcolsep}{1pt}
\rowcolors{2}{rowgray}{white}
\begin{tabular}{l|ccc}
\toprule
\textbf{Method} & \textbf{PSNR $\uparrow$} & \textbf{SSIM $\uparrow$} & \textbf{LPIPS $\downarrow$} \\
\midrule
IBRNet~\citep{wang2021ibrnetlearningmultiviewimagebased} & 15.99 & 0.484 & 0.532 \\
pixelNeRF~\citep{yu2021pixelnerfneuralradiancefields} & 20.43 & 0.589 & 0.550 \\
GPNR~\citep{suhail2022generalizablepatchbasedneuralrendering} & 24.11 & 0.793 & 0.255 \\
DBARF~\citep{chen2023dbarf} & 14.79 & 0.570 & 0.490 \\
FlowCAM~\citep{smith2023flowcam} & 18.24 & 0.455 & 0.597 \\
AttnRend~\citep{du2023learningrendernovelviews} & 24.78 & 0.820 & 0.213 \\
CoPoNeRF~\citep{hong2024unifying} & 19.54 & 0.638 & 0.398 \\
MuRF~\citep{xu2024murfmultibaselineradiancefields} & 26.10 & 0.858 & 0.143 \\
pixelSplat~\citep{charatan2024pixelsplat} & 25.89 & 0.858 & 0.142 \\
PixelGaussian~\citep{fei2024pixelgaussian} & 26.72 & - & 0.126 \\
MVSplat~\citep{chen2024mvsplat} & 26.39 & 0.869 & 0.128 \\
LatentSplat~\citep{wewer2024latentsplat} & 25.53 & 0.851 & 0.139 \\
MVSplat360~\citep{chen2024mvsplat360} & 26.41 & 0.869 & 0.126 \\
FreeSplat~\citep{wang2024freesplat} & 26.41 & 0.871 & 0.132 \\
eFreeSplat~\citep{min2024epipolarfree} & 26.45 & 0.865 & 0.126 \\ 
HiSplat~\citep{tang2024hisplat} & 27.21 & 0.881 & 0.117 \\
NoPoSplat~\citep{ye2024no} & 26.79 & 0.878 & 0.124 \\
PF3plat~\citep{hong2024pf3plat} & 23.59 & 0.782 & 0.181 \\
LRM~\citep{honglrm} & 24.78 & 0.820 & 0.213 \\ 
GS-LRM~\citep{zhang2024gs} & \cellcolor{third}28.10 & \cellcolor{third}0.892 & \cellcolor{third}0.114 \\
Long-LRM~\citep{ziwen2024long} & \cellcolor{second}28.54 & \cellcolor{second}0.895 & \cellcolor{second}0.109 \\
TranSplat~\citep{kim2025transplat} & 26.69 & 0.875 & 0.125 \\
DepthSplat~\citep{xu2024depthsplat} & 27.47 & 0.889 & \cellcolor{third}0.114 \\
FLARE~\citep{zhang2025flare} & 23.77 & 0.801 & 0.191 \\
RegGS~\citep{cc2025_reggs} & 24.27 & - & - \\
ProSplat~\citep{lu2025prosplat} & 23.58 & 0.809 & 0.136 \\
TinySplat~\citep{song2025tinysplat} & 27.43 & 0.889 & 0.119 \\
ReSplat~\citep{xu2025resplat} & \cellcolor{best}29.75 & \cellcolor{best}0.912 & \cellcolor{best}0.100 \\
\bottomrule
\end{tabular}
\end{table}

\definecolor{best}{RGB}{217,239,212}
\definecolor{second}{RGB}{237,248,233}
\definecolor{third}{RGB}{247,252,245}

\begin{table*}[t]
\centering
\caption{Pointmap Estimation on 7-Scenes~\citep{shotton2013scene}, NRGBD~\citep{azinovic2022neural}, and ETH3D~\citep{schops2017multi}. Metrics: Acc.$\downarrow$ (pred$\to$GT nearest-point distance, smaller is better), Comp.$\downarrow$ (GT$\to$pred nearest-point distance, smaller is better), NC$\uparrow$ (normal consistency, larger is better), Overall$\downarrow$ = (Acc.+Comp.)/2. The top part reports results under the sparse-view setting using 3 input views, while the bottom part reports results under the dense-view setting using 10 input views.}
\label{tab:dtu_7scenes_nrgbd_eth3d}

\rowcolors{2}{rowgray}{white}
\resizebox{\textwidth}{!}{
\begin{tabular}{l*{10}{c}}
\toprule
\multirow{2}{*}{\textbf{Method}} &
\multicolumn{3}{c}{\textbf{7-Scenes}} &
\multicolumn{3}{c}{\textbf{NRGBD}} &
\multicolumn{4}{c}{\textbf{ETH3D}} \\
\cmidrule(lr){2-4}\cmidrule(lr){5-7}\cmidrule(lr){8-11}
& Acc.$\downarrow$ & Comp.$\downarrow$ & NC$\uparrow$
& Acc.$\downarrow$ & Comp.$\downarrow$ & NC$\uparrow$
& Acc.$\downarrow$ & Comp.$\downarrow$ & NC$\uparrow$ & Overall$\downarrow$ \\
\midrule
DUSt3R~\citep{wang2024dust3r} & 0.786 & 1.748 & \cellcolor{third}0.597 & \cellcolor{second}0.453 & 2.520 & \cellcolor{second}0.515 & \cellcolor{best}0.496 & 9.567 & \cellcolor{best}0.563 & 5.032 \\
MASt3R~\citep{leroy2024grounding} & \cellcolor{second}0.330 & \cellcolor{second}0.754 & 0.557 & 0.749 & \cellcolor{third}1.906 & \cellcolor{third}0.511 & 0.739 & 8.552 & 0.507 & 4.645 \\
Spann3R~\citep{wang20243d} & 0.672 & 1.648 & 0.591 & 0.496 & 2.555 & 0.503 & 0.874 & 9.339 & 0.532 & 5.107 \\
Fast3R~\citep{yang2025fast3r} & 0.864 & 1.819 & 0.564 & \cellcolor{third}0.478 & 2.535 & 0.490 & 0.872 & 9.392 & 0.535 & 5.132 \\
$\pi$3~\citep{wang2025pi} & 0.927 & 1.392 & 0.520 & \cellcolor{best}0.355 & \cellcolor{best}1.514 & \cellcolor{best}0.522 & 0.757 & \cellcolor{third}8.436 & \cellcolor{second}0.548 & \cellcolor{third}4.596 \\
VGGT~\citep{wang2025vggt} & \cellcolor{third}0.397 & 0.885 & \cellcolor{second}0.611 & 0.511 & 2.120 & 0.490 & \cellcolor{second}0.624 & \cellcolor{second}8.386 & \cellcolor{third}0.543 & \cellcolor{second}4.505 \\
MapAnything~\citep{keetha2025mapanything} & 0.927 & \cellcolor{best}0.605 & \cellcolor{best}0.617 & 1.797 & \cellcolor{second}1.709 & 0.510 & 3.266 & \cellcolor{best}5.151 & 0.512 & \cellcolor{best}4.208 \\
Depth Anything 3~\citep{depthanything3} & \cellcolor{best}0.283 & \cellcolor{third}0.789 & 0.585 & 0.485 & 1.997 & 0.489 & \cellcolor{third}0.660 & 8.551 & 0.531 & 4.605 \\
\midrule
DUSt3R~\citep{wang2024dust3r} & 0.745 & 1.843 & 0.562 & \cellcolor{third}0.306 & 2.452 & 0.492 & \cellcolor{best}0.304 & 9.427 & 0.492 & 4.866 \\
MASt3R~\citep{leroy2024grounding} & \cellcolor{second}0.289 & \cellcolor{second}0.800 & \cellcolor{third}0.583 & 0.367 & \cellcolor{third}1.636 & 0.502 & 1.068 & 7.922 & 0.488 & 4.495 \\
Spann3R~\citep{wang20243d} & 0.669 & 1.698 & 0.568 & \cellcolor{best}0.241 & 2.396 & \cellcolor{best}0.553 & 0.874 & 9.339 & \cellcolor{second}0.532 & 5.107 \\
Fast3R~\citep{yang2025fast3r} & 0.804 & 1.870 & 0.549 & 0.319 & 2.445 & 0.482 & 0.807 & 9.270 & \cellcolor{third}0.516 & 5.038 \\
$\pi$3~\citep{wang2025pi} & 0.687 & 1.050 & 0.523 & \cellcolor{second}0.292 & \cellcolor{second}1.496 & \cellcolor{second}0.523 & \cellcolor{third}0.622 & \cellcolor{third}7.893 & 0.493 & \cellcolor{third}4.258 \\
VGGT~\citep{wang2025vggt} & \cellcolor{third}0.413 & 0.973 & 0.581 & 0.350 & 1.905 & 0.469 & \cellcolor{second}0.567 & \cellcolor{second}7.875 & \cellcolor{best}0.537 & \cellcolor{second}4.221 \\
MapAnything~\citep{keetha2025mapanything} & 0.821 & \cellcolor{best}0.588 & \cellcolor{second}0.592 & 1.537 & \cellcolor{best}1.367 & \cellcolor{third}0.510 & 3.774 & \cellcolor{best}4.123 & 0.514 & \cellcolor{best}3.949 \\
Depth Anything 3~\citep{depthanything3} & \cellcolor{best}0.278 & \cellcolor{third}0.851 & \cellcolor{best}0.601 & 0.378 & 1.818 & 0.490 & \cellcolor{third}0.622 & 7.906 & 0.505 & 4.264 \\
\bottomrule
\end{tabular}
}
\end{table*}

In feed-forward 3D reconstruction and view synthesis scenarios, multiple metrics are usually adopted for reliable evaluation. For novel view synthesis, common metrics include PSNR (Peak Signal-to-Noise Ratio), SSIM (Structural Similarity Index)~\citep{wang2004image}, and LPIPS (Learned Perceptual Image Patch Similarity)~\citep{zhang2018unreasonable}. They jointly measure the quality of generated images from different viewpoints. In terms of camera pose evaluation, RTA (Relative Translation Accuracy), RRA (Relative Rotation Accuracy), and AUC (Area Under Curve) are commonly used choices. Among them, RTA and RRA respectively report the relative angular errors of translation and rotation between image pairs; AUC, on the other hand, comprehensively reflects the overall performance by calculating the area under the accuracy curve under different angle error thresholds. For pointmap evaluation, common metrics include point cloud accuracy (precision), completeness (recall), and Chamfer distance. Accuracy calculates the average distance of the nearest neighbors from each predicted point to its corresponding ground-truth surface location, which is used to measure the precision of the prediction. Completeness calculates the average distance of nearest neighbors from ground-truth points to predicted reconstruction, reflecting the completeness of surface coverage. The Chamfer distance integrates measures of both accuracy and completeness. In the dynamic point tracking task, common metrics include OA (Occlusion Accuracy), $\sigma_{\mathrm{avg}}^{vis}$, and AJ (Average Jaccard)~\citep{doersch2022tap}. Among them, Occlusion Accuracy measures the binary accuracy of occlusion prediction; $\sigma_{\mathrm{avg}}^{vis}$ represents the proportion of points that are correctly tracked within the given pixel threshold; AJ integrates occlusion and prediction accuracy into a comprehensive score.

To provide a clear overview of state-of-the-art feed-forward 3D reconstruction and view synthesis methods, we compile results across multiple benchmarks and datasets. The following summarizes three key comparisons, each corresponding to a different evaluation table. For the DTU 3-view benchmark (Table~\ref{tab:dtu_3view}), early NeRF-based methods are typically evaluated on relatively small datasets. In this study, we select the DTU dataset and set the number of input views to three for all methods. The results show that MuRF~\citep{xu2024murfmultibaselineradiancefields} achieves the best performance on this benchmark. For the RealEstate10K benchmark (Table~\ref{tab:re10k_GS}), compared to DTU, RealEstate10K is a substantially larger dataset and provides a more extensive evaluation benchmark. Table~\ref{tab:re10k_GS} compiles reported results for 28 methods. Among the listed entries, ReSplat~\citep{xu2025resplat} achieves the highest PSNR and SSIM and the lowest LPIPS. Finally, for the point cloud pose reconstruction benchmarks (Table~\ref{tab:dtu_7scenes_nrgbd_eth3d}), in addition to novel view synthesis, we evaluate pointmap estimation on 7-Scenes~\citep{shotton2013scene}, NRGBD~\citep{azinovic2022neural}, and ETH3D~\citep{schops2017multi} under sparse-view and dense-view settings with 3 and 10 input views, respectively. All three datasets report Acc., Comp., and NC, while ETH3D additionally reports Overall. The results show that no single model dominates every dataset and metric. In the sparse-view setting, Depth Anything 3 attains the best accuracy on 7-Scenes, $\pi$3 leads all three metrics on NRGBD, and MapAnything obtains the best overall score on ETH3D as well as the best completeness and normal consistency on 7-Scenes. The dense-view setting preserves this ranking only in part: Depth Anything 3 and MapAnything remain the strongest on 7-Scenes and ETH3D, whereas on NRGBD the best accuracy and normal consistency shift to Spann3R~\citep{wang20243d} and the best completeness to MapAnything. Pointmap results should therefore be read per metric and per view budget rather than as a single ranking.


\section{Applications}\label{application}

Across these scenarios, a shared goal is to replace per-scene optimization with single-pass inference that is both scalable and robust under sparse or noisy input.
In autonomous driving, the focus is large-scale dynamic reconstruction with strict real-time and temporal-consistency requirements.
In robotics, fast generalizable 3D representations support downstream decision making, including manipulation and long-horizon navigation with explicit scene memory.
Beyond this, feed-forward 3D priors increasingly serve as a backbone for semantic scene understanding, SfM/SLAM, digital humans, and even geometry-aware video generation.

\subsection{Autonomous Driving}
\label{sec:autonomous_driving}

Autonomous driving poses unique challenges for feed-forward 3D reconstruction, including large-scale dynamic environments, sparse camera coverage, and the need for low-latency, temporally consistent scene representations. Recent work has focused on feed-forward, data-driven architectures that leverage learned priors to enable fast and robust 3D scene reconstruction.

For large-scale static reconstruction, SCube~\citep{ren2024scube} and InfiniCube~\citep{lu2024infinicube} adopt hierarchical voxel-to-Gaussian pipelines to generate city-scale 3D scenes from sparse and partial observations. To model dynamics, STORM~\citep{yang2024storm}, Driv3R~\citep{fei2024driv3r}, DrivingRecon~\citep{lu2024drivingrecon}, and WorldSplat~\citep{zhu2025worldsplat} directly predict 4D Gaussians to capture moving objects and maintain temporal consistency, with WorldSplat additionally integrating a generative diffusion model for novel-view video synthesis.  

To improve robustness under sparse or limited-overlap camera setups, DrivingForward~\citep{tian2025drivingforward}, EVolSplat~\citep{miao2025evolsplat}, and EDUS~\citep{miao2024efficient} leverage geometric priors and depth features for stable single-pass inference. Finally, task-specific pipelines such as BEV-GS~\citep{wu2025bev}, Omni-Scene~\citep{wei2025omni}, and DriveGen3D~\citep{wang2025drivegen3d} enable real-time reconstruction tailored to downstream driving tasks, including road-surface modeling, panoramic scene reconstruction, and dynamic street-level scene generation.
A newer divide is between reconstruction as the output and reconstruction as a task representation. VR-Drive trains reconstruction as an auxiliary signal for control, while Map-Det3D and geometry-grounded unified perception convert streaming scene state into metric boxes, depth, and occupancy~\citep{cho2025vrdrive,yang2026mapdet3d,xu2026geometrygroundedunif}. This improves task alignment but makes geometric quality harder to separate from detector or control performance.

\begin{figure*}[ht]
    \centering
    \includegraphics[width=\textwidth]{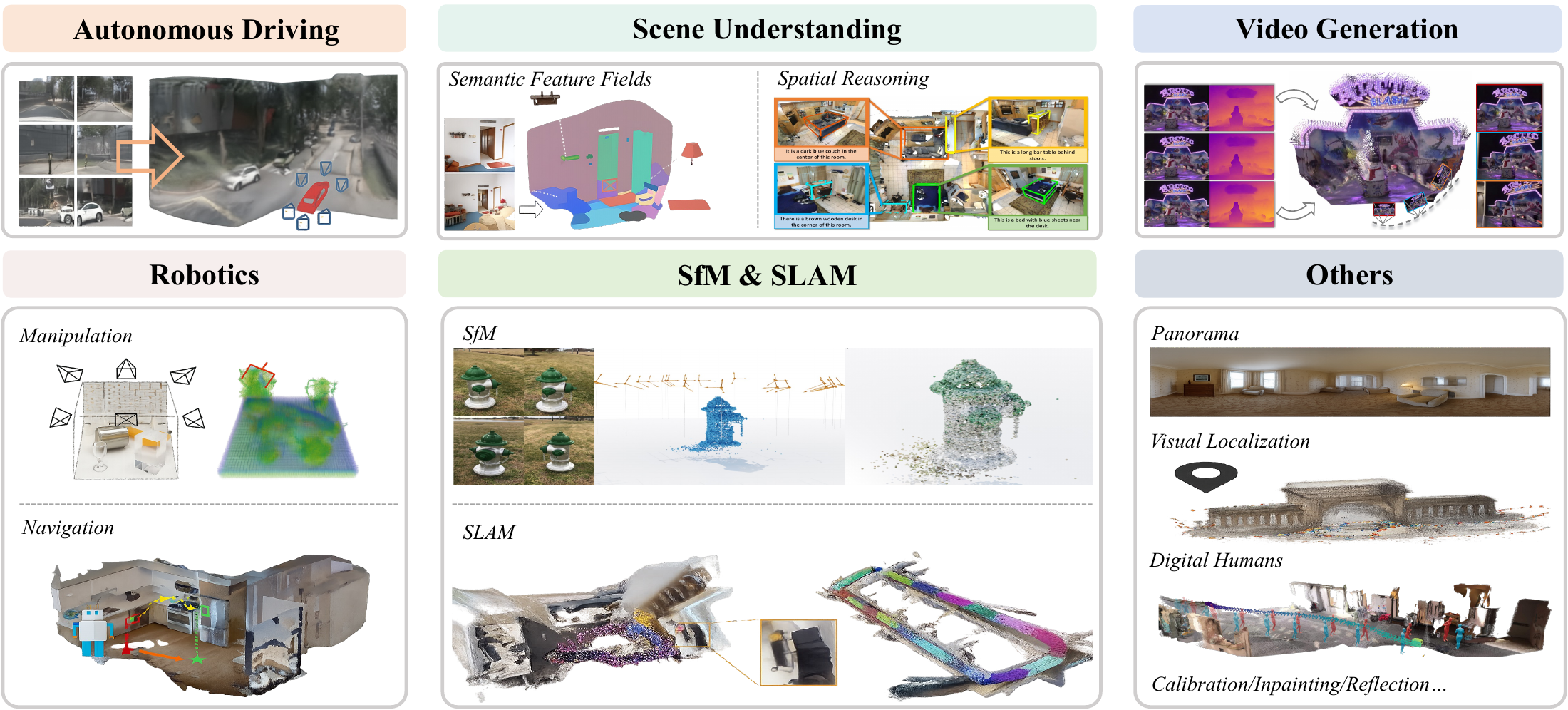}
    \caption{
        Visualization of application scenarios for feed-forward 3D reconstruction. 
        Adapted from prior work~\citep{dai2022graspnerf, guo2025igl, 
        chen2024slgaussian, wang2024vggsfm, zhang2025pansplat, maggio2025vggt, 
        wei2025omni,chen2020scanrefer, chen20254dnex,chen2025human3r}.
    }
    \label{fig:application}
\end{figure*}

\subsection{Robotics}
\label{sec:robotics}

\subsubsection{Manipulation.} 
Recent feed-forward 3D reconstruction methods have enabled fast and generalizable robotic manipulation by providing dense geometry and semantic representations. GraspNeRF~\citep{dai2022graspnerf} predicts TSDFs from multi-view images for efficient 6-DoF grasp detection, including challenging transparent and specular objects. ManiGaussian~\citep{lu2024manigaussian} and ManiGaussian++~\citep{yu2025manigaussian++} leverage feed-forward 3DGS to capture object geometry and dynamics, with ManiGaussian++ employing a hierarchical Gaussian world model to support complex multibody and bimanual operations. GAF~\citep{chai2025gaf}, QGFS~\citep{wang2024query}, GaussianGrasper~\citep{zheng2024gaussiangrasper}, and EmbodiedSplat~\citep{chhablani2025embodiedsplat} extend this paradigm to action reasoning, reinforcement learning, open-vocabulary instruction following, and personalized real-to-sim-to-real navigation, all within a feed-forward 3D framework.

\subsubsection{Navigation.} 
Feed-forward 3D scene representations also enhance robotic navigation by providing large-scale, temporally consistent maps for planning and localization. UnitedVLN~\citep{dai2024unitedvln} integrates 3DGS-based memory for panoramic observation and semantic aggregation, supporting vision-and-language navigation queries. VR-Robo~\citep{zhu2025vr} uses a GS–mesh hybrid memory to combine photorealistic 3DGS rendering with accurate physical simulation for robust sim-to-real transfer. GS-LTS~\citep{fu2025gs} leverages visual-language models (CLIP, SAM) to detect environmental changes and progressively refine 3DGS maps for object-centric navigation and adaptive planning. IGL-Nav~\citep{guo2025igl} constructs 3DGS from monocular video and uses coarse-to-fine matching to enable effective real-time image-goal localization.

\subsection{Scene Understanding}
\label{sec:perception}

\subsubsection{Semantic Feature Fields}

Feed-forward 3D reconstruction has recently enabled efficient integration of vision-language semantics into 3D representations, supporting open-vocabulary, part-aware, and temporally consistent 3D scene understanding without per-scene optimization. Semantic Gaussians~\citep{guo2024semantic}, SemanticSplat~\citep{li2025semanticsplat}, and UniForward~\citep{tian2025uniforward} embed 2D semantic features into 3D Gaussians while jointly reconstructing geometry and appearance. 

To improve generalization from sparse or unposed views, several methods adopt different strategies. SLGaussian~\citep{chen2024slgaussian} and SegMASt3R~\citep{jayanti2025segmast3r} leverage multi-view segmentation consistency, GSemSplat~\citep{wang2024gsemsplat} and LSM~\citep{fan2024large} employ feature aggregation and Transformer-based encoding, and PartField~\citep{liu2025partfield} and AlignGS~\citep{gao2025aligngs} focus on hierarchical or semantic-to-geometry regularization. Together, these works demonstrate how feed-forward pipelines efficiently construct semantically coherent 3D representations from limited 2D observations.
Recent systems split between attaching semantics to a persistent reconstructed state and using geometry as a task prior. MoonSeg3R and G2TAM add segmentation or prompted tracking to a reconstructed scene memory~\citep{du2025moonseg3r,zhu2026g2tam}. VGGT-Det, MV2GF, and glass-surface detection instead lift visual features through geometric predictions before task-specific decoding~\citep{cao2026vggtdet,yamane2026mv2gf,lu2026glasssurfacedetectio}. The first route supports joint state updates but couples errors across tasks, while the second transfers more easily but remains dependent on the reliability of the frozen geometric prior.



\subsubsection{Spatial Reasoning}

Feed-forward 3D methods are increasingly used to provide MLLMs with spatial awareness. Many approaches implicitly incorporate geometric priors from visual–geometry foundation models (e.g., VGGT~\citep{wang2025vggt}, CUT3R~\citep{wang2025continuous}) into the visual encoder to enhance structural reasoning and multi-view consistency. Representative works include 3DRS~\citep{huang2025mllmsneed3dawarerepresentation}, which distills rich 3D priors from VGGT, Spatial-MLLM~\citep{wu2025spatialmllmboostingmllmcapabilities}, which adds a VGGT-initialized spatial branch with space-aware frame sampling, VG-LLM~\citep{zheng2025learningvideos3dworld}, which fuses geometric features with visual tokens at the patch level, and VLM3R~\citep{fan2025vlm}, which replaces the geometric backbone with CUT3R for improved egocentric temporal understanding.

Other methods maintain explicit 3D representations to support spatio-temporal reasoning. For example, ST-LLM~\citep{zheng2025spatiotemporalllmreasoningenvironments} aligns egocentric video with offline point clouds and camera poses to provide structured 3D cues for temporal reasoning. Together, these approaches illustrate how feed-forward 3D pipelines can enhance both implicit and explicit spatial reasoning in vision-language models.
Geometry can also act as explicit supervision rather than only a latent feature. GAP-MLLM jointly predicts sparse pointmaps and semantic labels during multimodal pre-training, which strengthens 3D grounding but makes geometric accuracy compete with language objectives~\citep{zhang2026gapmllm}.

\subsection{SfM and SLAM}
\label{sec:slam}

Reconstructing 3D scene geometry and camera motion from images is a core challenge in computer vision and robotics. Feed-forward and differentiable architectures are increasingly used to unify Structure from Motion (SfM) and Simultaneous Localization and Mapping (SLAM), enabling direct prediction of geometry and camera poses without per-scene optimization.

For SfM, differentiable feed-forward pipelines such as VGGSfM~\citep{wang2024vggsfm} and Light3R-SfM~\citep{elflein2025light3r} replace traditional incremental optimization with end-to-end architectures that jointly infer geometry and camera parameters from multiple views, providing scalable and efficient global reconstruction.

For SLAM, end-to-end feed-forward approaches such as MASt3R-SLAM~\citep{murai2025mast3r}, SLAM3R~\citep{liu2025slam3r}, and VGGT-SLAM~\citep{maggio2025vggt} reconstruct local geometry and maintain global map consistency in real time without explicitly solving internal parameters. Advanced pipelines including ARTDECO~\citep{li2025artdeco}, EC3R-SLAM~\citep{hu2025ec3r}, MASt3R-Fusion~\citep{zhou2025mast3r}, and ViSTA-SLAM~\citep{zhang2025vista} further integrate multi-sensor fusion, hierarchical Gaussian decoders, or factor-graph optimization to achieve robust global alignment and high-fidelity mapping over long, complex sequences.

Collectively, these works illustrate the convergence of SfM and SLAM toward unified, differentiable, feed-forward 3D perception systems, combining global reconstruction accuracy with real-time mapping efficiency and robust trajectory estimation.

\subsection{Video Generation}
\label{sec:video_gen}

Video foundation models have advanced rapidly, yet temporal and spatial consistency remain major bottlenecks. Feed-forward 3D reconstruction provides an effective way to inject geometric priors into video generation, improving cross-frame consistency, novel-view realism, and physical plausibility. More broadly, the combination of video generation and feed-forward reconstruction connects pixel-level synthesis with structure-aware modeling, and can be roughly divided into two directions: reconstruction-enhanced video generation, where explicit 3D reconstruction improves video synthesis while the final output remains a video sequence, and video generation-based scene reconstruction, where video generative models serve as priors for producing explicit 3D or 4D scene representations or geometry-aware world models.

\noindent\textbf{Reconstruction-enhanced video generation.}
In this setting, feed-forward 3D reconstruction is used as a structural prior to improve geometric and temporal consistency in video generation. MVSplat360~\citep{chen2024mvsplat360} and JOG3R~\citep{huang2025jog3r} directly integrate reconstruction modules into the generation pipeline by coupling diffusion with 3D Gaussian or SfM-based geometry, thereby improving pose and structural consistency across frames. Other methods emphasize iterative refinement between generation and reconstruction. GenFusion~\citep{wu2025genfusion} and Envision~\citep{liu2025revision} form cyclic pipelines in which generated videos are converted into structured 3D or object-centric representations, refined with geometric or physical priors, and then fed back into the diffusion model to improve viewpoint diversity, reduce artifacts, and enforce motion consistency.

\noindent\textbf{Video generation-based scene reconstruction.}
In this direction, video generative models act as priors for producing explicit dynamic scene representations or geometry-aware world models. Some methods focus more directly on explicit 3D or 4D reconstruction. 4DNex~\citep{chen20254dnex}, Lyra~\citep{bahmani2025lyra}, and ShapeGen4D~\citep{yenphraphai2025shapegen4d} distill geometric knowledge from video diffusion models into structured 4D representations, enabling dynamic scene generation or temporally consistent 4D reconstruction from sparse inputs. Other methods explore geometry-aware control and long-horizon world modeling. Geometry Forcing~\citep{wu2025geometry}, Spmem~\citep{wu2025video}, and SteerX~\citep{park2025steerx} improve viewpoint consistency and structural alignment through geometry-grounded intermediate organization, spatial memory, or inference-time geometric rewards. EvoWorld~\citep{wang2025evoworld}, WorldForge~\citep{song2025worldforge}, and FantasyWorld~\citep{dai2025fantasyworld} further extend this trend toward geometry-aware world modeling, where video generation is coupled with explicit spatial memory, trajectory control, or implicit 3D fields for long-horizon consistent scene evolution.
HY-World 2.0~\citep{team2026hyworld} moves from an individual generator toward an integrated multimodal 3D-world system, combining panoramic generation, trajectory planning, memory-based world expansion, WorldMirror 2.0 reconstruction, and interactive rendering in one pipeline.

\subsection{Others}
\label{sec:other_app}

\noindent\textbf{Panorama.} 
360$^\circ$ panoramic scene reconstruction faces challenges such as wide baselines, severe spherical distortions, and high-resolution rendering demands. Recent feed-forward methods address these issues by leveraging spherical geometry, cost volumes, and Gaussian-based representations for efficient panoramic reconstruction and synthesis. Splatter-360~\citep{chen2025splatter} introduces a generalizable 3D Gaussian splatting framework with spherical cost volumes for wide-baseline panoramic reconstruction, while PanSplat~\citep{zhang2025pansplat} further improves high-resolution 4K panoramic synthesis through hierarchical spherical Gaussian representations, achieving strong rendering quality and memory efficiency. In parallel, PanoVGGT~\citep{guo2026panovggt} proposes a permutation-equivariant Transformer that jointly estimates camera poses, depth, and 3D structure from panoramas in a single forward pass, improving geometric reasoning under spherical distortions.


\noindent\textbf{Localization.}
Recent feed-forward localization methods increasingly move beyond explicit geometric matching toward direct pose prediction and neural scene-based reasoning. Some approaches, such as Reloc3r~\citep{dong2025reloc3r} and FastForward~\citep{barroso2025scene}, directly infer camera poses through relative pose regression or 3D-anchored feature representations, enabling efficient relocalization in a single forward pass. Other methods focus on richer scene correspondence and scene-level representations. Multi-View 3D Point Tracker~\citep{rajivc2025multi} predicts dense 3D correspondences from multi-view features using transformer-based correlation, while SAIL-Recon~\citep{deng2025sail} extends feed-forward scene regression to large-scale SfM by estimating poses from neural scene representations anchored to reference images.

\noindent\textbf{Digital humans.}
Beyond human-centric reconstruction, recent work has begun to explore joint human-scene modeling. Among recent feed-forward methods, Human3R~\citep{chen2025human3r} extends feed-forward reconstruction to unified 4D human-scene modeling by jointly estimating multiple SMPL-X bodies, dense 3D scenes, and camera trajectories in a single pass. This line of work suggests that feed-forward reconstruction can move beyond isolated avatar modeling toward holistic dynamic human-scene understanding.


\noindent\textbf{Calibration, inpainting and reflection.}
Beyond standard reconstruction, recent feed-forward methods have also been applied to related problems such as self-calibration, scene inpainting, and reflection-aware reconstruction, where conventional multi-stage optimization remains expensive under unknown intrinsics, cross-view ambiguity, or missing geometry. LoRA3D~\citep{lu2024lora3d} addresses self-calibration for scene-specific reconstruction, while BevSplat~\citep{wang2025bevsplat} improves cross-view localization through Gaussian-based BEV representations. InstaInpaint~\citep{you2025instainpaint} performs coherent 3D scene completion from 2D proposals via reference-guided feed-forward reconstruction, while Reflect3r~\citep{wu2025reflect3r} treats mirror reflections as virtual views to support single-image reconstruction and pose refinement.


\section{Future Directions}\label{future}

Despite the remarkable progress in feed-forward 3D reconstruction, the field still faces many challenges and opportunities that will shape the next generation of research. This section highlights several promising directions, ranging from fundamental questions about data and representations to broader conceptual developments in the discipline.

\subsection{Rigorous Benchmarks}
\label{sec:benchmark}

In the field of feed-forward 3D reconstruction, an increasing number of benchmarks are being introduced, reflecting the rapid progress of the research community. Nevertheless, notable limitations remain in both the evaluation criteria and the quality of available data. At present, most large-scale benchmarks provide only video sequences~\citep{ling2024dl3dv}, and only a small portion include ground-truth 3D point clouds for quantitative validation. The lack of comprehensive supervision limits the reliability of performance comparisons. In addition, many existing benchmarks do not account for the varying levels of difficulty that result from contextual gaps across different viewpoints, or they rely on fixed input views without variation. These practices reduce the fairness of evaluation, since models may be optimized to take advantage of specific view selections, leading to artificially enhanced performance, as observed in datasets such as RealEstate10K~\citep{zhou2018stereo} and ACID~\citep{liu2021infinite}.

Future benchmarks should emphasize larger-scale and high-fidelity data that provide accurate 3D ground truth together with video sequences. They should also organize difficulty levels according to variations in viewpoint and context, while maintaining standardized evaluation protocols to avoid unfair advantages caused by selective view sampling. These improvements will support more rigorous and equitable comparisons, promoting steady and transparent progress in feed-forward 3D reconstruction research.

\subsection{System Efficiency}
\label{sec:efficiency}

As feed-forward 3D reconstruction models grow in parameter count and feature resolution, higher accuracy comes with increased inference latency and hardware requirements. The combination of multi-view global attention, dense 3D representation, and high-resolution voxels or point pipelines leads to a super-linear increase in computational and memory cost. Under fixed bandwidth and FLOP budgets, the balance among throughput, latency, and accuracy becomes the main constraint of deployment. In large-scale, long-trajectory, and cross-sequence settings, mismatches among VRAM utilization, memory bandwidth and operator scheduling further exacerbate this limitation, reducing efficiency and actual scalability.

Recent studies have increasingly emphasized inference acceleration and model compactness. These efforts include structured sparsity and linear or approximate attention to reduce the cost of multi-view fusion~\citep{lin2022efficient,bello2024pronerf,wang2025zpressor,ziwen2024long,kang2025ilrm,shen2025fastvggttrainingfreeaccelerationvisual}; hierarchical or coarse-to-fine reconstruction to avoid inefficient 3D sampling~\citep{bello2024pronerf, fei2024pixelgaussian}; view selection and redundancy pruning to control input growth~\citep{song2025tinysplat, wang2025freesplat++, huang2025longsplat}; quantization, pruning, distillation, and low-rank or adapter layers, such as LoRA and adapter modules, to compress parameters and activations~\citep{fei2024pixelgaussian,wang2025zpressor,song2025tinysplat}. Future progress can be pursued in three main directions.

\begin{itemize}
    
    \item \textbf{Scalable and efficient reconstruction architectures.} Future progress may rely on architectures that better support large scenes, long input sequences, and high-resolution representations. Promising directions include hierarchical geometric priors, hybrid explicit--implicit representations, level-of-detail pipelines, and occupancy or visibility-aware acceleration.
    
    \item \textbf{Inference and memory optimization.} 
    Improving runtime efficiency remains critical for practical deployment. Techniques such as mixed precision, operator fusion, CUDA graphs, and out-of-core scheduling can help reduce latency, improve memory utilization, and stabilize throughput.

    \item \textbf{Deployment-oriented system design.} 
    Further advances will also require compression and hardware-aware optimization, including quantization, auto-tuned kernels, and heterogeneous execution across edge devices. Standardized benchmarks that jointly measure accuracy, latency, memory, and energy will be important for assessing real-world deployability.


\end{itemize}

\subsection{Scalable Representations}
\label{sec:representation}
The main paradigm is to adapt 3D representations such as NeRF~\citep{mildenhall2020nerf}, 3D Gaussian Splatting~\citep{kerbl20233d}, and explicit meshes from the domain of per-scene optimization. However, the representation optimized for fitting a single scene may not be suitable for predicting the structural content of the scene in a single forward pass with limited inputs. In large-scale settings, this limitation becomes more obvious, because existing feed-forward methods often struggle to maintain global coherence and fine-grained details in the reconstructed results.

There is a clear need for new 3D representations designed specifically for generalizable feed-forward reconstruction. Methods such as LVSM~\citep{jin2025lvsm} directly generate novel views without using an explicit 3D intermediate representation, which indicates a promising direction. Other insights can come from related areas such as video generation, where strong latent representations can often achieve impressive temporal and spatial consistency without explicit 3D geometry. Future representations need to be inherently scalable, which may involve hierarchical or compositional structures that effectively model a wide range of environments while preserving local details, thereby balancing scene scale and reconstruction quality.



\subsection{World Models}
\label{sec:world_model}

Feed-forward 3D reconstruction is increasingly positioned not only as a standalone reconstruction engine, but as a foundational component of world models, i.e., systems that maintain persistent, explorable, and actionable representations of scene state. Two distinct paradigms are emerging along this trajectory. \textit{Video world models} learn implicit world dynamics through video generation and leverage geometric priors for consistency. \textit{3D world models} explicitly construct and maintain structured 3D representations as the canonical scene state. Both paradigms benefit from the efficiency of feed-forward 3D reconstruction, yet they differ fundamentally in how the world state is represented, queried, and acted upon.

\noindent\textbf{Video world models.} Video world models treat the video generation process itself as a world simulator, where the temporal evolution of pixel sequences implicitly encodes scene dynamics, physical interactions, and viewpoint changes. A central theme across recent works~\citep{wu2025geometry,huang2025jog3r,wu2025video,ren2025gen3c,huang2025voyager,zhao2025spatia} is to inject geometric reasoning into the video generation pipeline, whether through explicit pose conditioning, geometry-grounded latent alignment, or persistent spatial memory that records and retrieves 3D scene layouts across generation steps. The scalability of this paradigm is a major strength, as video diffusion models trained on internet-scale data naturally capture complex appearance, lighting, and motion patterns. However, world state in these models remains entangled within high-dimensional latent spaces, making it difficult to perform precise spatial queries, enforce hard physical constraints, or support explicit object-level manipulation.
PERSIST~\citep{garcin2026persist} addresses the long-horizon failure of finite pixel histories by continually updating a latent 3D scene state, giving an interactive world model persistent spatial memory and improved geometric consistency as a trajectory grows.

\noindent\textbf{3D world models.} 3D world models maintain explicit, structured representations of the scene, such as point clouds, 3D Gaussians, meshes, or neural fields, that serve as the queryable and editable world state. Feed-forward 3D reconstruction plays a central role in this paradigm by enabling the real-time construction and progressive updating of such representations. Recent works in this direction~\citep{long2024wonder3d,liang2025wonderland,szymanowicz2025bolt3d,yu2025wonderworld,ni2025wonderturbo,park2025steerx,chen20254dnex,bahmani2025lyra,worldlabs2025marble} share a common pipeline that first recovers explicit 3D geometry from minimal input in a single feed-forward pass, then progressively grows the world through generative infilling and guided exploration, with several extending to dynamic 4D or text-conditioned scene creation. The explicit nature of this paradigm offers clear advantages for downstream reasoning, as spatial queries are straightforward, physical simulation can operate directly on the geometry, and compositional scene editing is naturally supported. Nevertheless, 3D world models currently lag behind their video counterparts in visual richness and the ability to hallucinate plausible content for unobserved regions.

Despite considerable progress, the optimal integration of these two paradigms remains an open challenge. First, in representation unification, it remains unclear how to bridge explicit 3D structures (pointmaps, 3D Gaussians, meshes) with latent world representations (video embeddings, action-conditioned predictors, symbolic memory) in a way that preserves the editability of the former and the generative capacity of the latter. Second, regarding action-conditioned prediction, current models largely operate in an open-loop fashion; closing the loop between agent actions and world state updates requires feed-forward 3D modules that can perform real-time, incremental updates conditioned on control signals. Third, in terms of persistent and scalable memory, both paradigms struggle with maintaining consistent world state over long time horizons and large spatial extents; architectures that unify geometric memory with generative infilling remain underexplored. Addressing these challenges will likely require feed-forward 3D generators that are deeply co-designed with, rather than appended to, large-scale generative frameworks, moving toward world models that are simultaneously geometrically grounded, visually rich, and interactively controllable.

\subsection{Unified Perception and Reconstruction}
\label{sec:understand}

The future of 3D reconstruction involves not only geometry and appearance but also understanding. Although some early studies regard semantic information as prior information, the real potential lies in deeper integration with foundation models (including LLMs and MLLMs). This integration opens several promising directions.

\noindent\textbf{Multi-Modal Reconstruction.} Models can use visual, textual, auditory, and other sensory inputs (such as thermal signals or inertial signals) to build richer, more accurate and semantically aware 3D scenes. Future research may develop joint training objectives that align geometric representations with text, audio, and semantic labels, enabling queries such as ``tell me where the stove is'' and grounding language in 3D space~\citep{wang2024gsemsplat,zheng2025learningvideos3dworld,xu2025uniuggunified3dunderstanding}. These models can also generate structured and queryable output, including object instances, spatial relationships and affordances, as well as geometry for robots and augmented reality.

\noindent\textbf{Interactive and Editable Scenes.} Interactive reconstruction can be realized by connecting the 3D model with the reasoning capabilities of LLMs. Users can use natural language to issue editing commands, such as ``turn the car red and move it forward''. This direction also points to the development of interactive world models generated through feed-forward inference as basic components of embodied intelligence, robotics and simulation~\citep{wu2025video,chen20254dnex,wu2025geometry}.




\subsection{Open Questions}
\label{sec:openq}
 \textbf{The Spectrum of Reconstruction and Generation.}
 Balancing reconstruction and generation is an important research direction. Determining which regions to reconstruct and which to generate is challenging, especially under occlusion or sparse sampling.
 Modern feed-forward models increasingly operate along the spectrum between these two extremes~\citep{honglrm,zhang2024gs}. Future research should examine this trade-off more clearly and explore how to guide the generative priors to enhance the fidelity of reconstruction, rather than compromise it.

\noindent \textbf{Feed-forward Prediction and Rapid Per-scene Tuning.} A key issue is whether the feed-forward model should be designed to achieve generalization without adaptation, or whether lightweight per-scene refinement should be accepted as a practical component for actual deployment~\citep{roessle2023ganerf,chen2023dbarf}. Achieving a balance between global generalization and effective scene-specific tuning may determine how efficiently such models scale in different environments and tasks.



\section{Conclusion}\label{conclusion}

This survey presents a systematic review of feed-forward 3D reconstruction, a paradigm that fundamentally addresses the scalability and efficiency limitations of classical per-scene optimization by learning to predict 3D representations directly from input images in a single forward pass. 
By organizing the field through a problem-driven taxonomy, we highlight a shared set of core challenges in feed-forward 3D reconstruction, including feature enhancement, geometry awareness, model efficiency, augmentation strategies, and temporal-aware modeling. More importantly, this shared set reveals the common design concerns underlying existing methods, such as robustness under sparse observations, geometric fidelity, computational efficiency, and temporal coherence. It further shows how different approaches address these recurring challenges through distinct architectural choices and trade-offs.
Beyond algorithmic advances, we reclassify benchmarks into geometry-oriented and visual-oriented categories, highlighting the need for more rigorous evaluation protocols that separate geometric accuracy from perceptual fidelity. This distinction is essential to avoid overfitting to view-synthesis metrics.
Furthermore, we review practical applications in autonomous driving, robotics, scene understanding, video generation, and others, showing that feed-forward reconstruction is evolving from a rendering-oriented technique into a broader 3D perception backbone for spatial intelligence systems.
Despite significant progress, several challenges remain open. Future research may focus on standardized benchmarks, scalable scene representations, improving geometric consistency, and deeper integration of reconstruction with generative modeling and semantic understanding.
Overall, we hope this survey provides a structured overview of the field and helps guide future research toward more robust, scalable, and geometry-aware 3D reconstruction systems.



\section*{Declarations}

\subsection*{Ethical Approval}
Not applicable. This article is a survey paper and does not involve any studies 
with human participants or animals performed by any of the authors.

\subsection*{Conflict of Interest}
The authors declare that they have no conflict of interest.

\subsection*{Funding}
This work was not supported by any specific funding.

\subsection*{Data Availability}
Data sharing is not applicable to this article as no datasets were generated or 
analyzed during the current study. This is a survey paper that reviews existing 
published literature.

\subsection*{Consent for Publication}
Not applicable. This article does not contain any individual person's data in any form.

\subsection*{Author Contribution}
Weijie Wang, Qihang Cao, and 
Sensen Gao contributed equally to this work and are co-first authors. They conducted 
the literature review, organized the taxonomy, and wrote the main body of the 
manuscript. Donny Y. Chen led the overall project. Haofei Xu, Wenjing Bian, Songyou Peng, Tat-Jen Cham, and Chuanxia 
Zheng contributed to the discussion, provided feedback, and helped revise the 
manuscript. Andreas Geiger, Jianfei Cai, Jia-Wang Bian, and Bohan Zhuang supervised 
the project, guided the research direction, and revised the final manuscript. All 
authors read and approved the final manuscript.

\subsection*{Consent to Participate}
Not applicable. This article does not involve any human participants.

\begin{appendices}

\end{appendices}

\clearpage
\markright{References}
\bibliographystyle{unsrtnat}
\bibliography{sn-bibliography}

\end{document}